\documentclass{article} 
\usepackage{iclr2027_conference,times}
\usepackage{fix-cm}

\usepackage{amsmath,amsfonts,bm}

\def\eqref#1{equation~\ref{#1}}

\def\1{\bm{1}}

\def\vtheta{{\bm{\theta}}}

\def\vb{{\bm{b}}}

\def\ve{{\bm{e}}}

\def\vh{{\bm{h}}}

\def\vk{{\bm{k}}}

\def\vm{{\bm{m}}}

\def\vo{{\bm{o}}}

\def\vq{{\bm{q}}}

\def\vs{{\bm{s}}}

\def\vu{{\bm{u}}}
\def\vv{{\bm{v}}}

\def\vy{{\bm{y}}}
\def\vz{{\bm{z}}}

\def\evy{{y}}

\def\mA{{\bm{A}}}
\def\mB{{\bm{B}}}

\def\mG{{\bm{G}}}
\def\mH{{\bm{H}}}
\def\mI{{\bm{I}}}
\def\mJ{{\bm{J}}}

\def\mP{{\bm{P}}}
\def\mQ{{\bm{Q}}}
\def\mR{{\bm{R}}}
\def\mS{{\bm{S}}}

\def\mU{{\bm{U}}}
\def\mV{{\bm{V}}}
\def\mW{{\bm{W}}}
\def\mX{{\bm{X}}}

\def\mZ{{\bm{Z}}}

\DeclareMathAlphabet{\mathsfit}{\encodingdefault}{\sfdefault}{m}{sl}
\SetMathAlphabet{\mathsfit}{bold}{\encodingdefault}{\sfdefault}{bx}{n}
\newcommand{\tens}[1]{\bm{\mathsfit{#1}}}

\def\tH{{\tens{H}}}
\def\tI{{\tens{I}}}

\def\gG{{\mathcal{G}}}

\newcommand{\R}{\mathbb{R}}

\usepackage{hyperref}
\usepackage{url}
\usepackage{booktabs, adjustbox}
\usepackage{tabularx}
\usepackage{graphicx}
\usepackage[capitalize,noabbrev,nameinlink]{cleveref}
\usepackage{wrapfig}
\usepackage{enumitem}
\usepackage{xspace}
\usepackage{amsmath}
\usepackage{tabularx}
\usepackage{makecell}
\usepackage{multirow}
\usepackage{subfig}

\newcommand{\jm}[1]{\textcolor{black}{#1}}

\newcommand{\method}{Ephris\xspace}

\newcommand{\mypar}[1]{\textbf{#1}\hspace{4pt}}

\usepackage{acronym}

\acrodef{IID}[IID]{independent and identically distributed}
\acrodef{GNN}[GNN]{graph neural network}
\acrodef{GT}[GT]{graph transformer}
\acrodef{GCN}[GCN]{graph convolutional network}
\acrodef{MLP}[MLP]{multi-layer perceptron}
\acrodef{GFM}[GFM]{graph foundation model}
\acrodef{TFM}[TFM]{tabular foundation model}
\acrodef{ICL}[ICL]{in-context learning}
\acrodef{BGAT}[BGAT]{Bi-axial Graph Attention Network}
\acrodef{SCM}[SCM]{structural causal model}
\acrodef{S2CM}[S$^2$CM]{Spatial Structural Causal Model}
\acrodef{MAB}[MAB]{Multihead Attention Block}
\acrodef{ISAB}[ISAB]{Induced Set Attention Block}
\acrodef{FFN}[FFN]{Feed Forward Network}
\acrodef{PFN}[PFN]{prior-data fitted network}
\acrodef{gICL}[graph ICL]{graph in-context learning}
\acrodef{tICL}[tabular ICL]{tabular in-context learning}
\acrodef{DAG}[DAG]{directed acyclic graph}
\acrodef{HP}[HP]{hyperparameter}
\usepackage{amsthm}
\usepackage{amssymb}
\theoremstyle{plain}

\theoremstyle{definition}

\theoremstyle{remark}

\title{Message Passing Does More with Less for \\
In-Context Learning on Graphs}

\author{
Dooho Lee$^{1,2}$,
Jinmo Lee$^{1,3}$,
Minho Jeong$^{1,3}$,
Kijung Shin$^{2}$,
Jaemin Yoo$^{1,3}$\thanks{Corresponding author: \texttt{jaeminyoo@snu.ac.kr}}
\\
$^1$Nums AI \quad
$^2$KAIST \quad
$^3$Seoul National University
}

\usepackage{pifont}

\iclrfinalcopy 
\begin{document}

\maketitle
\begin{abstract}
Achieving strong performance with graph neural networks (GNNs) typically requires training and hyperparameter tuning for each dataset, incurring repeated costs and effort.
Graph in-context learning (ICL) avoids this by using a single pretrained model to predict unknown node labels directly from labeled context nodes.
Existing approaches, however, rely on dense attention across nodes, making inference increasingly expensive as graphs grow.
In this work, we present Ephris, a new graph in-context learner built on sparse message passing, scaling linearly with the number of node-feature entries and graph edges.
Ephris is pretrained entirely on synthetic graphs generated from structural causal models with diverse graph structures and relational dynamics, exposing the model to varied dependencies among topology, features, and labels.
We evaluate Ephris on 51 node-classification datasets against 15 extensively tuned GNNs and existing graph ICL methods under both high- and low-label train/validation/test splits.
Across both settings, Ephris ranks first on all four aggregate measures: Elo, improvability, average rank, and accuracy.
Its inference cost remains comparable to training a single GNN once, while being over 10 times faster than previous graph ICL models.
Together, these results advance the performance-runtime Pareto frontier, demonstrating that strong graph ICL does not require dense attention.
Code and model weights are available at \url{https://github.com/nums-ai/ephris}.
\end{abstract}

\section{Introduction}
\label{sec:introduction}

Node classification arises across diverse graph-structured domains, including social, e-commerce, and road networks~\citep{hamilton2017inductive,hu2020open,liang2026towards}. \Acfp{GNN} achieve strong predictive performance with message passing between neighboring nodes, allowing node features and graph structure to jointly inform predictions~\citep{hamilton2017inductive,kipf2017semisupervised,veličković2018graph}. 
However, \Acp{GNN} typically require training and careful \ac{HP} tuning for each dataset, incurring substantial costs in time and effort whenever a new graph is encountered~\citep{luo2024classic,platonov2026a}. 

\Ac{ICL} offers an alternative paradigm: a pretrained model predicts missing labels on unseen datasets without parameter updates, using observed labels 
as context. \Acp{PFN}~\citep{muller2022transformers} learn this capability by pretraining Transformers~\citep{vaswani2017attention} on synthetic task priors, and \acp{TFM} enrich these priors with \acfp{SCM}~\citep{pearl2000models} to capture diverse feature-label relationships. 
Recently, \acp{TFM} have surpassed \ac{HP}-tuned supervised methods on tabular benchmarks~\citep{jäger2026tabpfn35technicalreport,zhang2026limix} and seen broader downstream adoption~\citep{hicham2026tabular,wu2026panmetai}, challenging the convention of training and tuning a separate model for each dataset.

For graphs, several methods extend this \ac{PFN} framework to node classification by combining graph-aware architectures with synthetic graph priors. NodePFN~\citep{choi2026learning} augments the TabPFN architecture~\citep{muller2022transformers} with a parallel message-passing branch and pretrains on synthetic graphs with varying homophily. GraphPFN~\citep{eremeev2026graphpfn} adds graph-attention adapters to pretrained LimiX~\citep{zhang2025limix} and trains them on tasks sampled from an \ac{SCM} augmented with graph convolution.
Their improvements over \ac{HP}-tuned \acp{GNN} on several datasets suggest that graph \ac{ICL} offers a practical alternative to dataset-specific training, as it has in tabular data.

\begin{wrapfigure}[18]{r}{0.44\linewidth}
    \vspace{-4pt}
    \centering
    \includegraphics[
        width=\linewidth,
    ]{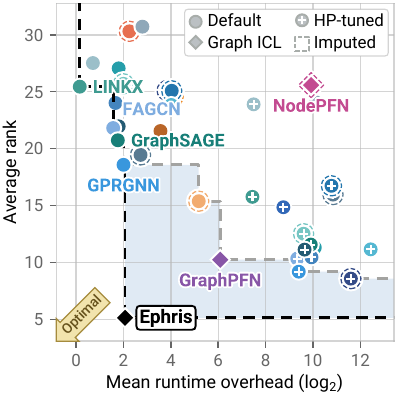}
    \vspace{-22pt}
    \caption{Performance-runtime Pareto plot across 51 datasets under the 50/25/25 split.}
    \label{fig:pareto_plot}
\end{wrapfigure}

However, existing methods face two key limitations that challenge the success of graph \ac{ICL}.
First, eliminating parameter updates does not necessarily make adaptation computationally efficient. Existing graph \ac{ICL} models augment \ac{TFM} backbones with graph-specific modules~\citep{choi2026learning,eremeev2026graphpfn}, thereby retaining the backbone's dense self-attention over labeled context nodes. Their computation therefore scales quadratically with context size, making inference increasingly expensive as more labeled nodes are provided and, in some datasets, even slower than training and tuning a \ac{GNN}.

Second, outperforming tuned \acp{GNN} in their original evaluations does not necessarily establish broad competitiveness.
NodePFN~\citep{choi2026learning} is evaluated only on graphs with fewer than 50,000 nodes and against 2 \ac{GNN} baselines, while GraphPFN~\citep{eremeev2026graphpfn} includes larger graphs but only datasets with at most 500 features and 4 \ac{GNN} baselines.
In our main experiments (\Cref{sec:experiments}), we evaluate both methods more broadly across 51 node-classification datasets, spanning up to 568,795 nodes and 8,710 features per node, against 15 \acp{GNN}.
Their reported gains do not consistently persist under this broader evaluation, with tuned \acp{GNN} remaining stronger on many datasets.

To overcome both limitations, we introduce \textbf{\method{}}, a new graph
in-context learner designed to push the performance-runtime frontier
(\Cref{fig:pareto_plot}).
We first develop a scalable architecture for graph \ac{ICL} that operates entirely through sparse message passing without dense cross-node attention, scaling linearly with node-feature entries and graph edges (\Cref{sec:architecture}).
We then introduce a synthetic graph prior that extends \acp{SCM} with diverse graph structures and relational dynamics, exposing the model to varied dependencies among topology, features, and labels (\Cref{sec:structure-conditioned-prior}).

We evaluate \method{} on 51 node-classification datasets against 15 tuned
\acp{GNN} and 6 \acp{GFM}, including NodePFN and GraphPFN, under high-label
(50/25/25) and low-label (10/10/80) train/validation/test splits.
Across this broad evaluation, no prior \ac{GFM} surpasses tuned \acp{GNN}
overall; GraphPFN comes closest, ranking 4th and 10th by Elo in the two
regimes.
In contrast, \method{} ranks 1st in Elo, improvability, average rank, and
accuracy (defined in \Cref{app:evaluation-metrics}) under both regimes.
It achieves this performance at the cost of a single \ac{GNN} training run,
while being more than $13\times$ faster than GraphPFN.
Together, these results show that dense attention is not necessary for strong
graph \ac{ICL}, advancing the performance-runtime frontier and making graph
\ac{ICL} a practical alternative to per-dataset training and tuning.
Code and pretrained weights are available at
\href{https://anonymous.4open.science/r/ephris/}{link}.
\section{Background and Related Work}
\label{sec:background}

\mypar{Graph in-context learning.}
Consider a graph $\gG=(\mathcal V,\mathcal E)$ with $N=|\mathcal V|$ nodes and $E=|\mathcal E|$ edges, with adjacency matrix $\mA\in\{0,1\}^{N\times N}$, where $A_{ij}=1$ if $(i,j)\in\mathcal E$ and $0$ otherwise. Node features are collected in $\mX\in\R^{N\times F}$, with each row an $F$-dimensional vector. Let $\vy\in\{1,\ldots,C\}^N$ denote node labels, where $y_i$ is the label of node $i$ and $C$ the number of classes. Nodes are partitioned into disjoint training and test nodes, which we call context nodes $\mathcal V_{\mathrm{ctx}}$ and query nodes $\mathcal V_{\mathrm{qry}}$, respectively.

Following prior graph \ac{ICL} models~\citep{choi2026learning,eremeev2026graphpfn}, we pretrain \method{} on synthetic graph tasks. Each task $\mathcal D=(\mA,\mX,\vy,\mathcal V_{\mathrm{ctx}})$ is sampled from a prior $\Pi$, with $\mathcal V_{\mathrm{qry}}=\mathcal V\setminus\mathcal V_{\mathrm{ctx}}$. Given the graph structure, node features, and context labels, the model predicts unobserved query labels by minimizing their expected average negative log-likelihood:
\begin{equation}
\mathcal L_{\mathrm{ICL}}(\vtheta)
=
-\mathbb E_{\mathcal D\sim\Pi}\!
\bigl(
\frac{1}{|\mathcal V_{\mathrm{qry}}|}
\sum_{i\in\mathcal V_{\mathrm{qry}}}
\log p_{\vtheta}\!\left(
\evy_i
\mid
i,\mA,\mX,
\vy_{\mathcal V_{\mathrm{ctx}}}
\right)
\bigr).
\label{eq:structured-icl-objective}
\end{equation}
This objective trains a single set of parameters across synthetic tasks, requiring the model to infer each task-specific prediction rule from its observed context.
At inference, the pretrained model applies this capability to unseen graphs without updating $\vtheta$.
For \method{}, $p_{\vtheta}$ is defined by the architecture in \Cref{sec:architecture}, while $\Pi$ is defined by the synthetic task prior in \Cref{sec:structure-conditioned-prior}.


\paragraph{Structural causal models.}
Recent \acp{TFM} use \acfp{SCM} to generate synthetic tasks with diverse feature-label relationships~\citep{hollmann2025accurate,qu2026tabiclv}.
An \ac{SCM} defines a \ac{DAG} $\mathcal G_{\mathrm{SCM}}$ specifying dependencies over potentially vector-valued intermediate variables.
For $N$ samples, $\mU^{(a)}$ collects the values of variable $a$, with one row per sample.
Let $\operatorname{pa}(a)$ denote its parents.
Variables are generated in topological order as
\begin{equation}
\mU^{(a)}
=
\mathcal T_a(
f_a(\{\mU^{(b)}\}_{b\in\operatorname{pa}(a)})
).
\label{eq:scm-generation}
\end{equation}
For root variables, $f_a(\varnothing)$ generates sampled noise.
For other variables, $f_a$ is drawn from function families such as \acp{MLP}, tree ensembles, and linear or quadratic mappings, inducing diverse relationships among variables.
The transform $\mathcal T_a$ applies operations such as normalization and noise injection.
Features and labels are extracted from the variables and transformed as needed, for example by discretizing continuous values into categorical features or class labels.

\mypar{Graph foundation models.}
\Acp{GFM} aim to transfer pretrained models across graph datasets~\citep{liu2024one,wang2024towards}.
Our work aligns with efforts to enable node classification across datasets with varying graph structures, feature and label spaces.
These methods differ in which components they transfer and how they adapt to a new dataset.
GraphAny~\citep{zhao2024graphany} transfers an aggregator over linear least-squares predictors fitted to each dataset.
GVT~\citep{lee2026view} and Node4All~\citep{lee2026node4all} transfer feature encoders while training lightweight downstream predictors.
G2T-FM~\citep{eremeev2025turning} augments node features with graph-derived features and applies a pretrained \ac{TFM} for prediction.
NodePFN~\citep{choi2026learning} and GraphPFN~\citep{eremeev2026graphpfn} instead use observed node labels as context for graph \ac{ICL}, avoiding dataset-specific parameter updates at all.
\method{} follows this graph \ac{ICL} approach, aiming to improve predictive performance and computational scalability.
We compare against all six methods in our evaluation.
\section{\method: A Scalable Graph In-context Learner}
\label{sec:architecture}




Recent in-context learning methods broadly follow two architectural paradigms:
\emph{bi-axial} architectures alternate dense attention across samples and features~\citep{hollmann2025accurate,zhang2025limix,zhang2026limix,zhang2026mitra}, while \emph{compress-then-ICL} architectures compress each sample into a fixed-dimensional representation before performing \ac{ICL} through attention across samples~\citep{qu2025tabicl,qu2026tabiclv}.
\method{} redesigns the latter architecture for graph-structured data through five stages:
\begin{equation*}
(\underbrace{\mX}_{N\times F},\vy_{\mathcal V_{\mathrm{ctx}}})
\xrightarrow{\text{\ding{172} Token.}}
\underbrace{\tH^{(0)}}_{N\times F\times d}
\xrightarrow{\text{\ding{173} Ref.}}
\underbrace{\tH^{(L_{\mathrm{ref}})}}_{N\times F\times d}
\xrightarrow{\text{\ding{174} Comp.}}
\underbrace{\mZ^{(0)}}_{N\times D}
\xrightarrow{\text{\ding{175} ICL}}
\underbrace{\mZ^{(L_{\mathrm{ICL}})}}_{N\times D}
\xrightarrow{\text{\ding{176} Head}}
\underbrace{\widehat{\mP}}_{N\times C}.
\label{eq:compress-then-icl}
\end{equation*}

\begin{figure}[t]
\centering
\includegraphics[width=\linewidth]{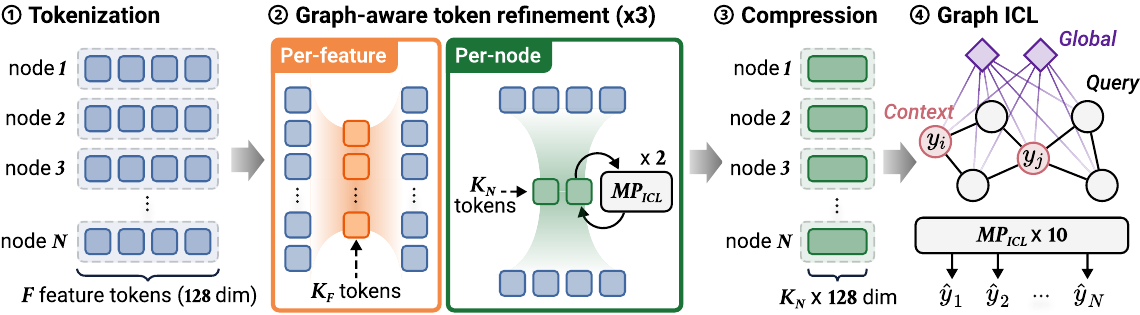}
\vspace{-5mm}
\caption{
\textbf{Architecture overview of \method.}
\method{} tokenizes each scalar node-feature value into a $d$-dimensional token, compresses feature tokens into fixed-size node representations while integrating distributional and graph context, and performs \ac{ICL} through stacked message passing.
}
\vspace{-3mm}
\label{fig:architecture_overview}
\end{figure}

\jm{We first outline these stages, which are designed to support diverse graph tasks with shared parameters (\S\ref{sec:architecture-overview}).
Then, we introduce the message-passing block $\operatorname{MP}_{\mathrm{ICL}}$, which is the core component of \method{} that is utilized in two different stages of the architecture (\S\ref{sec:graphmp}).
Finally, we analyze computational complexity and empirically compare scalability with prior graph \ac{ICL} models (\S\ref{subsec:complexity}).
An overview is shown in \Cref{fig:architecture_overview}, with full implementation details in \Cref{app:architecture}.}

\subsection{Architecture Overview}
\label{sec:architecture-overview}

\mypar{\ding{172} Tokenization.}
\jm{We first map each scalar node feature into a common $d$-dimensional token space.
A shared \ac{MLP} is used for all $F$ features.
We also incorporate context labels through learned class embeddings, allowing subsequent layers to use the observed label context; the class embeddings are added to an intermediate representation.
The resulting tokens form $\tH^{(0)}$.}

\mypar{\ding{173} Graph-aware token refinement.}
Before compression, we apply $L_{\mathrm{ref}}$ blocks, each consisting of per-feature refinement to capture feature distributions across nodes, followed by per-node refinement to incorporate graph structure into each node's feature tokens.

\emph{Per-feature refinement.}
For each feature, we use the induced attention~\citep{lee2019set} to exchange information across nodes: $K_F$ learned inducing tokens attend to the node tokens to \emph{gather} a summary, then node tokens attend to it to receive the \emph{broadcast}.
For feature $f$ in block $\ell$:
\begin{equation*}
\mJ_f^{(\ell)}
=
\operatorname{Gather}_{F}^{(\ell)}(\mH_{:f}^{(\ell-1)}),
\qquad
\widetilde{\mH}_{:f}^{(\ell)}
=
\operatorname{Broadcast}_{F}^{(\ell)}(\mH_{:f}^{(\ell-1)},\mJ_f^{(\ell)}),
\end{equation*}
where $\mJ_f^{(\ell)}$ contains the $K_F$ summary tokens gathered for feature $f$.
\jm{Note that we gather from both context and query nodes, incorporating the full observed distribution, while most \acp{TFM} restrict gathering to context samples so that each query prediction is independent of other queries.}

\emph{Per-node refinement.}
We then gather each node's feature tokens into a compact summary using $K_N$ learned inducing tokens.
Two message-passing blocks exchange information among these summaries, and broadcast returns the updates to the feature tokens:
\begin{equation*}
\mI_i^{(\ell)}
=
\operatorname{Gather}_{N}^{(\ell)}(\widetilde{\mH}_{i:}^{(\ell)}),
\quad
\widehat{\tI}^{(\ell)}
=
\operatorname{MP}_{\mathrm{ICL}}^{\circ 2}(\tI^{(\ell)},\mA),
\quad
\mH_{i:}^{(\ell)}
=
\operatorname{Broadcast}_{N}^{(\ell)}(\widetilde{\mH}_{i:}^{(\ell)},\widehat{\mI}_i^{(\ell)}),
\end{equation*}
where $\mI_i^{(\ell)}$ contains node $i$'s summary tokens, and $\tI^{(\ell)}$ collects them across nodes.
The message-passing block includes concatenating each node's summary tokens into a vector before propagation and splitting the updated vector back into tokens before broadcasting (details in \S\ref{sec:graphmp}).

\emph{Alternation.}
Alternating per-feature and per-node refinement across $L_{\mathrm{ref}}$ blocks allows feature distributions and graph structure to jointly guide which information is retained during compression.

\mypar{\ding{174} Compression.}
After refinement, an additional gather operation summarizes each node's features using $K_N$ inducing tokens.
Concatenating these tokens yields a representation of width $D=K_Nd$, independent of feature count.
The resulting matrix $\mZ^{(0)}$ enters the \ac{ICL} stage.

\mypar{\ding{175} Graph in-context learning.}
The compressed node representations pass through $L_{\mathrm{ICL}}$ successive $\operatorname{MP}_{\mathrm{ICL}}$ blocks, replacing the dense attention across samples used in existing in-context learners:
\begin{equation}
\mZ^{(L_{\mathrm{ICL}})}
=
\operatorname{MP}_{\mathrm{ICL}}^{\circ L_{\mathrm{ICL}}}(\mZ^{(0)},\mA).
\label{eq:graph-icl}
\end{equation}
Through successive rounds of message passing, the model infers a task-specific predictive rule relating node features and graph connectivity to the observed context labels, progressively updating query-node representations for label prediction.

\mypar{\ding{176} Prediction head.}
A shared head maps each query node's final representation to class logits.
As done in previous work~\citep{qu2026tabiclv,hollmann2025accurate}, we use ten learned class embeddings and ten output logits.
For $C\leq10$, softmax is applied to the first $C$ logits.
For $C>10$, error-correcting output codes (ECOC)~\citep{dietterich1994solving} decompose the task into classification problems with at most ten classes.
Each original class is scored by averaging the log probabilities assigned to its code entries, and softmax converts these scores into class probabilities.

\subsection{Message Passing Block for ICL}
\label{sec:graphmp}

The message passing block $\operatorname{MP}_{\mathrm{ICL}}$ is the central building block of \method{}.
The block is designed to accommodate diverse graph structures and predictive relationships across tasks with a single set of shared parameters.
Specifically, we introduce four complementary design choices below.

\textbf{Dynamic neighborhood aggregation.}
Graph structure provides plausible interactions, but the relevance of each connection for prediction can vary across tasks, graphs, and nodes.
We therefore use local attention~\citep{shi2020masked} to dynamically weight neighboring messages based on their current representations.
Let $\vh_i\in\mathbb R^{d_h}$ denote the representation of node $i$, where $d_h$ is the hidden dimension, and let $\mathcal N(i)$ denote its neighborhood including $i$ itself.
We compute
\begin{equation}
\alpha_{ij}
=
\operatorname{softmax}_{j\in\mathcal N(i)}
\left(
d_h^{-1/2}\vq_i^{\mathsf T}\vk_j
\right),
\qquad
\mathcal M_i(\vh)
=
{\textstyle\sum}_{j\in\mathcal N(i)}
\alpha_{ij}\vv_j,
\label{eq:graphmp-attention}
\end{equation}
where $\vq_i=\mW_Q\vh_i$, $\vk_i=\mW_K\vh_i$, and $\vv_i=\mW_V\vh_i$ are the query, key, and value projections.
This lets the model adapt which neighbors to emphasize as their representations evolve across layers.

\textbf{Neighborhood-aware attention.}
Neighborhood sizes also vary widely across nodes and graphs, ranging from only a few to thousands~\citep{hu2020open}.
With more neighbors competing within the same softmax, attention can become increasingly diffuse, a phenomenon known as attention dilution~\citep{zhang2024selective}.
To address this, we extend TabICLv2's query-aware attention scaling~\citep{qu2026tabiclv} to jointly account for neighborhood size:
\begin{equation}
\widetilde{\vq}_i
=
\vq_i
\odot f_{\vtheta}\!\left(\log |\mathcal N(i)|\right)
\odot
\left[1+\tanh g_{\vtheta}(\vq_i)\right],
\qquad
\alpha_{ij}
=
\operatorname{softmax}_{j\in\mathcal N(i)}
\left(
d_h^{-1/2}\widetilde{\vq}_i^{\mathsf T}\vk_j
\right),
\label{eq:graphmp-scaling}
\end{equation}
where $\odot$ denotes element-wise multiplication, and $f_{\vtheta}$ and $g_{\vtheta}$ are two-layer MLPs.
This lets the model adapt attention sharpness jointly to the current representation and neighborhood size.

\textbf{Local and global communication.}
Graph edges can be too sparse, noisy, or weakly aligned with the prediction task.
To improve robustness in these graphs, we introduce $K_G$ learned global nodes $\mathcal V_g$ that enable interactions beyond the observed edges.
Each global node connects bidirectionally to every original node~\citep{cai2023connection}.
For each node $i$, we augment its neighborhood as
\begin{equation}
\mathcal N^{+}(i)
=
\mathcal N(i)\cup\mathcal V_g,
\qquad
|\mathcal N^{+}(i)|
=
|\mathcal N(i)|+K_G,
\label{eq:augmented-neighborhood}
\end{equation}
and apply the same neighborhood-aware attention over $\mathcal N^{+}(i)$.
The global nodes allow arbitrary nodes to exchange information within two message-passing layers while adding only $\mathcal O(NK_G)$ interactions, which scale linearly in $N$ for fixed $K_G$.

\textbf{Deep residual propagation.}
The interaction range needed to infer a prediction rule can vary across graphs and tasks.
We therefore design $\operatorname{MP}_{\text{ICL}}$ for deep stacking using residual connections:
\begin{equation}
\bar{\vh}_i
=
\operatorname{LN}\bigl(\vh_i+\mathcal M_i(\vh)\bigr),
\qquad
\operatorname{MP}_{\text{ICL}}(\vh)_i
=
\operatorname{LN}\bigl(
\bar{\vh}_i+\operatorname{FFN}(\bar{\vh}_i)
\bigr),
\label{eq:graphmp}
\end{equation}
where $\operatorname{LN}$ and $\operatorname{FFN}$ denote layer normalization and a position-wise feed-forward network, respectively.
Stacking these blocks progressively expands each node's receptive field, allowing information to propagate over increasingly long graph distances.
Together, these four design choices let $\operatorname{MP}_{\text{ICL}}$ adapt what information to aggregate, where to communicate, and how far to propagate it, while retaining sparse, scalable computation.

\begin{wraptable}[8]{r}{0.35\linewidth}
    \centering
    \small
    \setlength{\tabcolsep}{3pt}
    \vspace{-5mm}
    \caption{Theoretical complexity.}
    \vspace{-1mm}
    \label{tab:complexity}
    \begin{tabular}{@{}ll@{}}
        \toprule
        Model & Computation \\
        \midrule
        NodePFN & $\mathcal O(NF+N^2+E)$ \\
        GraphPFN & $\mathcal O(N^2F+NF^2+EF)$ \\
        \midrule
        \textbf{\method{}} & $\mathcal O(NF+E)$ \\
        \bottomrule
    \end{tabular}
\end{wraptable}

\subsection{Computational Complexity}
\label{subsec:complexity}

\mypar{Theoretical complexity.}
We analyze computational complexity under fixed hidden dimensions, inducing-token counts, network depths, and context ratio.
Per-feature refinement costs $\mathcal O(NF)$, per-node refinement including message passing costs $\mathcal O(NF+E)$, and graph \ac{ICL} costs $\mathcal O(N+E)$.
Each prediction pass therefore requires $\mathcal O(NF+E)$ computation, scaling linearly with the number of node-feature entries and graph edges, rather than quadratically with the number of nodes or features as in prior graph \ac{ICL} models (\Cref{tab:complexity}).

\begin{wrapfigure}[10]{r}{0.35\textwidth}
    \vspace{-9mm}
    \centering\includegraphics[
        width=\linewidth,
    ]{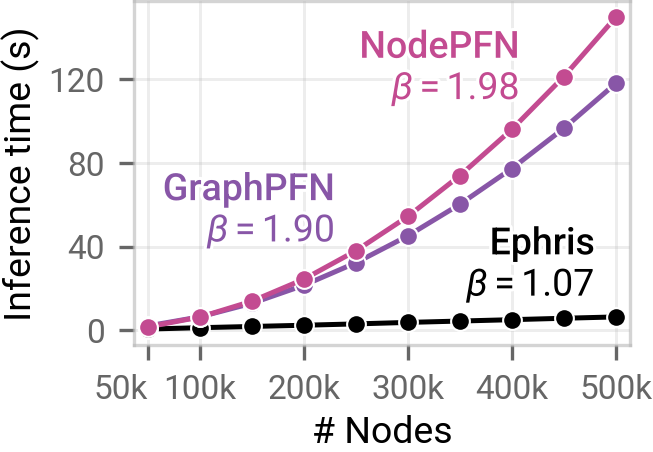}
    \vspace{-7mm}
    \caption{Inference runtime scaling; median over three seeds.}
    \label{fig:scaling}
\end{wrapfigure}

\mypar{Empirical scaling.}
We measure inference runtime on synthetic graphs ranging from 50K to 500K nodes, with 32 features, average degree 8, and a 50\% context ratio.
All methods run without ensembling on a single NVIDIA H200 GPU; runtime includes preprocessing and inference but excludes model loading.
We quantify scaling by fitting $T(N)=c+aN^\beta$ over the measured range.
As shown in \Cref{fig:scaling}, \method{} scales nearly linearly ($\beta=1.07$), whereas NodePFN ($\beta=1.98$) and GraphPFN ($\beta=1.90$) scale nearly quadratically. The resulting runtime gap widens rapidly with graph size.

\section{Pretraining with Diverse Synthetic Graphs}
\label{sec:structure-conditioned-prior}

We outline the synthetic graph generation pipeline that defines the pretraining prior $\Pi$ in \Cref{eq:structured-icl-objective}, followed by the training curriculum and optimization procedure.
Detailed descriptions of the prior and training setup are provided in \Cref{app:prior} and \Cref{app:implementation-details}, respectively.

\subsection{Synthetic Graph Prior}
\label{subsec:s2cm-graph}

Our pipeline has three stages: sampling a graph topology, generating features and labels conditioned on the sampled graph, and post-processing the features into diverse form of representations.

\mypar{Graph sampling.}
Real-world graphs often exhibit a mixture of connectivity patterns.
We capture this diversity by combining three graph-generation rules:
\emph{group-based} connectivity favors edges within or across sampled groups, as in stochastic block models~\citep{holland1983stochastic};
\emph{degree-heterogeneous} connectivity samples endpoints according to node-wise weights, producing heterogeneous degrees and hubs as in Chung-Lu graphs~\citep{chung2002connected};
and \emph{ordering-based} connectivity links nodes according to a sampled ordering, introducing route-like connection patterns.
We sample mixture weights for each graph so that these patterns coexist in varying proportions.

\mypar{Feature and label sampling.}
We generate features and labels conditioned on the sampled graph by incorporating graph propagation into the SCM.
Inspired by GraphPFN~\citep{eremeev2026graphpfn}, we extend \Cref{eq:scm-generation} as
\begin{equation}
\widetilde{\mU}^{(a)}
=
f_a(\{\mU^{(b)}\}_{b\in\operatorname{pa}(a)}),
\qquad
\mU^{(a)}
=
\mathcal T_a(\operatorname{Propagate}_a(\widetilde{\mU}^{(a)},\mA)).
\label{eq:s2cm-latent-generation}
\end{equation}
Here, $\widetilde{\mU}^{(a)}$ collects variable $a$'s values across nodes.
Before applying $\mathcal T_a$, propagation updates these values by aggregating neighbor information using operators such as mean, max, or min.
Graph structure thus shapes the intermediate variables from which features and labels are generated.

We also diversify the \emph{relational dynamics} governing how graph structure influences these variables.
For each variable selected for propagation, we sample one of three rules:
\emph{diffusion} repeatedly mixes node and neighbor states;
\emph{cascade} starts from a small set of active nodes and updates an inactive node only when agreement between its state and the aggregated states of active neighbors reaches a sampled threshold;
and \emph{degree-dependent mixing} adjusts the strength of neighbor influence based on node degree and neighbor agreement.
These rules capture gradual, conditional, and degree-dependent effects, respectively.
We further vary their behavior by sampling rule-specific mixing coefficients, propagation steps, and activation thresholds.

\mypar{Post-processing.}
Finally, we diversify how generated features appear to the model.
Real-world node features range from meaningful attributes to image or text embeddings whose information is distributed across dimensions~\citep{yan2023comprehensive}.
We therefore sample three representations:
\emph{tabular}, which preserves the generated features;
\emph{embedding}, which mixes all features through a linear or nonlinear transformation followed by standardization;
and \emph{hybrid}, which transforms only a subset.


\subsection{Training Schedule and Optimization  }
\label{subsec:optimization}

\begin{wraptable}[9]{r}{0.34\linewidth}
    \centering
    \small
    \vspace{-8pt}
    \setlength{\tabcolsep}{2.5pt}
    \vspace{-4pt}
    \caption{Two-stage training setup.}
    \label{tab:training-schedule}
    \vspace{-4pt}
    \begin{tabular}{@{}lcc@{}}
        \toprule
        & Stage 1 & Stage 2 \\
        \midrule
        \# Node & 512--1,024 & 128--16,384 \\
        \# Feature & 2--128 & 2--1,024 \\
        \# Steps & 50,000 & 10,000 \\
        \midrule
        Muon LR & $8{\times}10^{-4}$ & $4{\times}10^{-5}$ \\
        AdamW LR & $8{\times}10^{-4}$ & $10^{-5}$ \\
        \bottomrule
    \end{tabular}
\end{wraptable}

\paragraph{Training curriculum.}
Generalizing across graph datasets requires exposure to diverse graph sizes and feature dimensions, but larger graphs and higher-dimensional features can increase pretraining costs signficantly.
We therefore adopt a two-stage curriculum.
Stage~1 develops the model's in-context prediction capability on relatively small graphs over 50,000 steps.
Stage~2 then adapts the model to a broader range of graph sizes and feature dimensions over 10,000 additional steps.
Each step uses 64 freshly generated graphs, yielding 3.84 million graphs in total.

\paragraph{Optimization.}
We optimize eligible matrix-valued parameters with Muon~\citep{jordan2024muon} and the remaining parameters with AdamW~\citep{loshchilov2017decoupled}.
Each stage uses linear warmup followed by cosine decay.
In Stage~2, we use lower peak learning rates to preserve learned capabilities while adapting to larger graphs and higher-dimensional features.
\begin{figure}[t]
\centering
\vspace{-3mm}
\includegraphics[width=\linewidth]{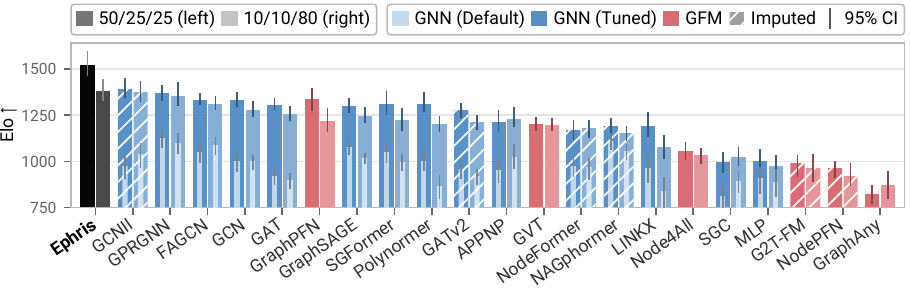}
\vspace{-7mm}
\caption{
\textbf{Elo across label regimes.}
Each bar nests a default \ac{GNN} within its \ac{HP}-tuned counterpart.
Hatching marks OOM results imputed with GCN (Default); error bars show 95\% confidence intervals.
Methods are ordered by mean Elo across regimes.
Ephris achieves the highest Elo in both.
}
\vspace{-3mm}
\label{fig:elo_comparison}
\end{figure}

\section{Experiments}
\label{sec:experiments}

\mypar{Model configuration.}
Our model has 50.16M parameters and uses 128-dimensional feature tokens.
Three refinement blocks combine per-feature refinement with 128 inducing tokens and per-node refinement with four inducing tokens.
Compression concatenates four tokens into a 512-dimensional node representation, which is then processed by ten $\operatorname{MP}_{\mathrm{ICL}}$ layers for graph \ac{ICL}.
Each $\operatorname{MP}_{\mathrm{ICL}}$ layer uses eight global nodes.
Pretraining follows the two-stage curriculum on 3.84M synthetic graphs with no exposure to real-world datasets, requiring 16 GPU-days in our environment.
Full model configurations and details of the experimental environment are provided in \Cref{app:implementation-details}.

\mypar{Baselines.}
We compare \method{} against 21 baselines: 15 \acp{GNN} and six recent \acp{GFM}.
The GNN baselines cover neighborhood aggregation~\citep{kipf2017semisupervised,hamilton2017inductive}, attention-based aggregation~\citep{veličković2018graph,brody2021attentive}, decoupled feature transformation and propagation~\citep{gasteiger2018predict,wu2019simplifying}, and architectures for deep propagation and heterophilous graphs~\citep{chen2020simple,chien2020adaptive,bo2021beyond}.
They also include scalable \acfp{GT} that capture graph-wide interactions through multi-hop representations, kernelized attention, or linear attention~\citep{chen2022nagphormer,wu2022nodeformer,deng2024polynormer,wu2023sgformer}.
The GFM baselines are GraphAny~\citep{zhao2024graphany}, GVT~\citep{lee2026view} Node4All~\citep{lee2026node4all}, G2T-FM~\citep{eremeev2025turning}, NodePFN~\citep{choi2026learning}, and GraphPFN~\citep{eremeev2026graphpfn}, introduced in \Cref{sec:background}.

\mypar{Evaluation protocol.}
We evaluate all methods on 51 node-classification datasets spanning six application domains.
The datasets range from 183 to 568,795 nodes, 12 to 8,710 features, and 2 to 70 classes, with adjusted label homophily from $-0.30$ to $0.94$.
Each dataset is evaluated under high-label and low-label regimes, using train/validation/test proportions of 50/25/25 and 10/10/80, respectively, with five splits per regime.
Following TabArena~\citep{erickson2026tabarena}, each \ac{GNN} is evaluated under two tuning budgets: \emph{default}, using default hyperparameters, and \emph{tuned}, selecting the best of 200 configurations by validation performance.
Supervised baselines use validation data for early stopping and hyperparameter selection.
\method{} uses only training labels as context and predicts in a single inference pass.
Dataset and baseline details, hyperparameter search spaces for each \ac{GNN}, and adaptation procedures for each \ac{GFM} are provided in \Cref{app:evaluation-details}.

\mypar{Metrics.}
We evaluate predictive performance and adaptation cost.
Performance is summarized by Elo, improvability, average rank, and average accuracy.
Elo aggregates split-level pairwise wins, draws, and losses, while improvability measures the normalized gap to the best-performing model~\citep{erickson2026tabarena}.
The first three metrics use AUROC for binary tasks and accuracy for multiclass tasks; average accuracy uses accuracy throughout.
Adaptation cost covers a single training run for default \acp{GNN}, the full hyperparameter search for tuned \acp{GNN}, each \ac{GFM}'s adaptation procedure, and inference for \method{}.
We report this cost as runtime overhead, defined as the log-aggregated slowdown relative to the fastest method.
Full definitions are in \Cref{app:evaluation-metrics}.

\subsection{Main Results}

Due to space constraints, we show selected plots here.
Complete results, including tables, bar plots, Pareto plots, subgroup radar charts, and pairwise win matrices, are provided in \Cref{app:full-results}.

\mypar{Predictive performance.}
A broader evaluation revealed a substantial gap between existing \acp{GFM} and extensively tuned \acp{GNN}.
In particular, GCNII~\citep{chen2020simple} and GPRGNN~\citep{chien2020adaptive} demonstrated leading results under both label regimes.
GraphPFN~\citep{eremeev2026graphpfn} emerged as the strongest existing \ac{GFM}, but did not outperform either overall.
Therefore, the advantages reported for existing \acp{GFM} do not hold for broader datasets and stronger baselines.

\method{} closes this gap, ranking first in Elo, improvability, average rank, and average accuracy in both label regimes.
Against tuned GCNII, the strongest baseline, it achieves pairwise win rates of 67\% and 54\% in the high- and low-label regimes, respectively.
These gains are particularly notable because supervised baselines use validation labels for early stopping and hyperparameter selection, whereas \method{} uses none.
Against GraphPFN, the strongest prior \ac{GFM}, it wins 71\% of comparisons in both regimes.
\method{} achieves these results through sparse message passing, demonstrating that dense all-pairs attention is not required for strong predictive performance in graph \ac{ICL}.

\begin{figure}[t]
\centering
\vspace{-4mm}
\includegraphics[width=\linewidth]{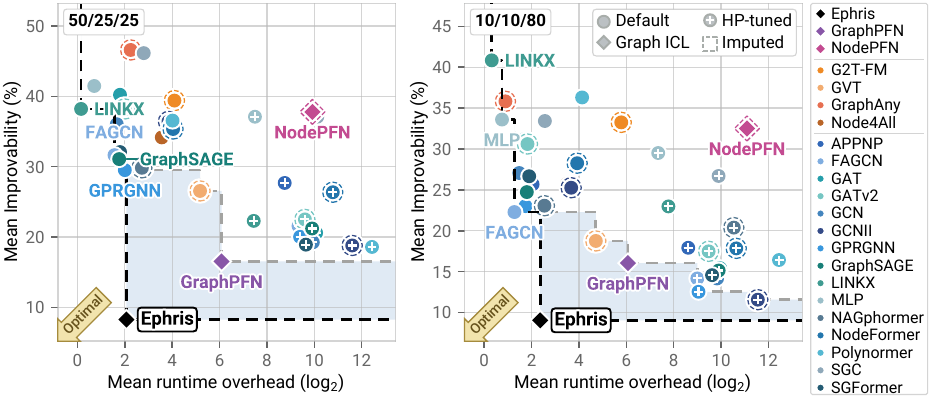}
\vspace{-6mm}
\caption{
\textbf{Performance-runtime trade-offs.}
Mean improvability versus runtime overhead under high-label (left) and low-label (right) regimes; lower is better on both axes.
Black and gray dashed lines show Pareto frontiers with and without Ephris, with the improvement shaded blue.
}
\vspace{-4mm}
\label{fig:main_pareto}
\end{figure}

\mypar{Performance-runtime trade-off.}
A central promise of \acp{GFM} is to reduce adaptation costs by reusing pretrained knowledge.
GVT~\citep{lee2026view} and GraphPFN~\citep{eremeev2026graphpfn} partially fulfill this promise: both improve on default \acp{GNN} at far lower cost than 200-trial \ac{HP} tuning, reaching the Pareto frontier in both label regimes (\Cref{fig:main_pareto}).
However, their adaptation costs remain substantial in practice.
Relative to a single GCN training run, GVT requires approximately $9.4$-$11.5\times$ the runtime and GraphPFN $21.4$-$24.1\times$ across the two label regimes.

\begin{wrapfigure}[17]{r}{0.44\textwidth}
    \centering
    \vspace{-17pt}
    \includegraphics[width=\linewidth]{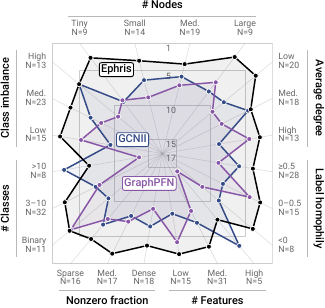}
    \vspace{-16pt}
    \caption{\textbf{Subgroup performance.}
    Average ranks under the high-label regime.}
    \label{fig:subgroup-rank-high-label}
\end{wrapfigure}

\method{} also closes this efficiency gap.
Across the two label regimes, its inference costs $1.33$--$1.84\times$ a single GCN training run and is $13.1$--$16.1\times$ faster than GraphPFN, demonstrating the efficiency of replacing dense attention with message passing for graph \ac{ICL}.
Together with its leading predictive performance, this pushes the performance-runtime Pareto frontier forward, with \method{} remaining the only \ac{GFM} on the frontier across all metrics (\Cref{fig:pareto-full}).

\mypar{Subgroup Analysis.}
We assess whether the overall gains of \method{} persist across graph and task characteristics.
\Cref{fig:subgroup-rank-high-label} reports average ranks across subgroups under the high-label regime, with complete results and subgroup definitions in \Cref{app:full-results}.
\method{} outperforms GraphPFN in nearly all subgroups under both label regimes, indicating broad improvements of graph \ac{ICL} rather than gains confined to particular datasets.
Against the runner-up, tuned GCNII, it leads in most subgroups but falls behind on datasets with more than 5,000 features or ten classes.
Both exceed the ranges seen during pretraining, suggesting that broader pretraining coverage may help close these gaps.

\subsection{Beyond Ordinary Graphs and Random Splits}
\label{subsec:beyond}
Having established strong performance across diverse graph datasets, we further test whether \method{} generalizes beyond graphs and random train-test splits, considering hypergraph transfer and distribution shift (\Cref{app:beyond-standard}).
On ten AllSet hypergraph datasets~\citep{chien2021you}, \method{} ranks first or second on eight using simple clique or incidence representations, suggesting that it also provide a promising alternative to per-dataset training and \ac{HP} tuning of supervised hypergraph models.
On the Graph Out-of-Distribution (GOOD) benchmark~\citep{gui2022good}, \method{} ranks first or second in 13 of 20 settings against baselines including methods specifically designed for OOD generalization.
However, its performance drops substantially in several remaining settings, indicating that robust generalization under distribution shift remains an important direction for future work.

\subsection{Ablation Studies}
\label{subsec:architecture-prior-study}

We ablate the key design choices of \method{} to understand their individual contributions.
Since full pretraining requires approximately 16 GPU-days, we conduct these analyses at 10\% of the Stage~1 scale, using 5,000 updates over 320,000 synthetic graphs and evaluating on the same 51 datasets as in the main experiments.
Due to space constraints, we summarize the key findings here; full results, configurations, and detailed analyses are provided in \Cref{app:ablation-studies}.

\begin{wrapfigure}[15]{r}{0.32\textwidth}
    \vspace{-10mm}
    \centering
    \includegraphics[width=\linewidth]{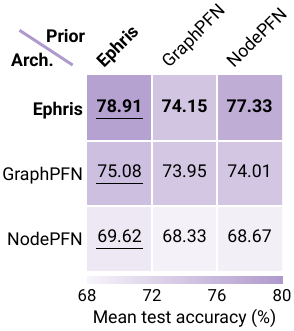}
    \vspace{-6mm}
    \caption{\textbf{Architecture–prior factorial study.} Bold: column best, underline: row best.}
    \label{fig:factorial_study}
\end{wrapfigure}

\mypar{Disentangling architecture and prior.}
\method{} introduces improvements along two axes: the architecture and the synthetic graph prior.
To disentangle their contributions, we conduct a \(3\times3\) factorial study crossing the architectures and priors of \method{}, GraphPFN~\citep{eremeev2026graphpfn} and NodePFN~\citep{choi2026learning}, training each architecture with every prior.
As shown in \Cref{fig:factorial_study}, our architecture achieves the highest mean accuracy under every prior, while our prior performs best across all architectures.
These consistent gains show that both contribute independently, rather than only through their specific combination.

\mypar{Component ablations.}
We ablate key components of the architecture and pretraining prior.
For the architecture, restricting per-feature gathering to context nodes, removing token updates after message passing, using static aggregation in MP$_\text{ICL}$, or removing neighborhood-size-aware scaling or global nodes all reduce both mean accuracy and win rate.
For the prior, removing propagation from the \ac{SCM} causes the largest degradation, reducing mean accuracy by roughly 20\%, suggesting that jointly modeling topology, features, and labels is critical.
Restricting the prior to any single relational dynamic also underperforms their mixture, while removing embedding-like post-processing also reduces performance.
Together, these results support the design choices of both the architecture and pretraining prior.
\section{Conclusion}

We introduced \method{}, a scalable graph in-context learner demonstrating that strong \ac{ICL} can be achieved through message passing over the observed graph without dense cross-node attention.
Pretrained on a synthetic prior spanning diverse graph structures, relational dynamics, and feature representations, \method{} transfers effectively to unseen graphs without per-dataset training or tuning.
Across 51 node-classification datasets, \method{} outperforms 15 extensively tuned \acp{GNN} and six recent \acp{GFM} across all aggregate metrics, including Elo, improvability, average rank, and accuracy, under both high- and low-label regimes.
It achieves this performance at roughly the cost of a single \ac{GNN} training run while running $13.1$--$16.1\times$ faster than GraphPFN, pushing the performance--runtime Pareto frontier forward.
Overall, the strong performance and fast inference of \method{} across diverse datasets suggest graph \ac{ICL} as a practical alternative to per-dataset \ac{GNN} training and tuning.

\mypar{Limitations and Future Work.}
Our ablation studies use reduced-budget proxy pretraining, so their conclusions may not fully transfer to full-scale training.
As shown in \Cref{subsec:beyond}, improving robustness to distribution shifts under non-IID grouped and temporal splits remains important for real-world applicability.
Our scope is also limited to node classification; extending scalable graph \ac{ICL} to link prediction and graph-level tasks is an important step toward general-purpose \acp{GFM}.

\newpage
\subsection*{AI use statement}
We used generative AI tools during code development and manuscript editing, primarily for code suggestions and improvements to clarity and grammar. The authors reviewed and tested all AI-assisted code and independently reviewed and revised all AI-assisted text. Generative AI was not used to generate experimental results or make research decisions. We take full responsibility for the final content of this work, including all AI-assisted content.

\subsection*{Ethics statement}
This work does not involve human subjects or the collection of personal data. Our model is pretrained entirely on synthetic graph tasks and evaluated on open node-classification benchmarks. We follow the licenses and intended research use of the datasets and methods used in our experiments, and encourage responsible use of the proposed model in downstream applications.

\subsection*{Reproducibility Statement}
We provide the model specifications, experimental settings, and evaluation procedures needed to reproduce our results throughout the appendix.
\Cref{app:architecture} describes the architecture and its constituent operations, while \Cref{app:prior} details graph sampling, relational dynamics, and feature post-processing in the synthetic pretraining prior.
\Cref{app:implementation-details} documents the model configuration, software and hardware environment, task-size sampling, two-stage curriculum, optimization settings, and inference procedure.

For evaluation, \Cref{app:evaluation-metrics,app:evaluation-details} specify the metrics, datasets, train/validation/test splits, baseline configurations, hyperparameter search spaces, and adaptation procedures for prior \acp{GFM}.
Complete aggregate results and supplementary comparisons are provided in \Cref{app:full-results}, and \Cref{app:ablation-studies} reports the configurations and results of our ablation studies.

The model checkpoint and code are available at \url{https://github.com/nums-ai/ephris}.
These artifacts allow researchers to evaluate \method{} directly without repeating the full pretraining process.
Together, the released artifacts and documented procedures support independent verification of our results and comparisons with future methods.



\newpage
\bibliography{iclr2027_conference}
\bibliographystyle{iclr2027_conference}

\newpage
\appendix
\crefname{appendix}{appendix}{appendices}
\Crefname{appendix}{Appendix}{Appendices}

\section{Evaluation Metrics}
\label[appendix]{app:evaluation-metrics}
We evaluate the high-label and low-label regimes
separately, each using 51 datasets and five splits
per dataset.
Default and tuned \acp{GNN} are treated as separate
participants, yielding 37 configurations.
Unless otherwise specified, the predictive score
$s_{m,d,r}$ for method $m$ on dataset $d$ and split
$r$ is AUROC for binary classification and accuracy
for multiclass classification.
Comparisons are computed within each dataset-split,
rather than from scores averaged across splits.

\mypar{Elo.}
For every pair of methods on the same dataset-split,
we assign an outcome of 1 for a win, 0.5 for an
exact tie, and 0 for a loss.
We fit a Bradley-Terry model to these outcomes,
weighting each split by $1/5$ so that datasets
contribute equally.
The resulting ratings are expressed on the Elo
scale and shifted so that default GCN has a rating
of 1000.
Higher ratings indicate stronger pairwise performance;
the reference rating only fixes the origin of the scale.

\mypar{Average rank.}
We rank all configurations within each dataset-split,
assigning tied methods the average of their occupied
ranks.
We then average ranks over the five splits within
each dataset and over the 51 datasets.
Lower values indicate better average standing.
In particular, we do not average predictive scores
across splits before computing ranks.

\mypar{Improvability~\citep{erickson2026tabarena}.}
Let $s_{*,d,r}=\max_j s_{j,d,r}$ denote the best
score among all configurations on dataset $d$
and split $r$.
The improvability of method $m$ is
\begin{equation}
I_{m,d,r}
=
100\,\frac{s_{*,d,r}-s_{m,d,r}}{1-s_{m,d,r}}.
\end{equation}
This measures the fraction of the remaining gap
to a perfect score that could be closed by matching
the best evaluated configuration.
For example, scores of 0.80 and 0.90 for the method
and the best configuration yield an improvability
of 50\%.
For AUROC, the denominator is the gap to perfect
AUROC, not a misclassification rate.
The best configuration has zero improvability;
cases with a denominator numerically close to zero
are also assigned zero.
We average split-level values within each dataset,
then average across datasets.
Lower mean improvability is better.

\mypar{Pairwise win rate.}
For methods $a$ and $b$, we compute
\begin{equation}
W_{a,b}
=
\frac{100}{D}
\sum_{d=1}^{D}
\frac{1}{S}
\sum_{r=1}^{S}
\left[
\mathbf{1}\{s_{a,d,r}>s_{b,d,r}\}
+
\frac{1}{2}\mathbf{1}\{s_{a,d,r}=s_{b,d,r}\}
\right],
\end{equation}
where $D=51$ and $S=5$.
Thus, ties count as half a win and each dataset
receives equal weight.
With five splits for every dataset, this equals
the tie-adjusted win rate over all 255 comparisons.
The overall leaderboard win rate averages pairwise
win rates over all other configurations.

\mypar{Average accuracy.}
We additionally report accuracy averaged across
splits and datasets, using accuracy for both binary
and multiclass tasks.
This metric therefore differs from Elo, average
rank, and improvability in its treatment of binary
classification, for which those metrics use AUROC.

\mypar{Runtime.}
Runtime measures the wall-clock cost of adapting a method to a new dataset.
It includes training and the full hyperparameter search for tuned supervised
methods, or the adaptation procedure for \acp{GFM}.
Let $t_{m,d}$ denote the runtime of method $m$ on dataset $d$.
We define the relative runtime overhead and its aggregate as
\begin{equation}
\rho_{m,d}
=
\frac{t_{m,d}}{\min_j t_{j,d}},
\qquad
\bar{\rho}_m
=
\exp\left(
\frac{1}{D}\sum_{d=1}^{D}\log\rho_{m,d}
\right).
\end{equation}
The geometric mean weights datasets equally and summarizes multiplicative
runtime differences.
Pairwise speedups are aggregated in the same way.

\section{Architecture Details}
\label[appendix]{app:architecture}
We describe the five stages of the architecture in \Cref{eq:compress-then-icl}.
The model uses feature-token width $d=128$, $K_F=128$ per-feature inducing tokens, and $K_N=4$ per-node inducing tokens, giving a compressed width of $D=K_Nd=512$.
Each message-passing stage maintains $K_G=8$ global nodes, which are updated throughout its successive blocks.

\paragraph{Tokenization.}
Numeric features are standardized, binary features are encoded as zero or one, and categorical features are ordinally encoded.
Each scalar value is then clipped to $[-8,8]$.
For a clipped value $\bar x$, we construct an 11-dimensional basis:
\begin{equation*}
\vb(\bar x)
=
[
\bar x,\;
\tanh(\bar x),\;
\operatorname{sign}(\bar x)\log(1+|\bar x|),\;
r_1(\bar x),\ldots,r_8(\bar x)
],
\quad
r_q(\bar x)=\exp(-(\bar x-c_q)^2/(2w^2)).
\end{equation*}
The eight radial basis functions have centers evenly spaced over $[-3,3]$ and bandwidth $w$.
A shared \ac{MLP} with gating maps this basis into a token of width $d$.

To provide information about the other features of the same node, the tokenizer also computes a node summary.
It takes the elementwise mean and root-mean-square of the basis vectors separately for numeric, binary, and categorical features, then concatenates these statistics.
A learned projection of the summary is added to each scalar representation through a gated residual connection, followed by RMS normalization.
This introduces node-level information without assigning positions to features.

Let $\vs_i$ denote this node summary.
The initial tokens are
\begin{equation*}
\vh_{if}^{(0)}
=
\operatorname{CellEncode}(\bar x_{if},\vs_i)
+
\ve_{\mathrm{type}}(t_f)
+
\mathbf1[i\in\mathcal V_{\mathrm{ctx}}]\ve_{\mathrm{feat}}(y_i).
\end{equation*}
Here, $\operatorname{CellEncode}$ includes the scalar projection and node-summary update, and $t_f$ identifies the feature type.
The context-label embedding is shared across all feature tokens of the same node.
Query nodes receive a zero label embedding.
The resulting tokens form the initial tensor $\tH^{(0)}$.

\paragraph{Graph-aware refinement.}
We apply $L_{\mathrm{ref}}=3$ refinement blocks.
Each block first exchanges distributional information across nodes for each feature, then incorporates structural information into each node's feature tokens.

\emph{Per-feature refinement.}
For each feature, $K_F$ learned inducing tokens attend to its tokens across all nodes to gather a summary.
The node tokens then attend to that summary to receive the broadcast:
\begin{equation*}
\mJ_f^{(\ell)}
=
\operatorname{Gather}_{F}^{(\ell)}(\mH_{:f}^{(\ell-1)}),
\qquad
\widetilde{\mH}_{:f}^{(\ell)}
=
\operatorname{Broadcast}_{F}^{(\ell)}(\mH_{:f}^{(\ell-1)},\mJ_f^{(\ell)}).
\end{equation*}
Here, $\mJ_f^{(\ell)}$ contains the summary tokens for feature $f$ in block $\ell$.
Both context and query nodes participate in gathering and broadcasting, while only context labels are available.
Features are processed independently with shared module parameters.
The updated tokens carry information about each feature's distribution across the graph.

\emph{Per-node refinement.}
We next gather across each node's features using $K_N$ learned inducing tokens.
Two successive $\operatorname{MP}_{\mathrm{ICL}}$ blocks exchange information among these node summaries before the updates are broadcast back to the feature tokens:
\begin{equation*}
\mI_i^{(\ell)}
=
\operatorname{Gather}_{N}^{(\ell)}(\widetilde{\mH}_{i:}^{(\ell)}),
\quad
\widehat{\tI}^{(\ell)}
=
\operatorname{MP}_{\mathrm{ICL},\ell}^{\circ2}(\tI^{(\ell)},\mA),
\quad
\mH_{i:}^{(\ell)}
=
\operatorname{Broadcast}_{N}^{(\ell)}(\widetilde{\mH}_{i:}^{(\ell)},\widehat{\mI}_i^{(\ell)}).
\end{equation*}
Here, $\mI_i^{(\ell)}$ contains node $i$'s summary tokens, and $\tI^{(\ell)}$ collects them across nodes.
The message-passing notation includes concatenating each node's tokens before propagation and splitting the updated vector back into tokens afterward.
Each two-block stage initializes its own global nodes and updates them alongside the original nodes through both blocks.
Alternating per-feature and per-node refinement allows distributional and structural information to guide the representations used for compression.

\paragraph{Compression.}
A per-node gather produces $K_N$ summary tokens, which are concatenated into a representation of width $D$, independent of the feature count.
Before graph \ac{ICL}, context-label embeddings are added to these representations, followed by layer normalization:
\begin{equation*}
\vz_i^{(0)}
=
\operatorname{LN}(
\operatorname{Concat}(\mI_i^{\mathrm{comp}})
+
\mathbf1[i\in\mathcal V_{\mathrm{ctx}}]\ve_{\mathrm{ICL}}(y_i)
).
\end{equation*}
Here, $\mI_i^{\mathrm{comp}}$ contains the tokens produced by the compression gather.
The ICL label embeddings have width $D$ and use a separate table from the tokenizer's width-$d$ embeddings.
Both tables are trainable and initialized with orthogonal class embeddings.
The resulting matrix $\mZ^{(0)}$ is the label-conditioned input to graph \ac{ICL}.

\paragraph{Graph in-context learning.}
The node representations pass through an independent stage of $L_{\mathrm{ICL}}=10$ successive message-passing blocks:
\begin{equation*}
\mZ^{(L_{\mathrm{ICL}})}
=
\operatorname{MP}_{\mathrm{ICL}}^{\circ L_{\mathrm{ICL}}}(\mZ^{(0)},\mA).
\end{equation*}
This stage initializes its own eight global nodes and updates them throughout the ten blocks.
Only the representations of the original nodes are retained at the output.
Together with the six blocks used during refinement, the model contains 16 message-passing blocks.

\paragraph{Prediction head.}
A shared head applies layer normalization and a linear projection to produce ten logits per node:
\begin{equation*}
\vo_i
=
\operatorname{Linear}_{D\rightarrow10}(
\operatorname{LN}(\vz_i^{(L_{\mathrm{ICL}})})
).
\end{equation*}
For $C\leq10$, softmax is applied to the first $C$ logits.
For $C>10$, ECOC decomposes the task into classification problems with at most ten classes, each evaluated using the same pretrained model.
Each original class is scored by averaging the log probabilities assigned to its code entries.
Softmax then converts these scores into probabilities over the original classes, with the query-node predictions used for evaluation.

\section{Synthetic Graph Prior Details}
\label[appendix]{app:prior}
This section details the three components of the synthetic prior in \Cref{sec:structure-conditioned-prior}: graph sampling, relational dynamics, and feature post-processing.
Together, they vary graph topology, how topology influences features and labels, and how generated features are represented.
Task-size sampling and the training setup are described in \Cref{app:implementation-details}.

\subsection{Graph Sampling}
\label{app:prior-graph-sampling}

Each graph combines group-based, degree-heterogeneous, and ordering-based connectivity.
We sample their relative contributions once per graph and select a rule for each proposed edge.
Topology is generated before features and labels, so neither determines which nodes are connected.

\paragraph{Node sampling weights.}
All rules sample source nodes from a shared distribution:
\begin{equation*}
p_i=\frac{\exp(\sigma g_i)}{\sum_{j=1}^{N}\exp(\sigma g_j)},
\qquad
g_i\sim\mathcal N(0,1),
\qquad
\sigma\sim\mathcal U(0.3,1.6).
\end{equation*}
Larger values of $\sigma$ concentrate source probability on fewer nodes, increasing out-degree heterogeneity.
The rules differ in how they choose a destination.

\paragraph{Group-based connectivity.}
We sample three group structures.
First, nodes are randomly partitioned into 2 to 8 approximately balanced coarse groups.
Each coarse group is divided into 1 to 4 approximately balanced fine groups.
A separate partition assigns nodes to 2 to 4 additional groups.

The coarse and fine rules sample a destination uniformly from the source node's group.
For the additional partition, a fair coin sampled once per graph selects within-group or cross-group connectivity.
In the cross-group case, each proposal selects another group uniformly, then samples a destination uniformly within it.
These rules produce both within-group concentration and cross-group connections.

\paragraph{Degree-heterogeneous connectivity.}
The destination is sampled independently from the same distribution as the source.
An ordered pair $(i,j)$ is therefore proposed with probability $p_i p_j$.
Nodes with larger weights are more likely to appear at either endpoint, producing heterogeneous degrees and hubs without prescribing an exact degree sequence.

\paragraph{Ordering-based connectivity.}
We sample a random node ordering and connect each sampled source to the node a fixed number of positions ahead, wrapping around at the end.
The offset is shared across the graph and sampled uniformly from 1 to $\min(N,32)-1$.
Writing the ordering as $\pi$, the destination of source $i$ is
\begin{equation*}
j=\pi((\pi^{-1}(i)+\Delta)\bmod N).
\end{equation*}
Here, positions are indexed from zero, and $\Delta$ is the sampled offset.
This introduces a common transition pattern among the proposed edges.
Because only sampled edges are retained and mixed with other rules, the resulting graph need not form a complete cycle.

\paragraph{Mixture weights.}
The implementation uses six entries: three group-based rules, one ordering-based rule, and two entries with the same degree-heterogeneous rule.
For each entry $k$, we independently sample $u_k\sim\mathcal U(0,1)$ and normalize the clipped weights:
\begin{equation*}
w_k=
\frac{\max(u_k,0.05)}
{\sum_{r=1}^{6}\max(u_r,0.05)}.
\end{equation*}
Each edge proposal samples a source from $p$, an entry from $w$, and a destination using that entry's rule.
The two degree-heterogeneous entries contribute jointly to the same connectivity pattern.

\paragraph{Edge construction.}
Given target average out-degree $d$, we draw $M=\min(N(N-1),\operatorname{round}(Nd))$ proposals with replacement, then remove self-edges and duplicates.
To vary reciprocity, we sample $\rho\sim\mathcal U(0,1)$ and independently add the reverse of each remaining edge with probability $\rho$.
After deduplication, any excess over $M$ is removed by uniformly selecting $M$ edges.
The realized graph can therefore contain fewer than $M$ edges, and isolated nodes are allowed.
This directed graph is used to generate features and labels.
The model's subsequent symmetrization and self-loop insertion are separate from this sampling procedure.

\subsection{Relational Dynamics}
\label{app:prior-relational-dynamics}

Relational dynamics determine how the sampled graph influences the intermediate variables used to generate features and labels.
We detail the propagation operator in \Cref{eq:s2cm-latent-generation}, using the SCM construction described in \Cref{eq:scm-generation}.

\paragraph{Selecting propagation settings.}
Let $L_{\mathrm{SCM}}$ be the number of retained SCM variables needed to generate the outputs.
We sample $q_{\mathrm{prop}}\sim\mathcal U(0,1)$ and select $\operatorname{round}(q_{\mathrm{prop}}L_{\mathrm{SCM}})$ variables uniformly without replacement.
Each selected variable independently samples a propagation rule, aggregation operator, and direction.
Propagation is the identity for unselected variables.

The direction is chosen uniformly from incoming, outgoing, and bidirected propagation.
These use the original edges, reversed edges, or their deduplicated union, respectively.
The remaining settings are summarized in \Cref{tab:prior-propagation-settings}.
All sampled settings remain fixed throughout propagation for a given variable.

\begin{table}[t]
\centering
\small
\caption{\textbf{Relational-dynamics settings.}
Discrete choices and continuous intervals are sampled uniformly.
Rule-specific parameters apply only to the indicated operations.}
\label{tab:prior-propagation-settings}
\begin{tabular}{@{}ll@{}}
\toprule
Setting & Choices or range \\
\midrule
Rule & Diffusion, cascade, degree-dependent mixing \\
Aggregation & Mean, max, min, softmax \\
Direction & Incoming, outgoing, bidirected \\
Mixing coefficient $\alpha$ & $[0.15,0.85]$ \\
Steps $T$ (diffusion and cascade) & $\{2,3,4,5,6\}$ \\
Seed fraction $s$ (cascade) & $[0.08,0.28]$ \\
Threshold $\theta$ (cascade) & $[0.20,0.55]$ \\
Temperature $\tau$ (softmax) & $[0.25,2.0]$ \\
\bottomrule
\end{tabular}
\end{table}

\paragraph{Neighborhood aggregation.}
For a selected SCM variable $a$, let $\vu_i^{(0)}$ denote its value at node $i$ before propagation, corresponding to row $i$ of $\widetilde{\mU}^{(a)}$.
We omit the variable index below and use $t$ to index propagation steps.
Let $\mathcal N(i)$ contain the neighbors that send information to node $i$ under the sampled direction.
The neighborhood aggregate is
\begin{equation*}
\vm_i^{(t)}
=
\operatorname{Agg}(\{\vu_j^{(t)}:j\in\mathcal N(i)\}).
\end{equation*}
Mean, max, and min operate coordinatewise.
Softmax aggregation weights neighbors by cosine similarity to the receiving node:
\begin{equation*}
\vm_i^{(t)}
=
\sum_{j\in\mathcal N(i)}
\frac{\exp(\operatorname{cos}(\vu_i^{(t)},\vu_j^{(t)})/\tau)}
{\sum_{k\in\mathcal N(i)}
\exp(\operatorname{cos}(\vu_i^{(t)},\vu_k^{(t)})/\tau)}
\vu_j^{(t)}.
\end{equation*}
For all rules, nodes without eligible sending neighbors remain unchanged.

\paragraph{Diffusion.}
Diffusion repeatedly mixes each node's current value with its neighborhood aggregate:
\begin{equation*}
\vu_i^{(t+1)}
=
(1-\alpha)\vu_i^{(t)}+\alpha\vm_i^{(t)},
\qquad t=0,\ldots,T-1.
\end{equation*}
Repeated updates allow neighbor influence to accumulate and propagate beyond immediate neighbors.

\paragraph{Cascade.}
Cascade starts from a small set of active nodes and updates other nodes only when their agreement with active neighbors reaches a threshold.
The initial active set contains $\max(1,\operatorname{round}(sN))$ nodes sampled uniformly without replacement.

At step $t$, let $\mathcal A_t$ be the active set and let $\vm_{i,\mathrm{act}}^{(t)}$ aggregate only neighbors in $\mathcal N(i)\cap\mathcal A_t$.
The newly activated nodes are
\begin{equation*}
\mathcal B_t
=
\{i\notin\mathcal A_t:
\mathcal N(i)\cap\mathcal A_t\ne\varnothing,\;
(1+\operatorname{cos}(\vu_i^{(t)},\vm_{i,\mathrm{act}}^{(t)}))/2
\geq\theta\}.
\end{equation*}
Only these nodes are updated:
\begin{equation*}
\vu_i^{(t+1)}
=
\begin{cases}
(1-\alpha)\vu_i^{(t)}+\alpha\vm_{i,\mathrm{act}}^{(t)}, & i\in\mathcal B_t,\\
\vu_i^{(t)}, & \text{otherwise},
\end{cases}
\qquad
\mathcal A_{t+1}=\mathcal A_t\cup\mathcal B_t.
\end{equation*}
The process runs for $T$ steps.
Once activated, a node retains its updated value and can influence further activations.
Each non-seed node is therefore updated at most once, making propagation conditional rather than continuous.

\paragraph{Degree-dependent mixing.}
This rule applies a single update whose strength depends on node degree and neighbor agreement.
We rank nodes by receiving degree under the sampled direction and assign buckets $b_i\in\{0,1,2\}$ to the lower, middle, and upper thirds.
Using the aggregate of the initial values, we compute
\begin{equation*}
\delta_i=(1-\operatorname{cos}(\vu_i^{(0)},\vm_i^{(0)}))/2,
\qquad
\gamma_i=1+\tfrac12 b_i+\delta_i,
\qquad
\vu_i^{(1)}
=
\vu_i^{(0)}+\alpha\gamma_i(\vm_i^{(0)}-\vu_i^{(0)}).
\end{equation*}
Higher degree and greater disagreement both increase the update strength.
The effective coefficient $\alpha\gamma_i$ can exceed one, so the update may extrapolate beyond the neighborhood aggregate.
This produces node-specific responses to the same propagation mechanism.

\subsection{Post-processing}
\label{app:prior-post-processing}

We choose uniformly among tabular, embedding, and hybrid representations.
These modes vary how feature information is distributed across dimensions while leaving labels unchanged.
The mode is selected before feature generation to determine the required source width.
Let $F$ denote the final feature count.

\paragraph{Tabular representation.}
The generator produces $F$ source columns, which are retained without additional mixing.
Numerical and categorical attributes therefore remain individually accessible.

\paragraph{Embedding representation.}
We transform a compact source matrix into an embedding of width $F_{\mathrm{emb}}$, where $F_{\mathrm{emb}}=F$ in this mode.
The source width $q$ is sampled log-uniformly between $\min(4,F_{\mathrm{emb}})$ and $\min(32,F_{\mathrm{emb}})$ and rounded to the nearest integer.

The source columns are first standardized.
With probability $1/2$, we apply a random linear map followed by GELU and standardize the result again:
\begin{equation*}
\mV=\operatorname{Std}(\mS),
\qquad
\mV\leftarrow\operatorname{Std}(\operatorname{GELU}(\mV\mW+\boldsymbol1\vb^\top)).
\end{equation*}
Here, $W_{ab}\sim\mathcal N(0,1/q)$ and $b_a\sim\mathcal N(0,1)$.
The operator $\operatorname{Std}$ centers each column and divides by its population standard deviation, bounded below by $10^{-6}$.

A random projection then distributes the source information across the embedding dimensions:
\begin{equation*}
\mB_0=\operatorname{Std}(\mV\operatorname{diag}(\vs)\mQ^\top),
\qquad
s_k=\frac{k^{-\eta}}{\sqrt{q^{-1}\sum_{r=1}^{q}r^{-2\eta}}}.
\end{equation*}
The columns of $\mQ$ form an orthonormal basis obtained by QR decomposition of a Gaussian matrix.
The sampled exponent $\eta$ controls how strongly the source directions contribute.

We add independent low-rank nuisance variation and Gaussian noise:
\begin{equation*}
\mB_1=\mB_0+\lambda\operatorname{Std}(\mR\mQ_\nu^\top)+\epsilon\mG.
\end{equation*}
Here, $\mR$ has $\nu$ independent standard Gaussian columns, $\mQ_\nu$ is an independent orthonormal projection, and $\mG$ contains independent standard Gaussian entries.
The nuisance term is omitted when $\nu=0$.
Finally, with equal probability, we either normalize each row to norm $\sqrt{F_{\mathrm{emb}}}$ or multiply it by an independent lognormal factor.
There is no further column standardization.

\begin{table}[t]
\centering
\small
\caption{\textbf{Embedding variation.}
Sampling distributions for projection, noise, and row scaling.
Lognormal scaling multiplies row $i$ by $\exp(\sigma_r z_i)$, with $z_i\sim\mathcal N(0,1)$.}
\label{tab:prior-embedding-settings}
\begin{tabular}{@{}ll@{}}
\toprule
Parameter & Distribution \\
\midrule
Spectral decay $\eta$ & $\mathcal U(0.5,2.0)$ \\
Nuisance rank $\nu$ & Uniform over $\{0,\ldots,\min(8,q)\}$ \\
Nuisance amplitude $\lambda$ & $\mathcal U(0,0.5)$ \\
Noise scale $\epsilon$ & $\operatorname{LogUniform}(0.01,0.35)$ \\
Lognormal scale $\sigma_r$ & $\mathcal U(0.05,0.5)$ \\
\bottomrule
\end{tabular}
\end{table}

\paragraph{Hybrid representation.}
We combine an embedding with untransformed source columns.
The number of retained columns $m$ is sampled uniformly from
\begin{equation*}
\{1,\ldots,\min(16,F-1,\max(1,\lfloor F/4\rfloor))\}.
\end{equation*}
We set $F_{\mathrm{emb}}=F-m$ and sample the source width $q$ as above.
Of the $q+m$ generated columns, the first $q$ are transformed into an embedding and the remaining $m$ are retained.
Concatenating the two parts and randomly permuting their columns produces the final $F$-dimensional representation.

\section{Implementation Details}
\label[appendix]{app:implementation-details}
This section describes the experimental environment and implementation details of \method{}, covering its architecture configuration, pretraining setup, and inference protocol.

\subsection{Experiment Environment}
\label[appendix]{app:environment}
Experiments were conducted on NVIDIA H200 GPUs.
The server was equipped with two Intel Xeon Platinum 8462Y+ CPUs (64 physical cores in total) and approximately 2 TiB of RAM, running Ubuntu 22.04.5 LTS.
Pretraining used Python 3.11.9 and PyTorch 2.11.0 with CUDA 13.0.

\subsection{Architecture Setup}
\label{app:architecture-setup}

We instantiate the architecture described in \Cref{sec:architecture} using the configuration in \Cref{tab:model-configuration}.
Each scalar node-feature value is embedded into a 128-dimensional token.
Per-feature refinement uses 128 inducing tokens with eight attention heads to model each feature across nodes.
Per-node refinement then applies three refinement blocks, each gathering feature tokens into four node-inducing tokens and processing them with two $\mathrm{MP}_{\mathrm{ICL}}$ layers.
After the final per-feature refinement and gather operation, the four tokens of each node are concatenated into a 512-dimensional representation.
Ten additional $\mathrm{MP}_{\mathrm{ICL}}$ layers perform graph \ac{ICL}, followed by a prediction head supporting up to ten classes.

\begin{table}[h]
\centering
\small
\caption{
\textbf{Architecture configuration of \method{}.}
}
\label{tab:model-configuration}
\begin{tabular}{@{}llr@{}}
\toprule
Component & Setting & Value \\
\midrule
Tokenization
& Feature-token dimension & 128 \\
\midrule
Per-feature refinement
& Inducing tokens & 128 \\
& Attention heads & 8 \\
\midrule
Per-node refinement
& Refinement blocks & 3 \\
& Node-inducing tokens per node & 4 \\
& $\mathrm{MP}_{\mathrm{ICL}}$ layers per block & 2 \\
& Concatenated node dimension & 512 \\
\midrule
$\mathrm{MP}_{\mathrm{ICL}}$
& Global nodes & 8 \\
& Attention heads & 8 \\
& Attention-scaling MLP hidden dimension & 64 \\
\midrule
Graph \ac{ICL}
& $\mathrm{MP}_{\mathrm{ICL}}$ layers & 10 \\
\midrule
Prediction
& Maximum output classes & 10 \\
\midrule
General
& FFN expansion factor & 4 \\
& Dropout & 0 \\
\bottomrule
\end{tabular}
\end{table}

\subsection{Pretraining Setup}
\label{app:pretraining-setup}

\paragraph{Two-stage curriculum.}
We pretrain \method{} on the synthetic node-classification tasks
described in \Cref{sec:structure-conditioned-prior}.
Stage~1 uses moderately sized graphs with up to 128 features.
Stage~2 broadens the distribution to include smaller and larger
graphs, with sampling bounds of 128 to 16,384 nodes and
2 to 1,024 features.
Both stages share the SCM, relational-dynamics, and
feature-observation settings.

Stage~2 starts from the final Stage~1 model weights,
with freshly initialized optimizer and scheduler states.
Each update uses 64 newly generated tasks.
The full curriculum comprises 50,000 Stage~1 updates and
10,000 Stage~2 updates, totaling 3.84M training tasks.
\Cref{tab:pretraining-configuration} summarizes the settings.

\begin{table*}[t]
\centering
\small
\caption{
\textbf{Two-stage pretraining configuration.}
Ranges specify sampling bounds before integer conversion.
Stage~2 restricts the feature-count and target-degree
supports using memory budgets.
}
\label{tab:pretraining-configuration}
\begin{tabular}{@{}llcc@{}}
\toprule
Component & Setting & Stage 1 & Stage 2 \\
\midrule
\multirow{6}{*}{Task distribution}
& Updates & 50,000 & 10,000 \\
& Graphs per update & 64 & 64 \\
& Node-count bounds & 512 to 1,024 & 128 to 16,384 \\
& Feature-count bounds & 2 to 128 & 2 to 1,024 \\
& Requested class count & 2 to 10 & 2 to 10 \\
& Context fraction & Uniform on $[0.3,0.9]$ & $0.8$ \\
\midrule
\multirow{3}{*}{Graph sampling}
& Target average out-degree bounds & 1.5 to 500 & 1.5 to 256 \\
& Node-feature budget $B_{NF}$ & None & $2^{21}$ \\
& Directed-edge budget $B_E$ & None & $2^{19}$ \\
\midrule
\multirow{4}{*}{Optimization}
& Muon peak learning rate & $8\times10^{-4}$ & $4\times10^{-5}$ \\
& AdamW peak learning rate & $8\times10^{-4}$ & $10^{-5}$ \\
& LR decay endpoint (both groups) & $4\times10^{-5}$ & $10^{-5}$ \\
& Warmup updates & 1,000 & 500 \\
\bottomrule
\end{tabular}
\end{table*}

\paragraph{Task generation.}
We sample graph sizes, feature counts, class counts, and target average out-degrees within the stage-specific ranges in \Cref{tab:pretraining-configuration}.
Stage~2 aims to expose the model to both wide feature sets and large graphs despite limited memory: wide features remain available on smaller graphs, while larger graphs are paired with fewer features.
We implement this trade-off through a node-feature budget $B_{NF}=2^{21}$.
After sampling the node count $N$, we sample the feature count up to the smaller of 1,024 and $B_{NF}/N$, rounded down.
Similarly, a directed-edge budget $B_E=2^{19}$ restricts the target average out-degree to at most the smaller of 256 and $B_E/N$, allowing denser small graphs while keeping larger graphs sparse.

\paragraph{Training context.}
For each generated task, we select $\lfloor rN\rfloor$ context nodes uniformly without replacement and reveal their labels to the model.
Stage~1 samples the context fraction $r$ uniformly from $[0.3,0.9]$, while Stage~2 fixes $r=0.8$ to stabilize training during adaptation to larger tasks.
A high context fraction reduces the loss fluctuations associated with sparse supervision, helping preserve the capabilities learned in Stage~1.
The model receives all node features and graph edges together with the context labels, and the classification loss is computed on the remaining query nodes.
This trains the model to infer query labels from the observed context within each graph.

\subsection{Optimization}
\label{app:optimization}
We optimize eligible matrix parameters with Muon~\citep{jordan2024muon} and the remaining parameters with AdamW~\citep{loshchilov2017decoupled}.
Muon uses momentum $0.95$, Nesterov updates, and five Newton-Schulz iterations to approximately orthogonalize each update.
For an $a\times b$ parameter matrix, the resulting update is scaled by $0.2\sqrt{\max(a,b)}$ before applying the scheduled learning rate.
The peak learning rates reported in \Cref{tab:pretraining-configuration} are specified before this matrix-dependent scaling.
AdamW uses $(\beta_1,\beta_2)=(0.9,0.95)$ and $\epsilon=10^{-8}$.

Each stage begins with linear warmup from zero to the respective peak learning rates, followed by cosine decay toward the same absolute endpoint for both parameter groups.
In Stage~2, the AdamW peak learning rate equals this endpoint, so it remains constant at $10^{-5}$ after warmup.
Over the same period, the Muon learning rate decays from its peak of $4\times10^{-5}$ to $10^{-5}$.

\subsection{Inference}
\label{app:inference}

We use the final Stage~2 checkpoint for all datasets without dataset-specific gradient updates.
As in pretraining, the model receives all node features and graph edges, while labels are revealed only for context nodes.
Feature preprocessing is fitted on the available node features.

We do not use test-time permutation ensembling, which combines predictions across feature and class permutations in many recent \acp{TFM}~\citep{qu2026tabiclv, grinsztajn2026tabpfn, cho2026causilotechnicalreport}.
Our architecture is invariant to feature permutations, so changing feature order does not affect predictions.
Class-label permutations, however, can change predictions even after the outputs are mapped back to the original labels.
We use a single deterministic feature and class permutation and leave the potential benefits of class-permutation ensembling to future work.

For datasets with more than ten classes, we apply the ECOC procedure described in \Cref{sec:architecture} with fixed redundancy four.
A dataset with $C$ classes uses $4\lceil\log_{10}C\rceil$ code columns, each defining a classification problem with at most ten outputs.
All code problems are evaluated using the same pretrained model weights and preprocessed features.
To recover predictions over the original classes, we average the log probabilities assigned to each class's code symbols across columns, then normalize the resulting scores over the original classes.

\section{Evaluation Details}
\label[appendix]{app:evaluation-details}
This section provides additional details of our evaluation.
We describe the datasets and baselines, the training and hyperparameter-tuning protocols for supervised \acp{GNN}, and the adaptation procedures of \acp{GFM}, including their design and our implementation.

\subsection{Datasets}
\label{app:datasets}

Our 51 datasets are selected to cover a broad range of node-classification settings rather than a narrow family of commonly used benchmarks.
They span six application domains and vary substantially in scale, dimensionality, label space, connectivity, and homophily, ranging from 183 to 568,795 nodes, 12 to 8,710 input features, and 2 to 70 classes, with average degrees from 2.30 to 88.28 and adjusted label homophily from $-0.30$ to $0.94$.
This diversity allows us to test whether conclusions persist across substantially different graph and task characteristics.
The full dataset list, together with domains, key statistics, and sources, is provided in \Cref{tab:dataset-catalog}.

\subsection{GNN Training and Hyperparameter Tuning}
\label{app:gnn-tuning}

We extensively tune all 15 supervised \acp{GNN} to establish strong per-dataset baselines.
When the original work specifies a hyperparameter search space, we follow it; when only selected configurations are reported, we construct a search space that includes the reported choices and varies the corresponding hyperparameters.
For conventional \acp{GNN}, we additionally tune choices often omitted from default implementations but known to substantially affect node-classification performance, including residual connections, layer normalization, pre-transformation, and dropout where applicable~\citep{luo2024classic}.
Our goal is to avoid comparisons against weak default configurations and provide a stringent test of pretrained methods.

For each method and dataset split, we sample 200 configurations from the resulting search space and select the final configuration solely by validation performance.
The complete baseline list, shared training settings, method-specific defaults, and search spaces are reported in \Cref{tab:search-spaces}.

\subsection{GFM Adaptation Protocols}
\label{app:gfm-adaptation}
Applying a pretrained \ac{GFM} to a new graph does not always mean inference alone.
Depending on the method, the original implementation may fit a new predictor, select among pretrained representations, or choose dataset-specific preprocessing and inference settings.
In our evaluation, we preserve these target-time procedures while keeping all pretrained backbones fixed.
We count any computation and validation required by these procedures as part of adaptation.

\mypar{GraphAny.}
GraphAny~\citep{zhao2025fully} combines several LinearGNN prediction channels using a pretrained inductive-attention model.
On each new graph, its original inference pipeline first recomputes the analytical LinearGNN predictors from the available training labels and then passes their predictions to the fixed pretrained model.
We follow the same procedure: the analytical predictors are recomputed for every dataset, while the pretrained model is never updated.
Nodes are processed in batches of 100,000.
No dataset-specific hyperparameter selection is performed.

\mypar{GVT.}
GVT~\citep{lee2026view} transfers a recurrent graph encoder whose depth can be chosen separately for each target dataset.
The original implementation freezes the encoder, produces node representations at depths 1--8, trains a lightweight predictor on each representation, and selects the depth using validation performance.
We reproduce this procedure directly.
For every depth, we train a two-layer prediction head with hidden width 128, ReLU activation, and no dropout using Adam with learning rate $0.005$ and zero weight decay.
Training runs for up to 2,500 epochs with patience 200, and we restore the checkpoint with the lowest validation CE or BCE, breaking ties by training loss.
Thus, both predictor fitting and depth selection contribute to GVT's per-dataset adaptation.

\mypar{Node4All.}
Node4All~\citep{lee2026node4all} transfers a fixed pretrained encoder and uses its node representations as input to a downstream predictor.
Following its original evaluation protocol, we keep the encoder frozen and fit a two-layer prediction head of hidden width 512 separately on each target dataset.
The full graph is encoded once, with features processed in chunks under a budget of 1,000,000 feature elements when necessary.
The head uses Adam with learning rate $0.001$ and zero weight decay for up to 2,500 epochs with patience 200, and validation loss selects the checkpoint.
Unlike GVT, there is no encoder-depth search; target-specific adaptation consists of fitting and selecting the prediction head.

\mypar{G2T-FM.}
G2T-FM~\citep{eremeev2025turning} augments node features with graph-derived information and presents the resulting representation to a pretrained tabular foundation model.
Its released code supports target-specific optimization, but we use the inference-only configuration in our evaluation so that the pretrained model remains fixed.
All training nodes are provided as labeled context, from which the model directly predicts the remaining nodes.
We use a single canonical configuration with inference batch size~1, seqlenred=1024, and PEARL batch size~8.
Because this configuration is fixed across datasets, no validation-based model or hyperparameter selection is required.

\mypar{GraphPFN.}
GraphPFN~\citep{eremeev2026graphpfn} supports graph-aware \ac{ICL}: in its inference mode, training nodes are supplied as labeled context and the pretrained model predicts the remaining nodes without gradient updates.
The released interface additionally exposes feature preprocessing, most notably optional PCA for high-dimensional inputs.
We therefore keep the model fixed and evaluate the two PCA configurations considered in our evaluation, selecting between them for each target dataset.
Each configuration uses the full graph, all training nodes as context, and a requested ensemble size of 10, which the implementation increases automatically when required.
Thus, GraphPFN requires no model fitting in our evaluation, but its preprocessing choice is dataset-specific.

\mypar{NodePFN.}
NodePFN~\citep{choi2026learning} also performs graph \ac{ICL} without updating its pretrained parameters.
Its original evaluation pipeline, however, specifies preprocessing and inference settings separately for each dataset, including dimensionality reduction, the retained dimensionality, and feature smoothing.
Applying NodePFN to a new dataset therefore requires choosing these settings even though the model itself is frozen.
To capture this adaptation cost, we construct a search space that subsumes the dataset-specific configurations reported by the original implementation and sample up to 200 configurations from it.
The search varies ensemble size over ${1,4,8,16,32}$, dimensionality reduction between none and truncated SVD with ${10,15,20,25,50}$ components, and the exposed smoothing choices.
We select the configuration using validation performance, then evaluate with all training nodes as context and the remaining nodes as queries in batches of~32.

\mypar{Use of validation data.}
These procedures differ importantly in how much target-specific selection they require.
GVT uses validation data to select both predictor checkpoints and recurrent depth; Node4All uses it for predictor checkpoint selection; and GraphPFN and NodePFN use it to resolve dataset-specific preprocessing or inference choices in our evaluation.
GraphAny and our inference-only G2T-FM configuration require no such selection.
Accordingly, a frozen backbone does not necessarily imply validation-free adaptation: several \acp{GFM} still use validation data indirectly to choose how the pretrained model is applied to each new dataset.
In contrast, \method{} uses a single fixed inference procedure across all datasets: it receives only the training labels as context, performs no parameter updates or dataset-specific configuration search, and never uses validation labels.
\begin{table*}[!p]
\centering
\caption{\textbf{Overview of the 51 node-classification datasets.}
We report each dataset's application domain and key characteristics:
number of nodes ($N$), features ($F$), classes ($C$), average degree ($\bar d$),
and adjusted label homophily ($h_{\mathrm{adj}}$).
Citations refer to the original dataset or benchmark sources.}
\vspace{8pt}
\label{tab:dataset-catalog}
\small
\setlength{\tabcolsep}{2.2pt}
\renewcommand{\arraystretch}{0.88}

\begin{tabular*}{\textwidth}{@{\extracolsep{\fill}}llrrrrrl@{}}
\toprule
\textbf{Dataset} &
\textbf{Domain} &
$N$ &
$F$ &
$C$ &
$\bar d$ &
$h_{\mathrm{adj}}$ &
\textbf{Citation} \\
\midrule
actor              & Web       & 7,600   & 932   & 5  & 7.02  & 0.003  & \citep{pei2020geom} \\
amazon\_computer   & Commerce  & 13,752  & 767   & 10 & 35.76 & 0.682  & \citep{shchur2018pitfalls} \\
amazon\_photo      & Commerce  & 7,650   & 745   & 8  & 31.13 & 0.785  & \citep{shchur2018pitfalls} \\
amazon\_ratings    & Commerce  & 24,492  & 300   & 5  & 7.60  & 0.140  & \citep{platonov2023critical} \\
amherst41          & Social    & 2,235   & 1,193 & 2  & 81.39 & 0.060  & \citep{lim2021large} \\
artnet-exp         & Social    & 50,405  & 75    & 2  & 11.12 & 0.155  & \citep{bazhenov2026graphland} \\
blogcatalog        & Social    & 5,196   & 8,189 & 6  & 66.11 & 0.272  & \citep{li2015unsupervised} \\
chameleon          & Web       & 890     & 2,325 & 5  & 19.90 & 0.030  & \citep{platonov2023critical} \\
citation\_citeseer & Scholarly & 4,230   & 602   & 6  & 2.52  & 0.938  & \citep{bojchevski2017deep} \\
citeseer           & Scholarly & 3,327   & 3,703 & 6  & 2.74  & 0.671  & \citep{yang2016revisiting} \\
city-reviews       & Commerce  & 148,801 & 204   & 2  & 15.66 & 0.591  & \citep{bazhenov2026graphland} \\
coauthor\_cs       & Scholarly & 18,333  & 6,805 & 15 & 8.93  & 0.785  & \citep{shchur2018pitfalls} \\
coauthor\_physics  & Scholarly & 34,493  & 8,415 & 5  & 14.38 & 0.872  & \citep{shchur2018pitfalls} \\
cora               & Scholarly & 2,708   & 1,433 & 7  & 3.90  & 0.771  & \citep{yang2016revisiting} \\
cora\_ml           & Scholarly & 2,995   & 2,879 & 7  & 5.45  & 0.749  & \citep{bojchevski2017deep} \\
cornell            & Web       & 183     & 1,703 & 5  & 3.03  & -0.220 & \citep{pei2020geom} \\
cornell5           & Social    & 18,660  & 4,735 & 2  & 84.76 & 0.091  & \citep{lim2021large} \\
dblp               & Scholarly & 17,716  & 1,639 & 4  & 5.97  & 0.679  & \citep{bojchevski2017deep} \\
deezer             & Social    & 28,281  & 128   & 2  & 6.56  & 0.030  & \citep{rozemberczki2020characteristic} \\
elliptic\_bitcoin  & Finance   & 203,769 & 165   & 2  & 2.30  & 0.516  & \citep{weber2019anti} \\
facebook\_large    & Social    & 22,470  & 4,714 & 4  & 15.20 & 0.821  & \citep{rozemberczki2021multi} \\
flickr             & Social    & 89,250  & 500   & 7  & 10.08 & 0.094  & \citep{zeng2019graphsaint} \\
full\_cora         & Scholarly & 19,793  & 8,710 & 70 & 6.41  & 0.556  & \citep{bojchevski2017deep} \\
genius             & Social    & 421,961 & 12    & 2  & 4.37  & -0.053 & \citep{lim2021large} \\
johnshopkins55     & Social    & 5,180   & 2,406 & 2  & 72.04 & 0.097  & \citep{lim2021large} \\
la                 & Transport & 240,587 & 37    & 10 & 2.84  & 0.618  & \citep{liang2026towards} \\
lastfm\_asia       & Social    & 7,624   & 7,842 & 18 & 7.29  & 0.856  & \citep{rozemberczki2020characteristic} \\
london             & Transport & 568,795 & 37    & 10 & 2.66  & 0.626  & \citep{liang2026towards} \\
ogbn\_arxiv        & Scholarly & 169,343 & 128   & 40 & 13.67 & 0.588  & \citep{hu2020open} \\
paris              & Transport & 114,127 & 37    & 10 & 3.20  & 0.570  & \citep{liang2026towards} \\
penn94             & Social    & 41,554  & 4,814 & 2  & 65.56 & 0.021  & \citep{lim2021large} \\
pubmed             & Scholarly & 19,717  & 500   & 3  & 4.50  & 0.686  & \citep{yang2016revisiting} \\
reed98             & Social    & 962     & 745   & 2  & 39.11 & 0.022  & \citep{lim2021large} \\
shanghai           & Transport & 183,917 & 37    & 10 & 2.85  & 0.613  & \citep{liang2026towards} \\
squirrel           & Web       & 2,223   & 2,089 & 5  & 42.28 & 0.009  & \citep{platonov2023critical} \\
tag\_bookchild     & Commerce  & 76,875  & 3,072 & 24 & 30.24 & 0.265  & \citep{wang2025generalization} \\
tag\_bookhis       & Commerce  & 41,551  & 3,072 & 12 & 12.11 & 0.519  & \citep{wang2025generalization} \\
tag\_citeseer      & Scholarly & 3,186   & 3,072 & 6  & 2.65  & 0.729  & \citep{wang2025generalization} \\
tag\_cora          & Scholarly & 2,708   & 3,072 & 7  & 3.90  & 0.771  & \citep{wang2025generalization} \\
tag\_cornell       & Web       & 191     & 3,072 & 5  & 2.87  & -0.225 & \citep{wang2025generalization} \\
tag\_pubmed        & Scholarly & 19,717  & 3,072 & 3  & 4.50  & 0.686  & \citep{wang2025generalization} \\
tag\_sportsfit     & Commerce  & 173,055 & 3,072 & 13 & 17.45 & 0.851  & \citep{wang2025generalization} \\
tag\_texas         & Web       & 187     & 3,072 & 4  & 2.99  & -0.294 & \citep{wang2025generalization} \\
tag\_washington    & Web       & 229     & 3,072 & 5  & 3.19  & -0.194 & \citep{wang2025generalization} \\
tag\_wikics        & Web       & 11,701  & 3,072 & 10 & 36.85 & 0.579  & \citep{wang2025generalization} \\
tag\_wisconsin     & Web       & 265     & 3,072 & 5  & 3.46  & -0.169 & \citep{wang2025generalization} \\
texas              & Web       & 183     & 1,703 & 4  & 3.05  & -0.298 & \citep{pei2020geom} \\
tolokers-2         & Social    & 11,758  & 19    & 2  & 88.28 & 0.093  & \citep{bazhenov2026graphland} \\
wiki               & Web       & 2,405   & 4,973 & 17 & 9.64  & 0.564  & \citep{yang2020scaling} \\
wiki\_cs           & Web       & 11,701  & 300   & 10 & 36.85 & 0.579  & \citep{mernyei2020wiki} \\
wisconsin          & Web       & 251     & 1,703 & 5  & 3.59  & -0.173 & \citep{pei2020geom} \\
\bottomrule
\end{tabular*}
\end{table*}
\begin{table*}[!p]
\centering
\caption{\textbf{Default configurations and hyperparameter search spaces for 15 supervised \acp{GNN}.}
We sample 200 configurations from the listed candidate values.
$d$ denotes hidden dimension and $L$ the number of layers.
Unless otherwise specified, methods use the shared training configuration and search space in the first row.}
\label{tab:search-spaces}
\vspace{8pt}
\scriptsize
\setlength{\tabcolsep}{3.2pt}
\renewcommand{\arraystretch}{1.10}

\begin{tabularx}{\textwidth}{
    @{}
    >{\raggedright\arraybackslash}p{0.105\textwidth}
    >{\raggedright\arraybackslash}p{0.105\textwidth}
    >{\raggedright\arraybackslash}p{0.285\textwidth}
    >{\raggedright\arraybackslash}X
    @{}
}
\toprule
\textbf{Method} & \textbf{Ref.} & \textbf{Default} & \textbf{Search space} \\
\midrule

\textit{Shared training}
& --
& Adam; lr $=10^{-2}$; wd $=5\!\times\!10^{-4}$;
2,500 epochs; patience $=100$
& lr $\in\{10^{-3},5\!\times\!10^{-3},10^{-2}\}$;
wd $\in\{0,5\!\times\!10^{-5},5\!\times\!10^{-4}\}$;
patience $\in\{20,100\}$
\\

\midrule

APPNP
& \citep{gasteiger2018predict}
& $d=64$; dropout $=.5$; $K=10$; $\alpha=.1$; prop. dropout $=0$
& $d\in\{64,128,256,512\}$;
dropout $\in\{0,.3,.5,.7\}$;
$K\in\{5,10,20\}$;
$\alpha\in\{.1,.2,.5\}$;
prop. dropout $\in\{0,.3,.5\}$
\\
\addlinespace[2pt]

FAGCN
& \citep{bo2021beyond}
& $d=32$; dropout $=.5$; $L=2$; $\epsilon=.3$; edge dropout $=.5$
& $d\in\{32,64,128\}$;
feature/edge dropout $\in\{0,.3,.5,.7\}$;
$L\in\{1,2,3,4\}$;
$\epsilon\in\{.1,.2,.3,.5\}$
\\
\addlinespace[2pt]

GAT
& \citep{veličković2018graph}
& $d=64$; heads $=1$; dropout $=.5$; attn. dropout $=.2$;
$L=2$; norm=None; residual=False; pre-transform=False
& $d\in\{64,128,256,512\}$;
heads $\in\{1,4\}$;
dropout $\in\{0,.3,.5,.7\}$;
attn. dropout $\in\{0,.2\}$;
$L\in\{1,2,4,8\}$;
norm $\in\{\texttt{none},\texttt{layernorm}\}$;
residual/pre-transform $\in\{\mathrm{F},\mathrm{T}\}$
\\
\addlinespace[2pt]

GATv2
& \citep{brody2021attentive}
& $d=64$; heads $=1$; dropout $=.5$; attn. dropout $=.2$;
$L=2$; norm=None; residual=False; pre-transform=False
& $d\in\{64,128,256,512\}$;
heads $\in\{1,4\}$;
dropout $\in\{0,.3,.5,.7\}$;
attn. dropout $\in\{0,.2\}$;
$L\in\{1,2,4,8\}$;
norm $\in\{\texttt{none},\texttt{layernorm}\}$;
residual/pre-transform $\in\{\mathrm{F},\mathrm{T}\}$
\\
\addlinespace[2pt]

GCN
& \citep{kipf2017semisupervised}
& $d=64$; dropout $=.5$; $L=2$;
norm=None; residual=False; pre-transform=False
& $d\in\{64,128,256,512\}$;
dropout $\in\{0,.3,.5,.7\}$;
$L\in\{1,2,4,8\}$;
norm $\in\{\texttt{none},\texttt{layernorm}\}$;
residual/pre-transform $\in\{\mathrm{F},\mathrm{T}\}$
\\
\addlinespace[2pt]

GraphSAGE
& \citep{hamilton2017inductive}
& $d=64$; dropout $=.5$; $L=2$;
norm=None; residual=False; pre-transform=False
& $d\in\{64,128,256,512\}$;
dropout $\in\{0,.3,.5,.7\}$;
$L\in\{1,2,4,8\}$;
norm $\in\{\texttt{none},\texttt{layernorm}\}$;
residual/pre-transform $\in\{\mathrm{F},\mathrm{T}\}$
\\
\addlinespace[2pt]

GCNII
& \citep{chen2020simple}
& $d=64$; dropout $=.6$; $L=16$; $\alpha=.1$; $\theta=.5$;
shared weights; norm=None; residual=False; pre-transform=False
& $d\in\{64,128,256,512\}$;
dropout $\in\{0,.3,.5,.7\}$;
$L\in\{4,8,16,32\}$;
$\alpha\in\{.1,.2,.5\}$;
$\theta\in\{.5,1,1.5\}$;
norm/residual/pre-transform choices
\\
\addlinespace[2pt]

GPRGNN
& \citep{chien2020adaptive}
& $d=64$; feature dropout $=.5$; prop. dropout $=0$;
$K=10$; $\alpha=.1$; init.=PPR
& $d\in\{64,128,256,512\}$;
both dropouts $\in\{0,.3,.5,.7\}$;
$K\in\{5,10,15,20\}$;
$\alpha\in\{.1,.2,.5,.9\}$;
init. $\in\{\texttt{ppr},\texttt{sgc},\texttt{random}\}$
\\
\addlinespace[2pt]

LINKX
& \citep{lim2021large}
& $d=32$; dropout $=.5$; $L=2$;
adjacency layers $=1$; feature layers $=1$
& $d\in\{32,64,128\}$;
dropout $\in\{0,.3,.5,.7\}$;
$L\in\{1,2,3\}$;
adjacency/feature layers $\in\{1,2\}$
\\
\addlinespace[2pt]

MLP
& --
& $d=64$; dropout $=.5$; $L=2$;
norm=None; identity skip=False
& $d\in\{64,128,256,512\}$;
dropout $\in\{0,.3,.5,.7\}$;
$L\in\{1,2,3,4\}$;
norm $\in\{\texttt{none},\texttt{layernorm}\}$;
identity skip $\in\{\mathrm{F},\mathrm{T}\}$
\\
\addlinespace[2pt]

NAGphormer
& \citep{chen2022nagphormer}
& AdamW; $d=512$; hops $=7$; $L=1$; heads $=8$;
dropout $=.1$; PE dim. $=15$; lr $=10^{-3}$; wd $=10^{-5}$
& $d\in\{128,256,512\}$;
hops $\in\{2,\ldots,20\}$;
$L\in\{1,\ldots,5\}$;
dropout $\in\{.1,.3,.5\}$;
lr $\in\{10^{-4},10^{-3},5\!\times\!10^{-3}\}$;
wd $\in\{10^{-5},10^{-4},5\!\times\!10^{-4}\}$
\\
\addlinespace[2pt]

NodeFormer
& \citep{wu2022nodeformer}
& $d=32$; $L=2$; heads $=4$; dropout $=0$;
random features $=30$; Gumbel samples $=10$; link weight $=1$;
bias order $=2$; transform=sigmoid;
lr $=10^{-3}$; wd $=5\!\times\!10^{-3}$
& $d\in\{32,64,128\}$;
$L\in\{2,3\}$;
heads $\in\{1,2,4\}$;
dropout $\in\{0,.3,.5\}$;
random features $\in\{30,50\}$;
samples $\in\{5,10\}$;
link weight $\in\{.01,.1,1\}$;
bias order $\in\{1,2,3\}$;
transform $\in\{\texttt{sigmoid},\texttt{identity}\}$;
activation/JK $\in\{\mathrm{F},\mathrm{T}\}$;
lr $\in\{10^{-5},10^{-4},10^{-3},10^{-2}\}$;
wd $\in\{0,5\!\times\!10^{-4},5\!\times\!10^{-3},.05\}$
\\
\addlinespace[2pt]

Polynormer
& \citep{deng2024polynormer}
& $d=64$; local/global layers $=7/2$; heads $=1$; $\beta=-1$;
input/model/global dropout $=.15/.5/.5$;
local epochs $=100$; lr $=10^{-3}$; wd $=5\!\times\!10^{-4}$
& $d\in\{64,128,256,512\}$;
local layers $\in\{5,7,10\}$;
global layers $\in\{1,2,3,4\}$;
heads $\in\{1,2,4,8\}$;
$\beta\in\{-1,.1,.5,.9\}$;
input dropout $\in\{0,.15,.2,.5\}$;
model dropout $\in\{.2,.3,.5,.7\}$;
global dropout $\in\{.3,.5\}$;
pre-LN $\in\{\mathrm{F},\mathrm{T}\}$;
local epochs $\in\{100,200\}$;
lr $\in\{3\!\times\!10^{-5},5\!\times\!10^{-4},10^{-3}\}$;
wd $\in\{0,5\!\times\!10^{-5},5\!\times\!10^{-4}\}$
\\
\addlinespace[2pt]

SGC
& \citep{wu2019simplifying}
& dropout $=0$; $K=2$
& dropout $\in\{0,.2,.5\}$;
$K\in\{1,\ldots,8\}$
\\
\addlinespace[2pt]

SGFormer
& \citep{wu2023sgformer}
& $d=64$; graph weight $=.5$;
Transformer/GNN layers $=1/2$;
Transformer/GNN dropout $=0/0$;
initial-feature reuse=False
& $d\in\{64,128,256,512\}$;
graph weight $\in\{.5,.8\}$;
Transformer/GNN dropout $\in\{0,.3,.5\}$;
GNN layers $\in\{1,2,4,8\}$;
initial-feature reuse $\in\{\mathrm{F},\mathrm{T}\}$
\\

\bottomrule
\end{tabularx}
\end{table*}

\newpage
\section{Beyond Standard Node Classification}
\label[appendix]{app:beyond-standard}

Our main evaluation considers graph node classification under random context-query splits.
We further examine how \method{} generalizes along two dimensions: distribution shifts between context and query nodes, and hypergraph-structured data unseen during pretraining.
For both, we follow existing benchmarks and their established protocols, using the same pretrained \method{} checkpoint as in the main experiments for single-pass inference.

\mypar{Robustness under distribution shift.}
We first ask whether \method{} remains effective when labeled context and query nodes follow different distributions.
We evaluate on four real-world node-classification datasets from the Graph Out-of-Distribution (GOOD) benchmark~\citep{gui2022good}.
Together, they define six dataset-attribute settings: word diversity and degree for Cora, publication time and degree for Arxiv, university for WebKB, and language for Twitch.
For each setting, GOOD provides a random \emph{no-shift} split and constructs \emph{covariate shift} by partitioning nodes according to the selected attribute.
For \emph{concept shift}, the partition additionally depends on the class label, altering the relationship between the attribute and label across training and test nodes.

We use the official splits provided by GOOD (\href{https://github.com/divelab/GOOD}{link}) and compare against the nine baselines reported in the original work.
These baselines are evaluated under two model-selection protocols: \emph{ID-selected} selects models using validation data from the training distribution, whereas \emph{OOD-selected} uses labeled validation data from the shifted distribution.
As in our main experiments, \method{} predicts using only the training-node labels as context, without access to any validation data.

As shown in \Cref{tab:good-covariate,tab:good-concept}, \method{} ranks first in all six no-shift settings, consistent with our main experiments.
Under distribution shift, its advantage narrows but remains competitive: \method{} achieves the best average rank against ID-selected baselines (4.00) and ranks fourth against OOD-selected baselines (5.08).
These results are encouraging given that several baselines are specifically designed for OOD generalization, whereas \method{} is not explicitly designed to handle distribution shift.

The degradation under distribution shift is concentrated in a few specific settings rather than being uniform across tasks.
\method{} degrades substantially under degree-based covariate shift on Arxiv, time-based concept shift on Arxiv, university-based covariate shift on WebKB, and language-based concept shift on Twitch, while remaining competitive in the other settings.
This overall competitiveness, together with these localized failures, suggests that \method{} can generalize under distribution shift, but its robustness remains case-dependent.
A promising direction is to incorporate distribution-shift scenarios into the synthetic graph prior during pretraining.

\mypar{Hypergraph transfer.}
We further test whether \method{}, pretrained exclusively on graphs, can transfer to hypergraph node classification.
As shown in \Cref{tab:hypergraph-transfer}, we evaluate on the ten hypergraph datasets from AllSet~\citep{chien2021you} using the same random 50/25/25 splits over 20 runs.
Since \method{} operates on graphs, we convert each hypergraph using either clique expansion, which connects nodes sharing a hyperedge, or an incidence representation, which introduces each hyperedge as an unlabeled node.
For both representations, we use the same frozen checkpoint as in the main experiments and perform inference from the 50\% labeled context, without hypergraph-specific training or validation-based model selection.
In contrast, the published baselines are trained separately on each target hypergraph and use the 25\% validation split for model selection.

Despite this difference, \method{} achieves or ties the highest reported accuracy on six of the ten datasets and remains competitive on most others.
Neither representation consistently dominates, suggesting that the preferred graph representation depends on the target hypergraph.
These results show that the learned graph \ac{ICL} mechanism transfers effectively to hypergraphs through simple graph representations.
More broadly, they suggest that the inference-only paradigm of \method{} may extend beyond replacing per-dataset \ac{GNN} training to reducing the need for specialized training on other graph-structured data such as hypergraphs.
\begin{table*}[p]
\centering
\begingroup
\footnotesize
\definecolor{goodFirst}{HTML}{D62728}
\definecolor{goodSecond}{HTML}{1565C0}
\definecolor{goodThird}{HTML}{16803C}
\setlength{\tabcolsep}{4pt}
\renewcommand{\arraystretch}{1.06}
\caption{\textbf{GOOD: no shift and covariate shift.}
Results on the official GOOD splits using either ID or OOD validation for baseline model selection.
Scores are mean $\pm$ SD in percent (ROC-AUC for Twitch; accuracy otherwise).
Baseline results are reported by GOOD~\citep{gui2022good}, while Ephris uses the same frozen checkpoint in both settings without model selection.
\textcolor{goodFirst}{Red}, \textcolor{goodSecond}{blue}, and \textcolor{goodThird}{green} mark the three best distinct scores in each column.
Average ranks are computed separately for ID- and OOD-based selection across the six dataset-attribute settings.}
\label{tab:good-covariate}
\fontsize{8}{8.4}\selectfont
\begin{tabular}{@{}lccccccc@{}}
\toprule
\multicolumn{8}{@{}l}{\textit{No shift --- ID-selected}} \\
\addlinespace[2pt]
Method & \multicolumn{2}{c}{Cora} & \multicolumn{2}{c}{Arxiv} & WebKB & Twitch & Avg. rank $\downarrow$ \\
\cmidrule(lr){2-3}\cmidrule(lr){4-5}
 & Word & Degree & Time & Degree & University & Language & \\
\midrule
ERM & $69.41\,{\scriptstyle\pm\,0.30}$ & $69.42\,{\scriptstyle\pm\,0.30}$ & $\textcolor{goodThird}{\mathbf{73.02}}\,{\scriptstyle\pm\,0.14}$ & $\textcolor{goodThird}{\mathbf{72.99}}\,{\scriptstyle\pm\,0.12}$ & $47.85\,{\scriptstyle\pm\,0.89}$ & $68.05\,{\scriptstyle\pm\,0.52}$ & 5.92 \\
IRM & $69.42\,{\scriptstyle\pm\,0.38}$ & $69.40\,{\scriptstyle\pm\,0.38}$ & $72.90\,{\scriptstyle\pm\,0.14}$ & $72.92\,{\scriptstyle\pm\,0.07}$ & $47.31\,{\scriptstyle\pm\,1.21}$ & $68.30\,{\scriptstyle\pm\,0.29}$ & 7.25 \\
VREx & $69.43\,{\scriptstyle\pm\,0.29}$ & $69.42\,{\scriptstyle\pm\,0.29}$ & $72.84\,{\scriptstyle\pm\,0.09}$ & $72.88\,{\scriptstyle\pm\,0.09}$ & $47.85\,{\scriptstyle\pm\,0.89}$ & $68.07\,{\scriptstyle\pm\,0.52}$ & 7.25 \\
GroupDRO & $69.46\,{\scriptstyle\pm\,0.25}$ & $69.40\,{\scriptstyle\pm\,0.30}$ & $72.91\,{\scriptstyle\pm\,0.12}$ & $72.98\,{\scriptstyle\pm\,0.10}$ & $47.85\,{\scriptstyle\pm\,0.89}$ & $\textcolor{goodThird}{\mathbf{69.19}}\,{\scriptstyle\pm\,0.28}$ & 5.75 \\
DANN & $69.25\,{\scriptstyle\pm\,0.33}$ & $69.24\,{\scriptstyle\pm\,0.34}$ & $73.00\,{\scriptstyle\pm\,0.12}$ & $72.97\,{\scriptstyle\pm\,0.10}$ & $47.85\,{\scriptstyle\pm\,0.89}$ & $68.07\,{\scriptstyle\pm\,0.52}$ & 7.00 \\
Deep Coral & $69.46\,{\scriptstyle\pm\,0.27}$ & $69.43\,{\scriptstyle\pm\,0.30}$ & $72.95\,{\scriptstyle\pm\,0.09}$ & $72.91\,{\scriptstyle\pm\,0.12}$ & $48.12\,{\scriptstyle\pm\,0.89}$ & $68.29\,{\scriptstyle\pm\,0.65}$ & 5.42 \\
Mixup & $\textcolor{goodSecond}{\mathbf{70.56}}\,{\scriptstyle\pm\,0.35}$ & $\textcolor{goodSecond}{\mathbf{70.87}}\,{\scriptstyle\pm\,0.47}$ & $\textcolor{goodSecond}{\mathbf{73.19}}\,{\scriptstyle\pm\,0.16}$ & $\textcolor{goodSecond}{\mathbf{73.03}}\,{\scriptstyle\pm\,0.14}$ & $\textcolor{goodThird}{\mathbf{51.88}}\,{\scriptstyle\pm\,1.34}$ & $67.09\,{\scriptstyle\pm\,0.34}$ & 3.50 \\
EERM & $\textcolor{goodThird}{\mathbf{70.10}}\,{\scriptstyle\pm\,0.22}$ & $\textcolor{goodThird}{\mathbf{70.38}}\,{\scriptstyle\pm\,0.24}$ & OOM & OOM & $50.54\,{\scriptstyle\pm\,0.46}$ & $\textcolor{goodSecond}{\mathbf{70.80}}\,{\scriptstyle\pm\,0.08}$ & 5.33 \\
SRGNN & $69.05\,{\scriptstyle\pm\,0.54}$ & $69.08\,{\scriptstyle\pm\,0.53}$ & $72.99\,{\scriptstyle\pm\,0.04}$ & $\textcolor{goodThird}{\mathbf{72.99}}\,{\scriptstyle\pm\,0.02}$ & $\textcolor{goodSecond}{\mathbf{52.96}}\,{\scriptstyle\pm\,1.04}$ & $67.69\,{\scriptstyle\pm\,0.13}$ & 6.58 \\
\midrule
\textbf{Ephris} & $\textcolor{goodFirst}{\mathbf{71.59}}\,{\scriptstyle\pm\,0.32}$ & $\textcolor{goodFirst}{\mathbf{71.57}}\,{\scriptstyle\pm\,0.31}$ & $\textcolor{goodFirst}{\mathbf{74.98}}\,{\scriptstyle\pm\,0.06}$ & $\textcolor{goodFirst}{\mathbf{74.97}}\,{\scriptstyle\pm\,0.05}$ & $\textcolor{goodFirst}{\mathbf{81.45}}\,{\scriptstyle\pm\,0.00}$ & $\textcolor{goodFirst}{\mathbf{73.06}}\,{\scriptstyle\pm\,0.01}$ & \textbf{1.00} \\
\bottomrule
\end{tabular}
\par\medskip
\begin{tabular}{@{}lccccccc@{}}
\toprule
\multicolumn{8}{@{}l}{\textit{Covariate shift --- ID-selected}} \\
\addlinespace[2pt]
Method & \multicolumn{2}{c}{Cora} & \multicolumn{2}{c}{Arxiv} & WebKB & Twitch & Avg. rank $\downarrow$ \\
\cmidrule(lr){2-3}\cmidrule(lr){4-5}
 & Word & Degree & Time & Degree & University & Language & \\
\midrule
ERM & $64.44\,{\scriptstyle\pm\,0.55}$ & $55.76\,{\scriptstyle\pm\,0.82}$ & $70.64\,{\scriptstyle\pm\,0.47}$ & $58.53\,{\scriptstyle\pm\,0.16}$ & $11.64\,{\scriptstyle\pm\,0.90}$ & $47.73\,{\scriptstyle\pm\,0.72}$ & 5.67 \\
IRM & $\textcolor{goodSecond}{\mathbf{64.83}}\,{\scriptstyle\pm\,0.25}$ & $\textcolor{goodThird}{\mathbf{55.77}}\,{\scriptstyle\pm\,0.46}$ & $70.55\,{\scriptstyle\pm\,0.33}$ & $\textcolor{goodFirst}{\mathbf{58.70}}\,{\scriptstyle\pm\,0.12}$ & $11.91\,{\scriptstyle\pm\,2.62}$ & $48.05\,{\scriptstyle\pm\,0.16}$ & 3.83 \\
VREx & $64.49\,{\scriptstyle\pm\,0.55}$ & $55.46\,{\scriptstyle\pm\,0.87}$ & $70.54\,{\scriptstyle\pm\,0.33}$ & $\textcolor{goodThird}{\mathbf{58.59}}\,{\scriptstyle\pm\,0.21}$ & $10.58\,{\scriptstyle\pm\,1.02}$ & $47.70\,{\scriptstyle\pm\,0.70}$ & 6.83 \\
GroupDRO & $64.49\,{\scriptstyle\pm\,0.66}$ & $55.44\,{\scriptstyle\pm\,0.91}$ & $70.67\,{\scriptstyle\pm\,0.31}$ & $58.46\,{\scriptstyle\pm\,0.21}$ & $12.96\,{\scriptstyle\pm\,1.95}$ & $47.23\,{\scriptstyle\pm\,0.26}$ & 6.00 \\
DANN & $\textcolor{goodThird}{\mathbf{64.72}}\,{\scriptstyle\pm\,0.22}$ & $55.50\,{\scriptstyle\pm\,0.60}$ & $70.57\,{\scriptstyle\pm\,0.40}$ & $58.56\,{\scriptstyle\pm\,0.16}$ & $\textcolor{goodSecond}{\mathbf{15.34}}\,{\scriptstyle\pm\,1.02}$ & $47.72\,{\scriptstyle\pm\,0.73}$ & 4.67 \\
Deep Coral & $64.63\,{\scriptstyle\pm\,0.38}$ & $55.52\,{\scriptstyle\pm\,0.93}$ & $70.59\,{\scriptstyle\pm\,0.29}$ & $\textcolor{goodSecond}{\mathbf{58.63}}\,{\scriptstyle\pm\,0.21}$ & $\textcolor{goodThird}{\mathbf{14.29}}\,{\scriptstyle\pm\,2.92}$ & $46.64\,{\scriptstyle\pm\,0.70}$ & 4.83 \\
Mixup & $63.07\,{\scriptstyle\pm\,1.52}$ & $\textcolor{goodFirst}{\mathbf{57.21}}\,{\scriptstyle\pm\,1.12}$ & $\textcolor{goodFirst}{\mathbf{71.05}}\,{\scriptstyle\pm\,0.31}$ & $57.43\,{\scriptstyle\pm\,0.27}$ & $10.85\,{\scriptstyle\pm\,0.66}$ & $\textcolor{goodThird}{\mathbf{51.33}}\,{\scriptstyle\pm\,1.50}$ & 5.00 \\
EERM & $60.80\,{\scriptstyle\pm\,0.61}$ & $55.23\,{\scriptstyle\pm\,0.40}$ & OOM & OOM & $11.90\,{\scriptstyle\pm\,0.37}$ & $\textcolor{goodSecond}{\mathbf{52.48}}\,{\scriptstyle\pm\,0.76}$ & 7.83 \\
SRGNN & $64.49\,{\scriptstyle\pm\,0.19}$ & $54.67\,{\scriptstyle\pm\,0.36}$ & $\textcolor{goodThird}{\mathbf{70.70}}\,{\scriptstyle\pm\,0.42}$ & $57.48\,{\scriptstyle\pm\,0.07}$ & $\textcolor{goodFirst}{\mathbf{16.14}}\,{\scriptstyle\pm\,3.35}$ & $46.17\,{\scriptstyle\pm\,0.98}$ & 6.17 \\
\midrule
\textbf{Ephris} & $\textcolor{goodFirst}{\mathbf{66.41}}\,{\scriptstyle\pm\,0.22}$ & $\textcolor{goodSecond}{\mathbf{57.20}}\,{\scriptstyle\pm\,0.41}$ & $\textcolor{goodSecond}{\mathbf{70.91}}\,{\scriptstyle\pm\,0.16}$ & $56.73\,{\scriptstyle\pm\,0.46}$ & $6.75\,{\scriptstyle\pm\,0.42}$ & $\textcolor{goodFirst}{\mathbf{57.14}}\,{\scriptstyle\pm\,0.04}$ & \textbf{4.17} \\
\bottomrule
\end{tabular}
\par\medskip
\begin{tabular}{@{}lccccccc@{}}
\toprule
\multicolumn{8}{@{}l}{\textit{Covariate shift --- OOD-selected}} \\
\addlinespace[2pt]
Method & \multicolumn{2}{c}{Cora} & \multicolumn{2}{c}{Arxiv} & WebKB & Twitch & Avg. rank $\downarrow$ \\
\cmidrule(lr){2-3}\cmidrule(lr){4-5}
 & Word & Degree & Time & Degree & University & Language & \\
\midrule
ERM & $\textcolor{goodThird}{\mathbf{64.86}}\,{\scriptstyle\pm\,0.38}$ & $56.30\,{\scriptstyle\pm\,0.49}$ & $71.08\,{\scriptstyle\pm\,0.23}$ & $58.91\,{\scriptstyle\pm\,0.23}$ & $14.29\,{\scriptstyle\pm\,3.24}$ & $48.95\,{\scriptstyle\pm\,3.19}$ & 5.17 \\
IRM & $64.77\,{\scriptstyle\pm\,0.36}$ & $56.28\,{\scriptstyle\pm\,0.63}$ & $71.04\,{\scriptstyle\pm\,0.16}$ & $58.98\,{\scriptstyle\pm\,0.28}$ & $13.49\,{\scriptstyle\pm\,0.75}$ & $47.21\,{\scriptstyle\pm\,0.98}$ & 6.92 \\
VREx & $64.80\,{\scriptstyle\pm\,0.28}$ & $56.30\,{\scriptstyle\pm\,0.50}$ & $\textcolor{goodThird}{\mathbf{71.12}}\,{\scriptstyle\pm\,0.24}$ & $\textcolor{goodThird}{\mathbf{58.99}}\,{\scriptstyle\pm\,0.16}$ & $14.29\,{\scriptstyle\pm\,3.24}$ & $48.99\,{\scriptstyle\pm\,3.20}$ & 4.33 \\
GroupDRO & $64.72\,{\scriptstyle\pm\,0.34}$ & $56.29\,{\scriptstyle\pm\,0.43}$ & $\textcolor{goodSecond}{\mathbf{71.15}}\,{\scriptstyle\pm\,0.20}$ & $\textcolor{goodFirst}{\mathbf{59.08}}\,{\scriptstyle\pm\,0.16}$ & $\textcolor{goodThird}{\mathbf{17.20}}\,{\scriptstyle\pm\,0.76}$ & $47.20\,{\scriptstyle\pm\,0.44}$ & 5.08 \\
DANN & $64.77\,{\scriptstyle\pm\,0.42}$ & $56.10\,{\scriptstyle\pm\,0.59}$ & $71.05\,{\scriptstyle\pm\,0.29}$ & $\textcolor{goodSecond}{\mathbf{59.00}}\,{\scriptstyle\pm\,0.18}$ & $15.08\,{\scriptstyle\pm\,0.37}$ & $48.98\,{\scriptstyle\pm\,3.22}$ & 5.42 \\
Deep Coral & $64.72\,{\scriptstyle\pm\,0.36}$ & $56.35\,{\scriptstyle\pm\,0.38}$ & $71.07\,{\scriptstyle\pm\,0.21}$ & $58.97\,{\scriptstyle\pm\,0.20}$ & $13.76\,{\scriptstyle\pm\,1.30}$ & $49.64\,{\scriptstyle\pm\,2.44}$ & 5.42 \\
Mixup & $\textcolor{goodSecond}{\mathbf{65.23}}\,{\scriptstyle\pm\,0.56}$ & $\textcolor{goodFirst}{\mathbf{58.20}}\,{\scriptstyle\pm\,0.67}$ & $\textcolor{goodFirst}{\mathbf{71.34}}\,{\scriptstyle\pm\,0.14}$ & $57.60\,{\scriptstyle\pm\,0.31}$ & $\textcolor{goodSecond}{\mathbf{17.46}}\,{\scriptstyle\pm\,1.94}$ & $\textcolor{goodSecond}{\mathbf{52.27}}\,{\scriptstyle\pm\,0.78}$ & 2.50 \\
EERM & $61.98\,{\scriptstyle\pm\,0.10}$ & $\textcolor{goodThird}{\mathbf{56.88}}\,{\scriptstyle\pm\,0.32}$ & OOM & OOM & $\textcolor{goodFirst}{\mathbf{24.61}}\,{\scriptstyle\pm\,4.86}$ & $\textcolor{goodThird}{\mathbf{51.34}}\,{\scriptstyle\pm\,1.41}$ & 6.17 \\
SRGNN & $64.66\,{\scriptstyle\pm\,0.21}$ & $54.78\,{\scriptstyle\pm\,0.10}$ & $70.83\,{\scriptstyle\pm\,0.10}$ & $57.52\,{\scriptstyle\pm\,0.10}$ & $13.23\,{\scriptstyle\pm\,2.93}$ & $47.30\,{\scriptstyle\pm\,1.43}$ & 8.83 \\
\midrule
\textbf{Ephris} & $\textcolor{goodFirst}{\mathbf{66.41}}\,{\scriptstyle\pm\,0.22}$ & $\textcolor{goodSecond}{\mathbf{57.20}}\,{\scriptstyle\pm\,0.41}$ & $70.91\,{\scriptstyle\pm\,0.16}$ & $56.73\,{\scriptstyle\pm\,0.46}$ & $6.75\,{\scriptstyle\pm\,0.42}$ & $\textcolor{goodFirst}{\mathbf{57.14}}\,{\scriptstyle\pm\,0.04}$ & \textbf{5.17} \\
\bottomrule
\end{tabular}
\endgroup
\end{table*}

\begin{table*}[p]
\centering
\begingroup
\footnotesize
\definecolor{goodFirst}{HTML}{D62728}
\definecolor{goodSecond}{HTML}{1565C0}
\definecolor{goodThird}{HTML}{16803C}
\setlength{\tabcolsep}{4pt}
\renewcommand{\arraystretch}{1.06}
\caption{\textbf{GOOD: concept shift.}
Results on the official GOOD splits using either ID or OOD validation for baseline model selection.
Scores are mean $\pm$ SD in percent (ROC-AUC for Twitch; accuracy otherwise).
Baseline results are reported by GOOD~\citep{gui2022good}, while Ephris uses the same frozen checkpoint in both settings without model selection.
\textcolor{goodFirst}{Red}, \textcolor{goodSecond}{blue}, and \textcolor{goodThird}{green} mark the three best distinct scores in each column.
Average ranks are computed separately for ID- and OOD-based selection across the six dataset-attribute settings.}
\label{tab:good-concept}
\fontsize{8}{8.4}\selectfont
\begin{tabular}{@{}lccccccc@{}}
\toprule
\multicolumn{8}{@{}l}{\textit{ID-selected}} \\
\addlinespace[2pt]
Method & \multicolumn{2}{c}{Cora} & \multicolumn{2}{c}{Arxiv} & WebKB & Twitch & Avg. rank $\downarrow$ \\
\cmidrule(lr){2-3}\cmidrule(lr){4-5}
 & Word & Degree & Time & Degree & University & Language & \\
\midrule
ERM & $64.20\,{\scriptstyle\pm\,0.56}$ & $60.38\,{\scriptstyle\pm\,0.33}$ & $\textcolor{goodFirst}{\mathbf{65.70}}\,{\scriptstyle\pm\,0.42}$ & $\textcolor{goodSecond}{\mathbf{61.77}}\,{\scriptstyle\pm\,0.29}$ & $24.77\,{\scriptstyle\pm\,0.43}$ & $\textcolor{goodThird}{\mathbf{48.57}}\,{\scriptstyle\pm\,0.17}$ & 4.17 \\
IRM & $64.16\,{\scriptstyle\pm\,0.61}$ & $\textcolor{goodThird}{\mathbf{61.00}}\,{\scriptstyle\pm\,0.34}$ & $\textcolor{goodSecond}{\mathbf{65.69}}\,{\scriptstyle\pm\,0.55}$ & $61.49\,{\scriptstyle\pm\,0.36}$ & $24.16\,{\scriptstyle\pm\,0.80}$ & $\textcolor{goodSecond}{\mathbf{49.77}}\,{\scriptstyle\pm\,0.82}$ & 5.50 \\
VREx & $64.20\,{\scriptstyle\pm\,0.54}$ & $60.05\,{\scriptstyle\pm\,0.72}$ & $65.40\,{\scriptstyle\pm\,0.54}$ & $61.61\,{\scriptstyle\pm\,0.32}$ & $24.77\,{\scriptstyle\pm\,0.43}$ & $48.56\,{\scriptstyle\pm\,0.18}$ & 6.00 \\
GroupDRO & $\textcolor{goodThird}{\mathbf{64.38}}\,{\scriptstyle\pm\,0.34}$ & $60.03\,{\scriptstyle\pm\,0.88}$ & $\textcolor{goodThird}{\mathbf{65.57}}\,{\scriptstyle\pm\,0.66}$ & $61.59\,{\scriptstyle\pm\,0.56}$ & $24.77\,{\scriptstyle\pm\,0.43}$ & $47.44\,{\scriptstyle\pm\,1.08}$ & 5.67 \\
DANN & $64.29\,{\scriptstyle\pm\,0.33}$ & $59.65\,{\scriptstyle\pm\,0.94}$ & $65.42\,{\scriptstyle\pm\,0.53}$ & $61.43\,{\scriptstyle\pm\,0.40}$ & $24.77\,{\scriptstyle\pm\,0.43}$ & $\textcolor{goodThird}{\mathbf{48.57}}\,{\scriptstyle\pm\,0.18}$ & 6.17 \\
Deep Coral & $\textcolor{goodThird}{\mathbf{64.38}}\,{\scriptstyle\pm\,0.36}$ & $60.22\,{\scriptstyle\pm\,0.55}$ & $65.53\,{\scriptstyle\pm\,0.63}$ & $\textcolor{goodSecond}{\mathbf{61.77}}\,{\scriptstyle\pm\,0.37}$ & $24.77\,{\scriptstyle\pm\,0.43}$ & $47.46\,{\scriptstyle\pm\,0.32}$ & 4.75 \\
Mixup & $64.22\,{\scriptstyle\pm\,0.33}$ & $\textcolor{goodFirst}{\mathbf{63.49}}\,{\scriptstyle\pm\,0.23}$ & $64.01\,{\scriptstyle\pm\,0.50}$ & $60.60\,{\scriptstyle\pm\,1.01}$ & $\textcolor{goodSecond}{\mathbf{27.83}}\,{\scriptstyle\pm\,1.53}$ & $\textcolor{goodFirst}{\mathbf{51.87}}\,{\scriptstyle\pm\,0.37}$ & 4.50 \\
EERM & $63.35\,{\scriptstyle\pm\,0.03}$ & $57.46\,{\scriptstyle\pm\,0.87}$ & OOM & OOM & $24.77\,{\scriptstyle\pm\,0.43}$ & $44.22\,{\scriptstyle\pm\,0.81}$ & 9.42 \\
SRGNN & $\textcolor{goodSecond}{\mathbf{64.90}}\,{\scriptstyle\pm\,0.03}$ & $59.96\,{\scriptstyle\pm\,0.89}$ & $65.37\,{\scriptstyle\pm\,0.22}$ & $\textcolor{goodThird}{\mathbf{61.74}}\,{\scriptstyle\pm\,0.10}$ & $\textcolor{goodThird}{\mathbf{25.08}}\,{\scriptstyle\pm\,1.13}$ & $48.27\,{\scriptstyle\pm\,1.10}$ & 5.00 \\
\midrule
\textbf{Ephris} & $\textcolor{goodFirst}{\mathbf{66.30}}\,{\scriptstyle\pm\,0.22}$ & $\textcolor{goodSecond}{\mathbf{63.22}}\,{\scriptstyle\pm\,0.26}$ & $63.60\,{\scriptstyle\pm\,0.20}$ & $\textcolor{goodFirst}{\mathbf{62.28}}\,{\scriptstyle\pm\,0.41}$ & $\textcolor{goodFirst}{\mathbf{48.62}}\,{\scriptstyle\pm\,0.00}$ & $44.72\,{\scriptstyle\pm\,0.01}$ & \textbf{3.83} \\
\bottomrule
\end{tabular}
\par\medskip
\begin{tabular}{@{}lccccccc@{}}
\toprule
\multicolumn{8}{@{}l}{\textit{OOD-selected}} \\
\addlinespace[2pt]
Method & \multicolumn{2}{c}{Cora} & \multicolumn{2}{c}{Arxiv} & WebKB & Twitch & Avg. rank $\downarrow$ \\
\cmidrule(lr){2-3}\cmidrule(lr){4-5}
 & Word & Degree & Time & Degree & University & Language & \\
\midrule
ERM & $\textcolor{goodThird}{\mathbf{64.60}}\,{\scriptstyle\pm\,0.17}$ & $60.54\,{\scriptstyle\pm\,0.44}$ & $67.32\,{\scriptstyle\pm\,0.24}$ & $\textcolor{goodSecond}{\mathbf{62.99}}\,{\scriptstyle\pm\,0.20}$ & $27.83\,{\scriptstyle\pm\,0.76}$ & $57.32\,{\scriptstyle\pm\,0.18}$ & 5.25 \\
IRM & $\textcolor{goodThird}{\mathbf{64.60}}\,{\scriptstyle\pm\,0.16}$ & $\textcolor{goodThird}{\mathbf{61.23}}\,{\scriptstyle\pm\,0.32}$ & $\textcolor{goodThird}{\mathbf{67.41}}\,{\scriptstyle\pm\,0.16}$ & $\textcolor{goodThird}{\mathbf{62.97}}\,{\scriptstyle\pm\,0.27}$ & $27.52\,{\scriptstyle\pm\,0.43}$ & $\textcolor{goodSecond}{\mathbf{59.17}}\,{\scriptstyle\pm\,0.85}$ & 4.00 \\
VREx & $64.57\,{\scriptstyle\pm\,0.18}$ & $60.58\,{\scriptstyle\pm\,0.42}$ & $67.37\,{\scriptstyle\pm\,0.27}$ & $\textcolor{goodFirst}{\mathbf{63.00}}\,{\scriptstyle\pm\,0.33}$ & $27.83\,{\scriptstyle\pm\,0.38}$ & $57.37\,{\scriptstyle\pm\,0.14}$ & 4.92 \\
GroupDRO & $\textcolor{goodSecond}{\mathbf{64.62}}\,{\scriptstyle\pm\,0.17}$ & $60.65\,{\scriptstyle\pm\,0.31}$ & $\textcolor{goodFirst}{\mathbf{67.45}}\,{\scriptstyle\pm\,0.15}$ & $62.88\,{\scriptstyle\pm\,0.24}$ & $28.14\,{\scriptstyle\pm\,1.12}$ & $\textcolor{goodFirst}{\mathbf{60.27}}\,{\scriptstyle\pm\,0.62}$ & 3.25 \\
DANN & $64.51\,{\scriptstyle\pm\,0.19}$ & $60.78\,{\scriptstyle\pm\,0.38}$ & $67.28\,{\scriptstyle\pm\,0.16}$ & $62.91\,{\scriptstyle\pm\,0.22}$ & $26.91\,{\scriptstyle\pm\,0.63}$ & $\textcolor{goodThird}{\mathbf{57.46}}\,{\scriptstyle\pm\,0.14}$ & 6.00 \\
Deep Coral & $64.58\,{\scriptstyle\pm\,0.18}$ & $60.58\,{\scriptstyle\pm\,0.40}$ & $\textcolor{goodSecond}{\mathbf{67.42}}\,{\scriptstyle\pm\,0.22}$ & $62.85\,{\scriptstyle\pm\,0.29}$ & $\textcolor{goodThird}{\mathbf{28.75}}\,{\scriptstyle\pm\,1.13}$ & $56.97\,{\scriptstyle\pm\,0.23}$ & 5.08 \\
Mixup & $64.44\,{\scriptstyle\pm\,0.10}$ & $\textcolor{goodFirst}{\mathbf{63.65}}\,{\scriptstyle\pm\,0.39}$ & $64.84\,{\scriptstyle\pm\,0.59}$ & $61.28\,{\scriptstyle\pm\,0.87}$ & $\textcolor{goodSecond}{\mathbf{31.19}}\,{\scriptstyle\pm\,0.43}$ & $55.28\,{\scriptstyle\pm\,0.12}$ & 6.17 \\
EERM & $63.09\,{\scriptstyle\pm\,0.36}$ & $58.38\,{\scriptstyle\pm\,0.04}$ & OOM & OOM & $27.83\,{\scriptstyle\pm\,4.12}$ & $51.94\,{\scriptstyle\pm\,4.52}$ & 9.17 \\
SRGNN & $\textcolor{goodSecond}{\mathbf{64.62}}\,{\scriptstyle\pm\,0.07}$ & $61.08\,{\scriptstyle\pm\,0.09}$ & $67.17\,{\scriptstyle\pm\,0.23}$ & $62.09\,{\scriptstyle\pm\,0.58}$ & $27.52\,{\scriptstyle\pm\,0.43}$ & $56.05\,{\scriptstyle\pm\,0.22}$ & 6.17 \\
\midrule
\textbf{Ephris} & $\textcolor{goodFirst}{\mathbf{66.30}}\,{\scriptstyle\pm\,0.22}$ & $\textcolor{goodSecond}{\mathbf{63.22}}\,{\scriptstyle\pm\,0.26}$ & $63.60\,{\scriptstyle\pm\,0.20}$ & $62.28\,{\scriptstyle\pm\,0.41}$ & $\textcolor{goodFirst}{\mathbf{48.62}}\,{\scriptstyle\pm\,0.00}$ & $44.72\,{\scriptstyle\pm\,0.01}$ & \textbf{5.00} \\
\bottomrule
\end{tabular}
\endgroup
\end{table*}

\begin{table*}[p]
\centering
\footnotesize
\definecolor{hgFirst}{HTML}{D62728}
\definecolor{hgSecond}{HTML}{1565C0}
\definecolor{hgThird}{HTML}{16803C}
\setlength{\tabcolsep}{3pt}
\renewcommand{\arraystretch}{1.06}
\caption{
\textbf{Transfer to hypergraph node classification.}
Test accuracy (\%, mean $\pm$ standard deviation) under random 50/25/25 splits.
We evaluate two graph representations: \emph{clique}, which connects nodes that share a hyperedge, and \emph{incidence}, which introduces unlabeled hyperedge nodes with mean member features and a node-kind indicator.
Both \method{} variants use the same 20 splits and perform inference only, using the same pretrained checkpoint as in the main experiments without additional training.
Published baseline results are taken from AllSet~\citep{chien2021you} (Table~2), which uses the same random 50/25/25 splits over 20 runs.
\textcolor{hgFirst}{Red}, \textcolor{hgSecond}{blue} indicate the first-, second-highest accuracies, respectively, with ties sharing the same color.
}
\vspace{4pt}
\label{tab:hypergraph-transfer}
\begin{tabular*}{\textwidth}{@{\extracolsep{\fill}}lccccc@{}}
\toprule
Method & Cora & CiteSeer & PubMed & Cora-CA & DBLP-CA \\
\midrule
\multicolumn{6}{@{}l}{\textit{Published: AllSet (ICLR 2022), Table 2}} \\
AllSetTransformer & $78.59{\scriptstyle\pm1.47}$ & $\textcolor{hgThird}{\mathbf{73.08}}{\scriptstyle\pm1.20}$ & $88.72{\scriptstyle\pm0.37}$ & $83.63{\scriptstyle\pm1.47}$ & $91.53{\scriptstyle\pm0.23}$ \\
AllDeepSets & $76.88{\scriptstyle\pm1.80}$ & $70.83{\scriptstyle\pm1.63}$ & $\textcolor{hgThird}{\mathbf{88.75}}{\scriptstyle\pm0.33}$ & $81.97{\scriptstyle\pm1.50}$ & $91.27{\scriptstyle\pm0.27}$ \\
MLP & $75.17{\scriptstyle\pm1.21}$ & $72.67{\scriptstyle\pm1.56}$ & $87.47{\scriptstyle\pm0.51}$ & $74.31{\scriptstyle\pm1.89}$ & $84.83{\scriptstyle\pm0.22}$ \\
CECGN & $76.17{\scriptstyle\pm1.39}$ & $70.16{\scriptstyle\pm1.31}$ & $86.45{\scriptstyle\pm0.43}$ & $77.05{\scriptstyle\pm1.26}$ & $88.00{\scriptstyle\pm0.26}$ \\
CEGAT & $76.41{\scriptstyle\pm1.53}$ & $70.63{\scriptstyle\pm1.30}$ & $86.81{\scriptstyle\pm0.42}$ & $76.16{\scriptstyle\pm1.19}$ & $88.59{\scriptstyle\pm0.29}$ \\
HNHN & $76.36{\scriptstyle\pm1.92}$ & $72.64{\scriptstyle\pm1.57}$ & $86.90{\scriptstyle\pm0.30}$ & $77.19{\scriptstyle\pm1.49}$ & $86.78{\scriptstyle\pm0.29}$ \\
HGNN & $79.39{\scriptstyle\pm1.36}$ & $72.45{\scriptstyle\pm1.16}$ & $86.44{\scriptstyle\pm0.44}$ & $82.64{\scriptstyle\pm1.65}$ & $91.03{\scriptstyle\pm0.20}$ \\
HCHA & $79.14{\scriptstyle\pm1.02}$ & $72.42{\scriptstyle\pm1.42}$ & $86.41{\scriptstyle\pm0.36}$ & $82.55{\scriptstyle\pm0.97}$ & $90.92{\scriptstyle\pm0.22}$ \\
HyperGCN & $78.45{\scriptstyle\pm1.26}$ & $71.28{\scriptstyle\pm0.82}$ & $82.84{\scriptstyle\pm8.67}$ & $79.48{\scriptstyle\pm2.08}$ & $89.38{\scriptstyle\pm0.25}$ \\
UniGCNII & $78.81{\scriptstyle\pm1.05}$ & $73.05{\scriptstyle\pm2.21}$ & $88.25{\scriptstyle\pm0.40}$ & $83.60{\scriptstyle\pm1.14}$ & $\textcolor{hgThird}{\mathbf{91.69}}{\scriptstyle\pm0.19}$ \\
HAN (full batch)$^{*}$ & $\textcolor{hgSecond}{\mathbf{80.18}}{\scriptstyle\pm1.15}$ & $\textcolor{hgSecond}{\mathbf{74.05}}{\scriptstyle\pm1.43}$ & $86.21{\scriptstyle\pm0.48}$ & $\textcolor{hgThird}{\mathbf{84.04}}{\scriptstyle\pm1.02}$ & $90.89{\scriptstyle\pm0.23}$ \\
HAN (mini batch)$^{*}$ & $\textcolor{hgThird}{\mathbf{79.70}}{\scriptstyle\pm1.77}$ & $\textcolor{hgFirst}{\mathbf{74.12}}{\scriptstyle\pm1.52}$ & $85.32{\scriptstyle\pm2.25}$ & $81.71{\scriptstyle\pm1.73}$ & $90.17{\scriptstyle\pm0.65}$ \\
\midrule
\multicolumn{6}{@{}l}{\textit{Frozen Ephris evaluation (20 runs)}} \\
Ephris (clique) & $\textcolor{hgFirst}{\mathbf{80.76}}{\scriptstyle\pm1.17}$ & $72.76{\scriptstyle\pm1.38}$ & $\textcolor{hgFirst}{\mathbf{90.77}}{\scriptstyle\pm0.35}$ & $\textcolor{hgSecond}{\mathbf{84.51}}{\scriptstyle\pm1.05}$ & $\textcolor{hgFirst}{\mathbf{92.20}}{\scriptstyle\pm0.23}$ \\
Ephris (incidence) & $79.24{\scriptstyle\pm1.51}$ & $71.63{\scriptstyle\pm1.45}$ & $\textcolor{hgSecond}{\mathbf{90.72}}{\scriptstyle\pm0.40}$ & $\textcolor{hgFirst}{\mathbf{85.01}}{\scriptstyle\pm1.02}$ & $\textcolor{hgSecond}{\mathbf{92.01}}{\scriptstyle\pm0.22}$ \\
\bottomrule
\end{tabular*}
\par\medskip
\begin{tabular*}{\textwidth}{@{\extracolsep{\fill}}lccccc@{}}
\toprule
Method & Zoo & 20News & Mushroom & NTU2012 & ModelNet40 \\
\midrule
\multicolumn{6}{@{}l}{\textit{Published: AllSet (ICLR 2022), Table 2}} \\
AllSetTransformer & $\textcolor{hgFirst}{\mathbf{97.50}}{\scriptstyle\pm3.59}$ & $\textcolor{hgSecond}{\mathbf{81.38}}{\scriptstyle\pm0.58}$ & $\textcolor{hgFirst}{\mathbf{100.00}}{\scriptstyle\pm0.00}$ & $88.69{\scriptstyle\pm1.24}$ & $\textcolor{hgThird}{\mathbf{98.20}}{\scriptstyle\pm0.20}$ \\
AllDeepSets & $\textcolor{hgThird}{\mathbf{95.39}}{\scriptstyle\pm4.77}$ & $81.06{\scriptstyle\pm0.54}$ & $\textcolor{hgSecond}{\mathbf{99.99}}{\scriptstyle\pm0.02}$ & $88.09{\scriptstyle\pm1.52}$ & $96.98{\scriptstyle\pm0.26}$ \\
MLP & $87.18{\scriptstyle\pm4.44}$ & $\textcolor{hgFirst}{\mathbf{81.42}}{\scriptstyle\pm0.49}$ & $\textcolor{hgFirst}{\mathbf{100.00}}{\scriptstyle\pm0.00}$ & $85.52{\scriptstyle\pm1.49}$ & $96.14{\scriptstyle\pm0.36}$ \\
CECGN & $51.54{\scriptstyle\pm11.19}$ & OOM & $95.27{\scriptstyle\pm0.47}$ & $81.52{\scriptstyle\pm1.43}$ & $89.92{\scriptstyle\pm0.46}$ \\
CEGAT & $47.88{\scriptstyle\pm14.03}$ & OOM & $96.60{\scriptstyle\pm1.67}$ & $82.21{\scriptstyle\pm1.23}$ & $92.52{\scriptstyle\pm0.39}$ \\
HNHN & $93.59{\scriptstyle\pm5.88}$ & $\textcolor{hgThird}{\mathbf{81.35}}{\scriptstyle\pm0.61}$ & $\textcolor{hgFirst}{\mathbf{100.00}}{\scriptstyle\pm0.01}$ & $\textcolor{hgThird}{\mathbf{89.11}}{\scriptstyle\pm1.44}$ & $97.84{\scriptstyle\pm0.25}$ \\
HGNN & $92.50{\scriptstyle\pm4.58}$ & $80.33{\scriptstyle\pm0.42}$ & $98.73{\scriptstyle\pm0.32}$ & $87.72{\scriptstyle\pm1.35}$ & $95.44{\scriptstyle\pm0.33}$ \\
HCHA & $93.65{\scriptstyle\pm6.15}$ & $80.33{\scriptstyle\pm0.80}$ & $98.70{\scriptstyle\pm0.39}$ & $87.48{\scriptstyle\pm1.87}$ & $94.48{\scriptstyle\pm0.28}$ \\
HyperGCN & N/A & $81.05{\scriptstyle\pm0.59}$ & $47.90{\scriptstyle\pm1.04}$ & $56.36{\scriptstyle\pm4.86}$ & $75.89{\scriptstyle\pm5.26}$ \\
UniGCNII & $93.65{\scriptstyle\pm4.37}$ & $81.12{\scriptstyle\pm0.67}$ & $\textcolor{hgThird}{\mathbf{99.96}}{\scriptstyle\pm0.05}$ & $\textcolor{hgFirst}{\mathbf{89.30}}{\scriptstyle\pm1.33}$ & $98.07{\scriptstyle\pm0.23}$ \\
HAN (full batch)$^{*}$ & $85.19{\scriptstyle\pm8.18}$ & OOM & $90.86{\scriptstyle\pm2.40}$ & $83.58{\scriptstyle\pm1.46}$ & $94.04{\scriptstyle\pm0.41}$ \\
HAN (mini batch)$^{*}$ & $75.77{\scriptstyle\pm7.10}$ & $79.72{\scriptstyle\pm0.62}$ & $93.45{\scriptstyle\pm1.31}$ & $80.77{\scriptstyle\pm2.36}$ & $91.52{\scriptstyle\pm0.96}$ \\
\midrule
\multicolumn{6}{@{}l}{\textit{Frozen Ephris evaluation (20 runs)}} \\
Ephris (clique) & $93.85{\scriptstyle\pm4.74}$ & $80.86{\scriptstyle\pm0.55}$ & $\textcolor{hgFirst}{\mathbf{100.00}}{\scriptstyle\pm0.00}$ & $88.23{\scriptstyle\pm1.40}$ & $\textcolor{hgSecond}{\mathbf{98.26}}{\scriptstyle\pm0.18}$ \\
Ephris (incidence) & $\textcolor{hgSecond}{\mathbf{96.54}}{\scriptstyle\pm4.30}$ & $81.23{\scriptstyle\pm0.48}$ & $\textcolor{hgFirst}{\mathbf{100.00}}{\scriptstyle\pm0.00}$ & $\textcolor{hgSecond}{\mathbf{89.17}}{\scriptstyle\pm1.26}$ & $\textcolor{hgFirst}{\mathbf{98.34}}{\scriptstyle\pm0.12}$ \\
\bottomrule
\end{tabular*}
\par\medskip
\begin{minipage}{\textwidth}
\scriptsize
\textbf{Note.} $^{*}$HAN uses additional preprocessing in the AllSet source paper.
OOM and N/A retain AllSet's reported out-of-memory and numerical-instability entries, respectively.
\end{minipage}
\end{table*}

\newpage
\section{Ablation Studies}
\label[appendix]{app:ablation-studies}
In this section, we use controlled proxy experiments to isolate the contributions of \method{}'s key design choices.
Full pretraining requires approximately 16 GPU-days, making repeated full-scale ablations impractical.
We therefore train all variants under the same reduced budget, using 10\% of Stage~1 training: 5,000 steps over 320,000 synthetic graphs.
All other optimization settings follow the Stage~1 configuration in \Cref{tab:pretraining-configuration}, including 5\% learning-rate warmup followed by cosine decay.
We evaluate every variant on the same 51 datasets as in the main experiments.

Because these proxy runs stop well before full pretraining, their absolute performance remains below that of the fully pretrained model, and the relative effects of individual design choices may change with further training.
We therefore use them to compare variants under a common early-training budget rather than to estimate full-scale effect sizes.
We first disentangle the contributions of the architecture and pretraining prior, and then ablate the individual design choices within each.

\mypar{Disentangling architecture and prior.}
\method{} introduces improvements along two axes: the architecture and the synthetic graph prior.
To separate their contributions, we conduct a \(3\times3\) factorial study using the architectures and priors of \method{}, GraphPFN~\citep{eremeev2026graphpfn}, and NodePFN~\citep{choi2026learning}.
We train every architecture with every prior under the same 5,000-step budget, allowing architectures to be compared under a fixed prior and priors under a fixed architecture.
GraphPFN supports both training from scratch and adaptation from a pretrained LimiX backbone; we use the latter, consistent with its released checkpoint, and train each prior from the same pretrained backbone.
Although this gives GraphPFN an initialization advantage over \method{} and NodePFN, which are trained from scratch, it more closely follows the original GraphPFN pipeline.

As shown in \Cref{fig:factorial_study}, clear differences emerge even within this reduced training budget.
Holding the prior fixed, the \method{} architecture achieves the highest mean accuracy under all three priors.
This consistency suggests that its gains are not tied to our synthetic prior.
Two explanations may contribute: message passing may provide an inductive bias better suited to graph \ac{ICL}, or its more constrained interaction structure may learn more efficiently than dense attention under the limited training budget.
Our proxy experiment does not distinguish these effects.

Holding the architecture fixed, the \method{} prior likewise performs best across all three architectures.
The comparison is informative because the three priors construct graphs differently.
NodePFN generates features before constructing graph structure from label-induced communities, whereas GraphPFN and \method{} incorporate sampled graph structure into feature and label generation.
The observed ordering does not establish whether feature-first or structure-first generation is inherently preferable, as NodePFN outperforms GraphPFN.
However, \method{} consistently improves over GraphPFN despite sharing its structure-first formulation, suggesting that the additional diversity in graph generation and explicitly sampled relational dynamics contributes beyond this high-level choice.

Overall, the factorial study shows that the gains from our architecture and prior are not specific to their pairing: each transfers across substantially different choices of the other.
We next examine the individual design choices responsible for these gains.

\begin{figure*}[t]
\centering
\begin{minipage}[t]{0.49\textwidth}
\centering
\includegraphics[width=\linewidth]{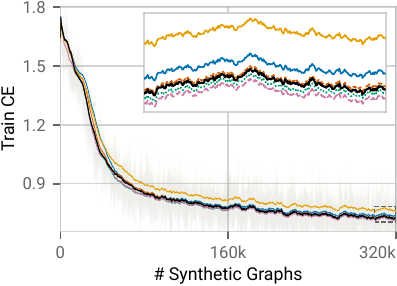}
\par\medskip
{\footnotesize\setlength{\tabcolsep}{3pt}
\renewcommand{\arraystretch}{1.1}
\begin{tabularx}{\linewidth}{@{}>{\raggedright\arraybackslash}Xrrr@{}}
\toprule
Variant & Acc. $\uparrow$ & Rank $\downarrow$ & Win $\uparrow$ \\
\midrule
\textcolor[HTML]{000000}{\rule{0.7em}{0.7em}}\,Full model & \textbf{78.91} & \textbf{2.84} & -- \\
\midrule
\textcolor[HTML]{0072B2}{\rule{0.7em}{0.7em}}\,Uniform aggregation & 78.79 & 3.03 & 37.25 \\
\textcolor[HTML]{D55E00}{\rule{0.7em}{0.7em}}\,w/o global nodes & 76.60 & 4.75 & 19.61 \\
\textcolor[HTML]{009E73}{\rule{0.7em}{0.7em}}\,w/o degree scaling & 78.48 & 3.13 & 39.22 \\
\midrule
\textcolor[HTML]{CC79A7}{\rule{0.7em}{0.7em}}\,Context-only gather & 78.07 & 2.99 & 43.14 \\
\textcolor[HTML]{E69F00}{\rule{0.7em}{0.7em}}\,w/o feature broadcast & 77.44 & 4.26 & 31.37 \\
\bottomrule
\end{tabularx}
}
\end{minipage}
\hfill
\begin{minipage}[t]{0.49\textwidth}
\centering
\includegraphics[width=\linewidth]{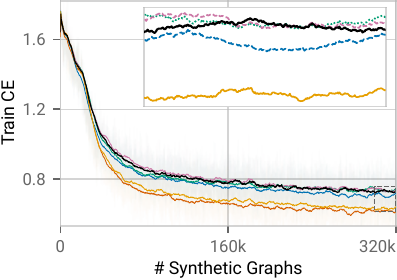}
\par\medskip
{\footnotesize\setlength{\tabcolsep}{3pt}
\renewcommand{\arraystretch}{1.1}
\begin{tabularx}{\linewidth}{@{}>{\raggedright\arraybackslash}Xrrr@{}}
\toprule
Variant & Acc. $\uparrow$ & Rank $\downarrow$ & Win $\uparrow$ \\
\midrule
\textcolor[HTML]{000000}{\rule{0.7em}{0.7em}}\,Full model & \textbf{78.91} & \textbf{2.27} & -- \\
\midrule
\textcolor[HTML]{D55E00}{\rule{0.7em}{0.7em}}\,w/o relational updates & 60.61 & 5.61 & 3.92 \\
\midrule
\textcolor[HTML]{0072B2}{\rule{0.7em}{0.7em}}\,Diffusion only & 77.64 & 3.32 & 35.29 \\
\textcolor[HTML]{009E73}{\rule{0.7em}{0.7em}}\,Cascade only & 76.89 & 3.39 & 21.57 \\
\textcolor[HTML]{CC79A7}{\rule{0.7em}{0.7em}}\,Degree only & 78.44 & 3.34 & 23.53 \\
\midrule
\textcolor[HTML]{E69F00}{\rule{0.7em}{0.7em}}\,No embedding & 77.38 & 3.06 & 33.33 \\
\bottomrule
\end{tabularx}
}
\end{minipage}
\caption{
\textbf{Architecture (left) and prior (right) ablations.}
Models are pretrained for 5,000 steps.
Curves show training cross-entropy smoothed with a trailing 100-step mean, with faint traces indicating raw losses; insets magnify the final 300k--320k training graphs.
Tables report mean accuracy (Acc., \%), average rank (Rank), and the percentage of strict wins against the full model (Win, \%) across 51 datasets under high-label regime.
Ranks are computed separately within the architecture and prior ablations, and table swatches indicate the corresponding training curves.
}
\label{fig:ablations}
\end{figure*}

\newcommand{\ablUniform}{\textcolor[HTML]{0072B2}{\rule{0.7em}{0.7em}}}
\newcommand{\ablGlobal}{\textcolor[HTML]{D55E00}{\rule{0.7em}{0.7em}}}
\newcommand{\ablDegree}{\textcolor[HTML]{009E73}{\rule{0.7em}{0.7em}}}
\newcommand{\ablContext}{\textcolor[HTML]{CC79A7}{\rule{0.7em}{0.7em}}}
\newcommand{\ablWriteback}{\textcolor[HTML]{E69F00}{\rule{0.7em}{0.7em}}}

\newcommand{\priorNoRel}{\textcolor[HTML]{D55E00}{\rule{0.7em}{0.7em}}}
\newcommand{\priorDiffusion}{\textcolor[HTML]{0072B2}{\rule{0.7em}{0.7em}}}
\newcommand{\priorCascade}{\textcolor[HTML]{009E73}{\rule{0.7em}{0.7em}}}
\newcommand{\priorDegree}{\textcolor[HTML]{CC79A7}{\rule{0.7em}{0.7em}}}
\newcommand{\priorNoEmbed}{\textcolor[HTML]{E69F00}{\rule{0.7em}{0.7em}}}

\mypar{Architecture Ablations.}
The left panel of \Cref{fig:ablations} reports training cross-entropy and downstream performance for architectural variants trained on the same prior distribution.
For the message-passing block, we replace local attention with uniform aggregation (\ablUniform), remove global nodes (\ablGlobal), and remove neighborhood-aware scaling (\ablDegree).
We further ablate our graph-aware feature refinement by restricting symmetric gathering to context nodes (\ablContext) and removing the \emph{broadcast} after message passing, preventing graph-structural information from updating the feature tokens (\ablWriteback).
All five variants degrade downstream performance, but their training losses show a different pattern.
Uniform aggregation and removing the broadcast increase training cross-entropy, whereas the other variants achieve comparable or even lower training loss despite worse downstream performance.
This contrast suggests that our architectural choices contribute not only to fitting the synthetic pretraining tasks, but also to transferring to real-world graphs.

\mypar{Prior Ablations.}
The right panel of \Cref{fig:ablations} ablates the main components of our synthetic graph prior.
We remove relational propagation entirely (\priorNoRel), restrict it to diffusion (\priorDiffusion), cascade (\priorCascade), or node-dependent mixing (\priorDegree), and remove embedding-like post-processing (\priorNoEmbed).
Removing relational propagation causes by far the largest degradation, reducing mean accuracy from 78.91\% to 60.61\%.
This confirms that sampling graph topology alone is insufficient; the topology must also shape feature and label generation during pretraining.
Restricting relational propagation to any single dynamic also underperforms their mixture, while removing embedding-like post-processing further reduces performance.
Together, these results support diversity in both relational dynamics and feature representations when constructing the synthetic graph prior.

\newpage
\section{Full Experimental Results}
\label[appendix]{app:full-results}
This section provides the complete results underlying the main evaluation in
\Cref{sec:experiments}.
We report aggregate performance across all 51 datasets, examine its trade-off
with adaptation runtime, and further analyze pairwise and subgroup results to
characterize where the observed gains arise.

\begin{figure}[t]
    \centering

    \subfloat[\textbf{Elo.}]{
        \includegraphics[width=0.97\linewidth]{results/latex/leaderboard/elo-paired.pdf}
        \label{fig:leaderboard-elo}
    }
    
    \subfloat[\textbf{Improvability.}]{
        \includegraphics[width=0.97\linewidth]{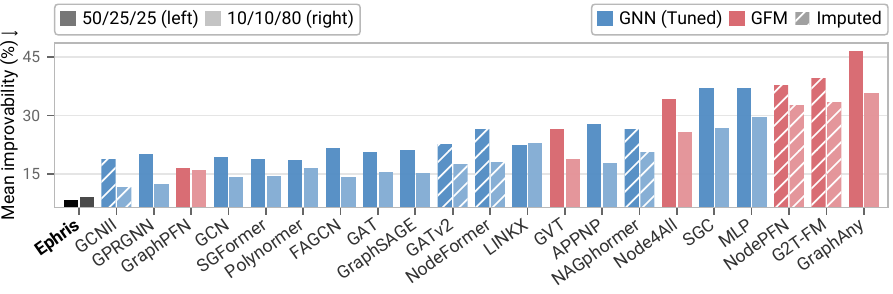}
        \label{fig:leaderboard-improvability}
    }
    
    \subfloat[\textbf{Average rank.}]{
        \includegraphics[width=0.97\linewidth]{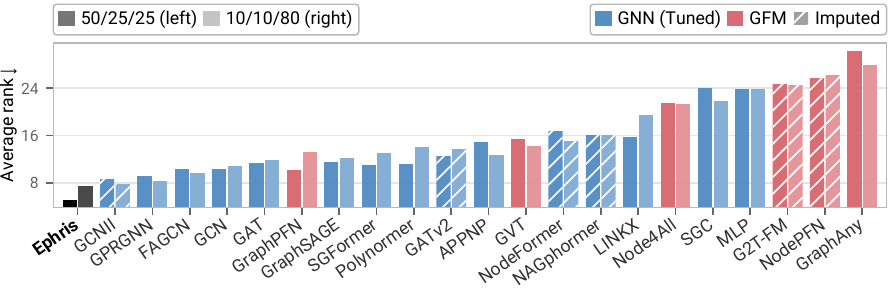}
        \label{fig:leaderboard-rank}
    }

    \subfloat[\textbf{Average accuracy.}]{
        \includegraphics[width=0.97\linewidth]{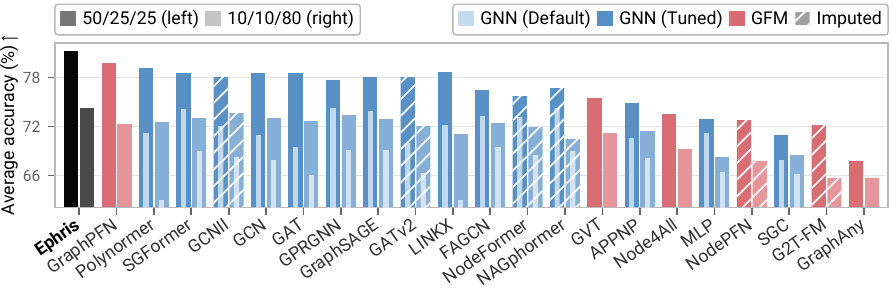}
        \label{fig:leaderboard-accuracy}
    }

    \caption{
\textbf{Aggregate performance across evaluation metrics.}
Paired comparisons under the high- and low-label regimes for
\textbf{(a)} Elo,
\textbf{(b)} improvability,
\textbf{(c)} average rank, and
\textbf{(d)} average accuracy.
Each bar shows a default \ac{GNN} alongside its \ac{HP}-tuned counterpart.
Hatching marks results imputed for OOM runs.
Methods are ordered by their mean performance across the two label regimes.
}
    \label{fig:bar-plot-full}
\end{figure}

\mypar{Aggregate performance.}
\Cref{tab:leaderboard-paired,fig:bar-plot-full} report the complete aggregate
results under the high-label (50/25/25) and low-label (10/10/80) regimes.
\method{} ranks first in all four metrics in both regimes: Elo, improvability,
average rank, and average accuracy.
In the high-label regime, \method{} achieves an Elo of 1521 compared with
1388 for the runner-up tuned GCNII, while reducing mean improvability from
18.81\% to 8.29\% and average rank from 8.57 to 5.13.
It also achieves the highest average accuracy at 81.27\%.
The margins narrow under the low-label regime, where \method{} achieves an
Elo of 1381 compared with 1371 for the runner-up tuned GCNII, but remains
first in improvability, average rank, and average accuracy at 9.06\%, 7.55,
and 74.23\%, respectively.

HP tuning substantially improves many supervised GNNs, with tuned GCNII and
GPRGNN emerging as the strongest GNN baselines across the two regimes.
Among prior graph ICL methods, GraphPFN is consistently the strongest.
We therefore use these methods as the primary reference points in the
comparisons below.

\begin{figure}[t]
    \centering

    \subfloat[\textbf{Elo.}]{
        \includegraphics[width=0.98\linewidth]{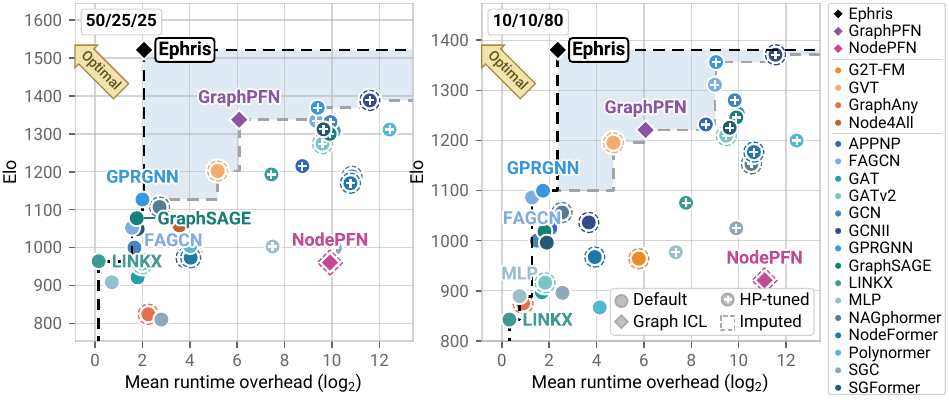}
        \label{fig:pareto-elo}
    }

    \subfloat[\textbf{Average rank.}]{
        \includegraphics[width=0.98\linewidth]{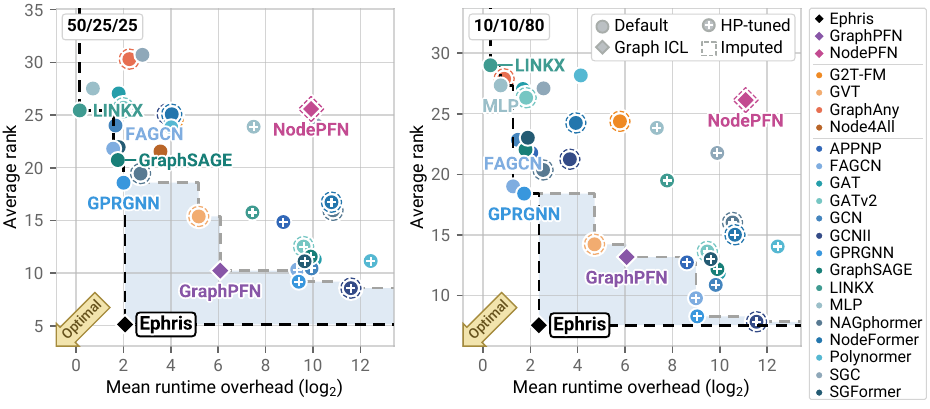}
        \label{fig:pareto-rank}
    }

    \subfloat[\textbf{Average accuracy.}]{
        \includegraphics[width=0.98\linewidth]{results/latex/main_plot/fit-time-average-rank-paired.pdf}
        \label{fig:pareto-accuracy}
    }

    \caption{
    \textbf{Performance-runtime Pareto plots.}
    Trade-offs between average runtime overhead and
    \textbf{(a)} Elo,
    \textbf{(b)} average rank, and
    \textbf{(c)} average accuracy.
    Each point represents an evaluated method, and the Pareto frontier identifies methods that are not jointly dominated in performance and runtime.
    }
    \label{fig:pareto-full}
\end{figure}

\mypar{Performance and runtime.}
The predictive gains of \method{} also come with substantially lower
adaptation cost.
Runtime overhead is aggregated geometrically across datasets, so differences
on the $\log_2$ scale in \Cref{tab:leaderboard-paired} correspond directly
to multiplicative differences in average runtime.
Compared with the strongest tuned GNNs, \method{} is approximately
$739\times$ and $584\times$ faster than GCNII in the high-label and
low-label regimes, respectively, and $161\times$ and $103\times$ faster
than GPRGNN.
Compared with GraphPFN, the strongest prior graph ICL baseline,
\method{} is approximately $16.1\times$ and $13.1\times$ faster while
achieving stronger predictive performance across all four aggregate metrics.
Accordingly, \method{} occupies a favorable performance and runtime region
across the metrics in \Cref{fig:pareto-full}, improving predictive performance
without increasing target-time computation.

\begin{figure*}[t]
    \centering

    \begin{minipage}{0.49\linewidth}
        \centering
        \includegraphics[width=\linewidth]{
        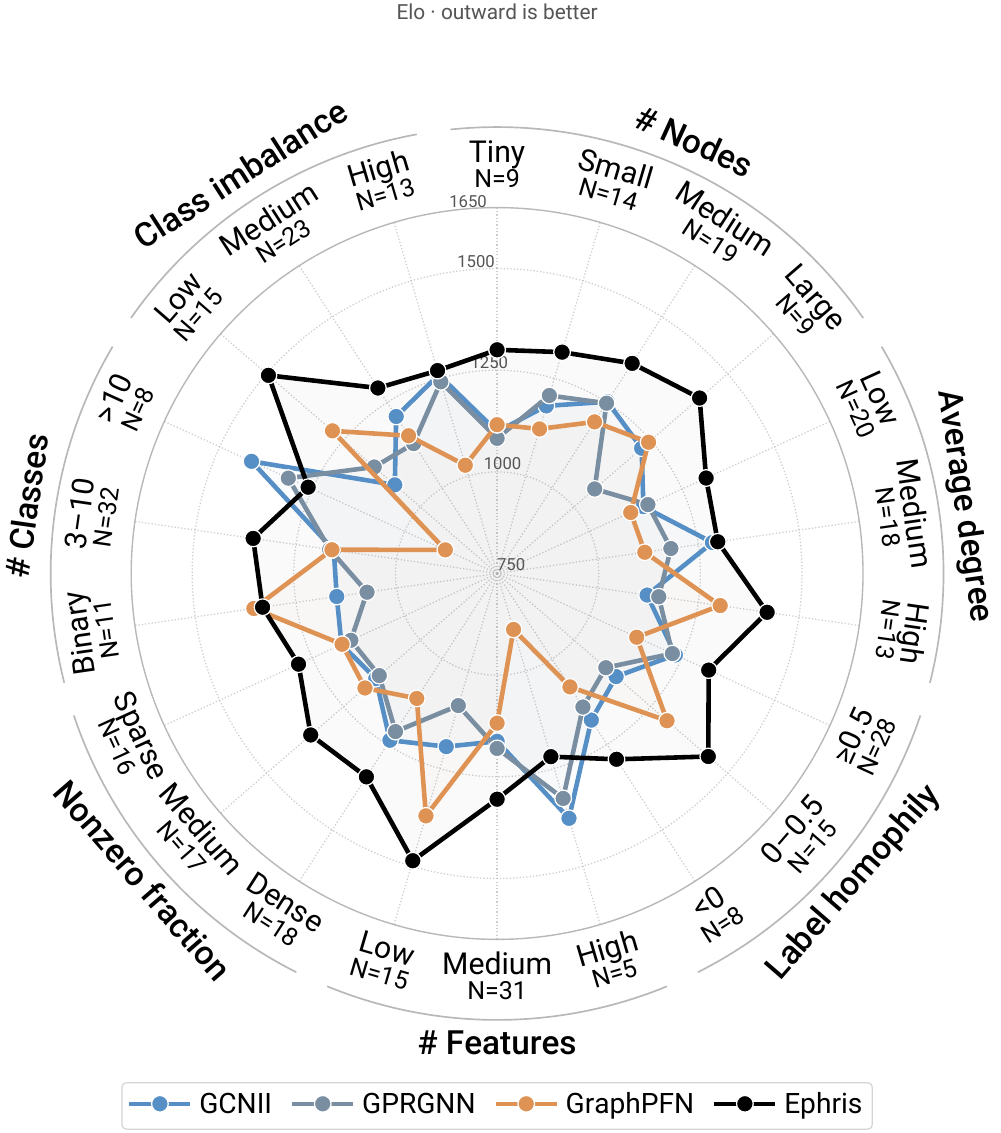}
        {\small (a) Elo}
    \end{minipage}
    \hfill
    \begin{minipage}{0.49\linewidth}
        \centering
        \includegraphics[width=\linewidth]{
        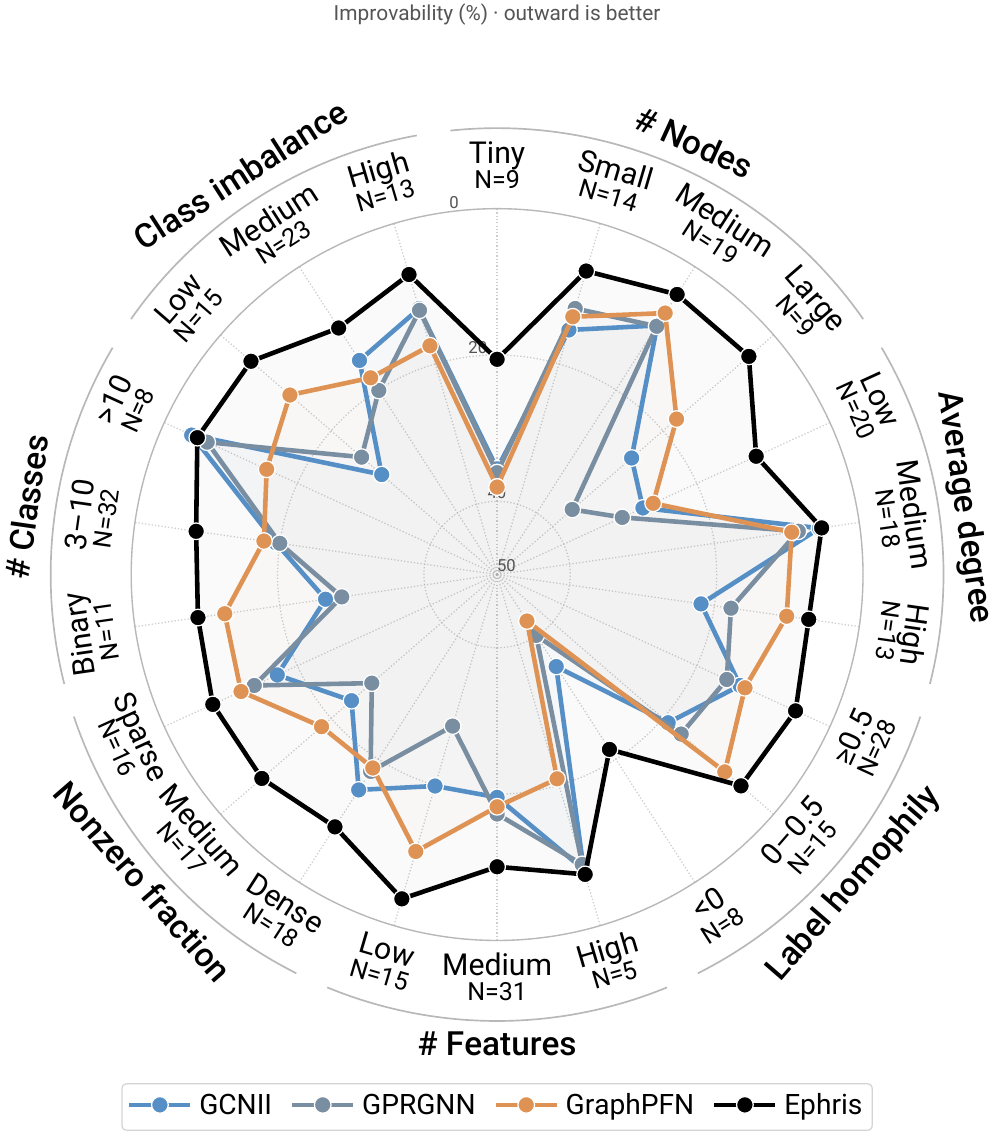}
        {\small (b) Improvability}
    \end{minipage}

    \vspace{2mm}

    \begin{minipage}{0.49\linewidth}
        \centering
        \includegraphics[width=\linewidth]{
        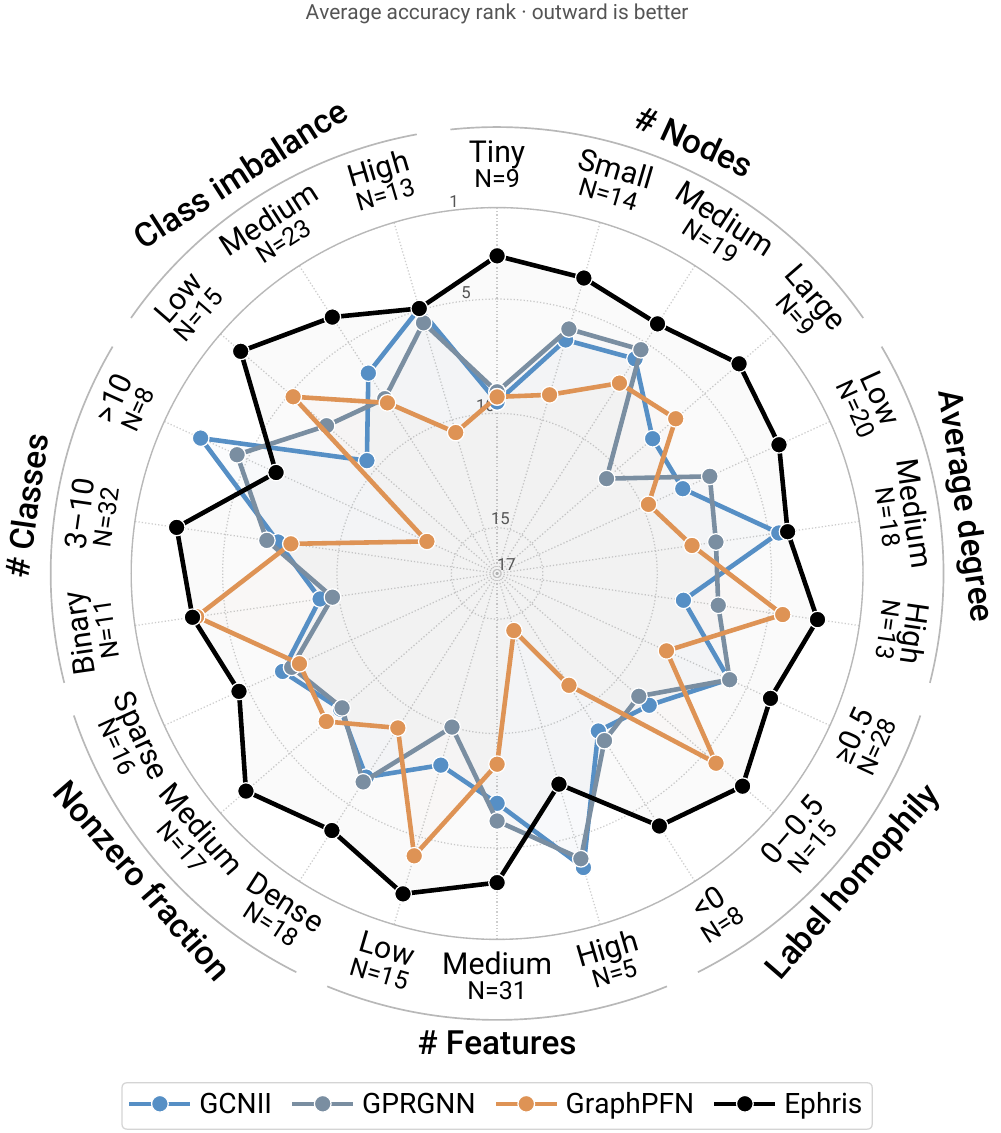}
        {\small (c) Average rank}
    \end{minipage}
    \hfill
    \begin{minipage}{0.49\linewidth}
        \centering
        \includegraphics[width=\linewidth]{
        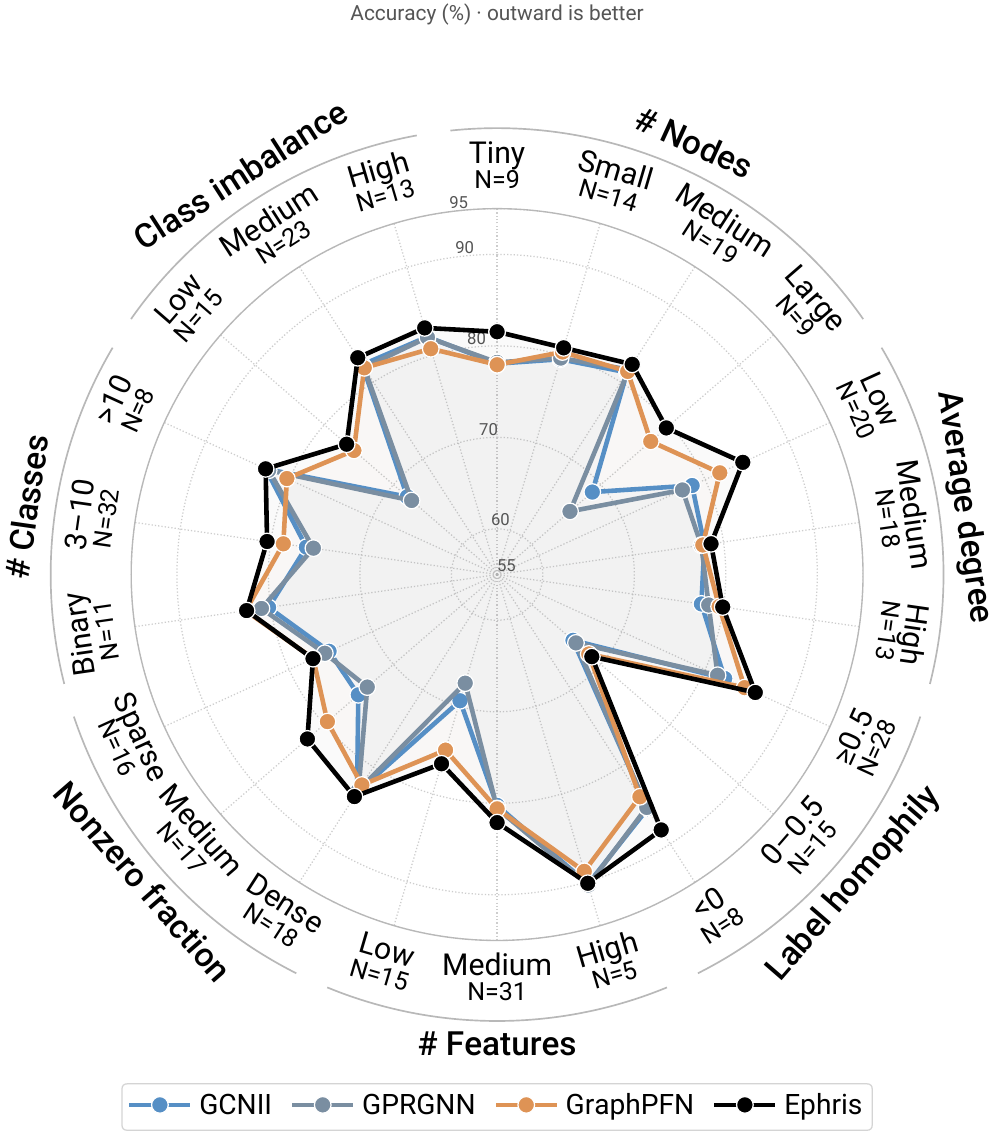}
        {\small (d) Average accuracy}
    \end{minipage}

    \caption{
    \textbf{Performance across dataset subgroups under the high-label regime (50/25/25).}
    Radar plots show subgroup performance measured by
    (a) Elo,
    (b) improvability,
    (c) average rank, and
    (d) average accuracy.
    Each axis corresponds to a dataset subgroup, providing a more detailed view of performance across different dataset characteristics.
    }
    \label{fig:subgroup-radar-high}
\end{figure*}

\begin{figure*}[t]
    \centering

    \begin{minipage}{0.49\linewidth}
        \centering
        \includegraphics[width=\linewidth]{
        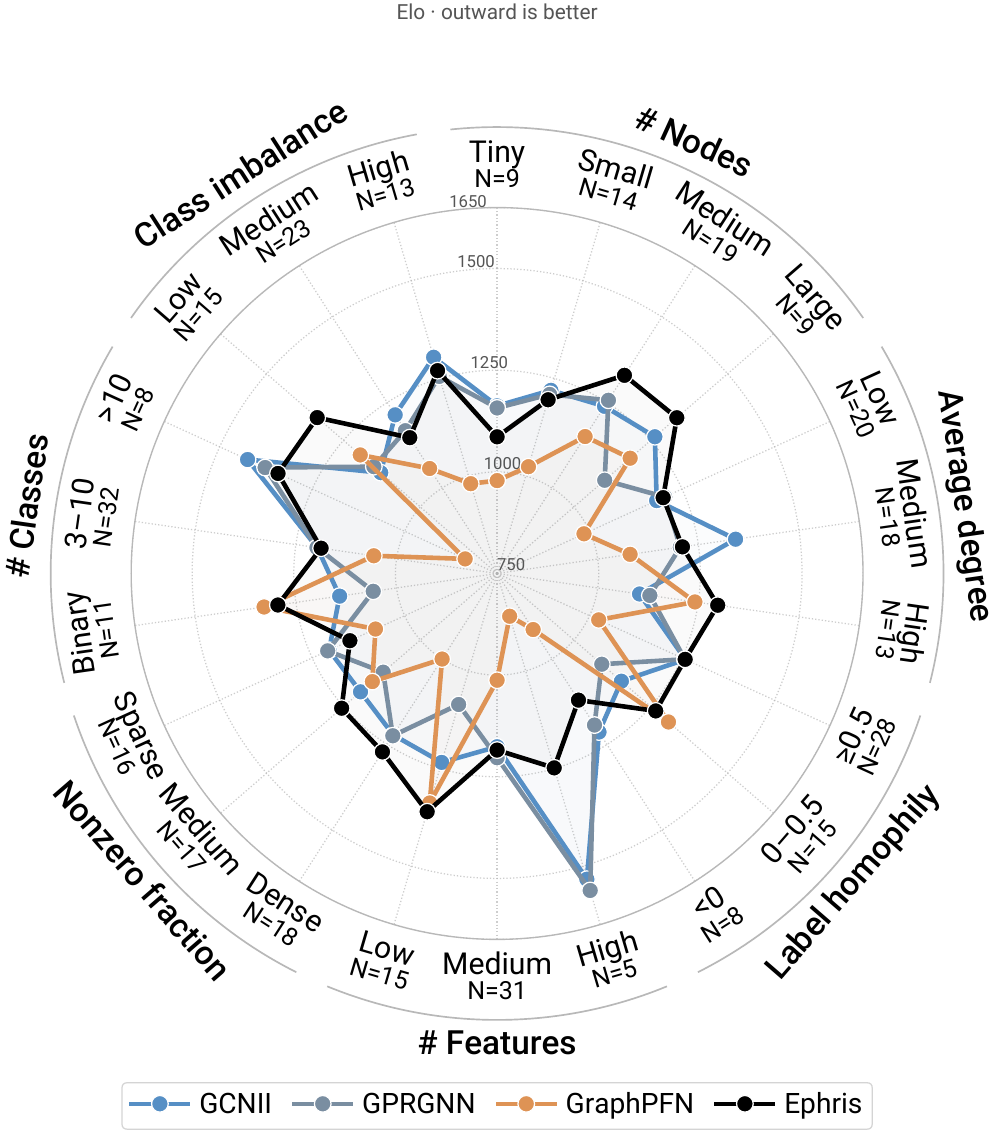}
        {\small (a) Elo}
    \end{minipage}
    \hfill
    \begin{minipage}{0.49\linewidth}
        \centering
        \includegraphics[width=\linewidth]{
        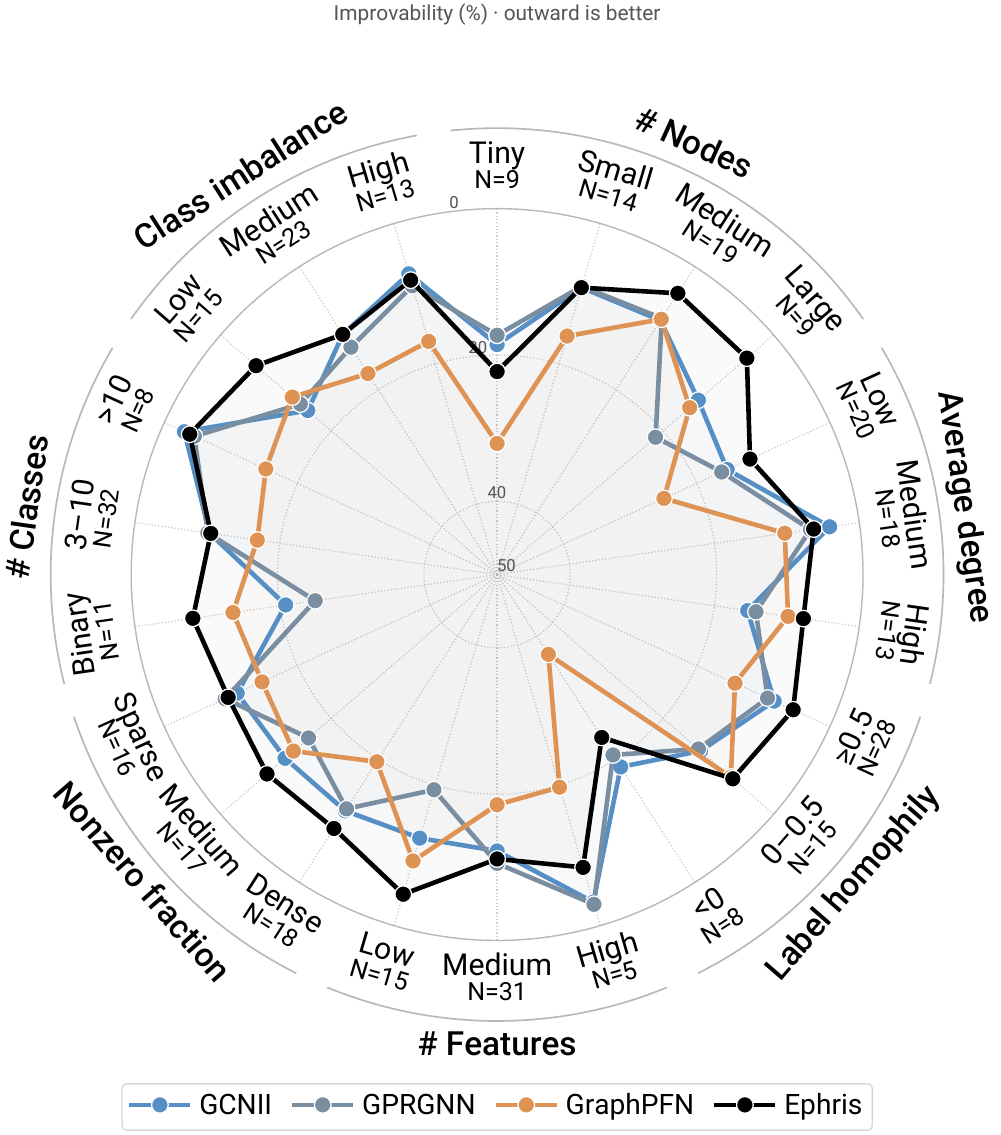}
        {\small (b) Improvability}
    \end{minipage}

    \vspace{2mm}

    \begin{minipage}{0.49\linewidth}
        \centering
        \includegraphics[width=\linewidth]{
        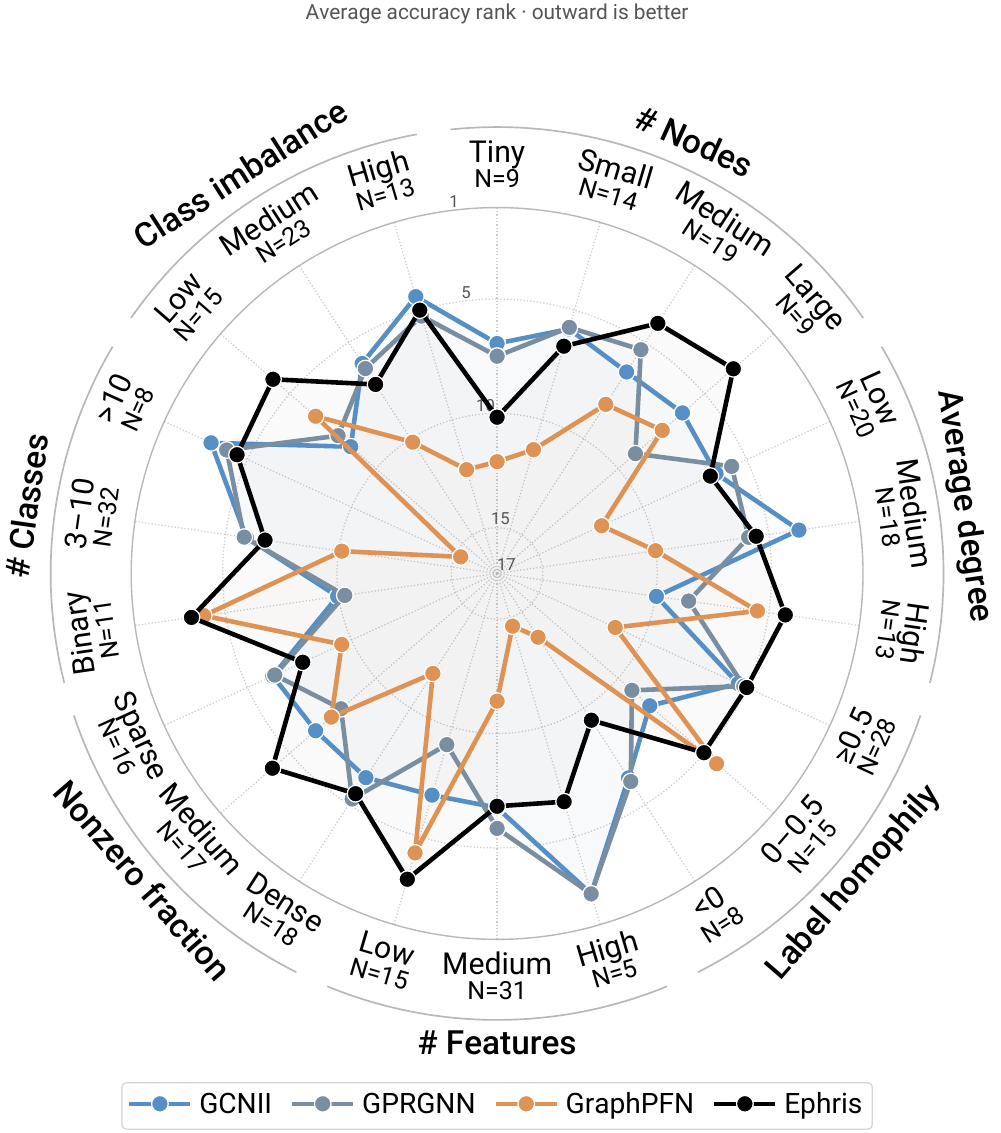}
        {\small (c) Average rank}
    \end{minipage}
    \hfill
    \begin{minipage}{0.49\linewidth}
        \centering
        \includegraphics[width=\linewidth]{
        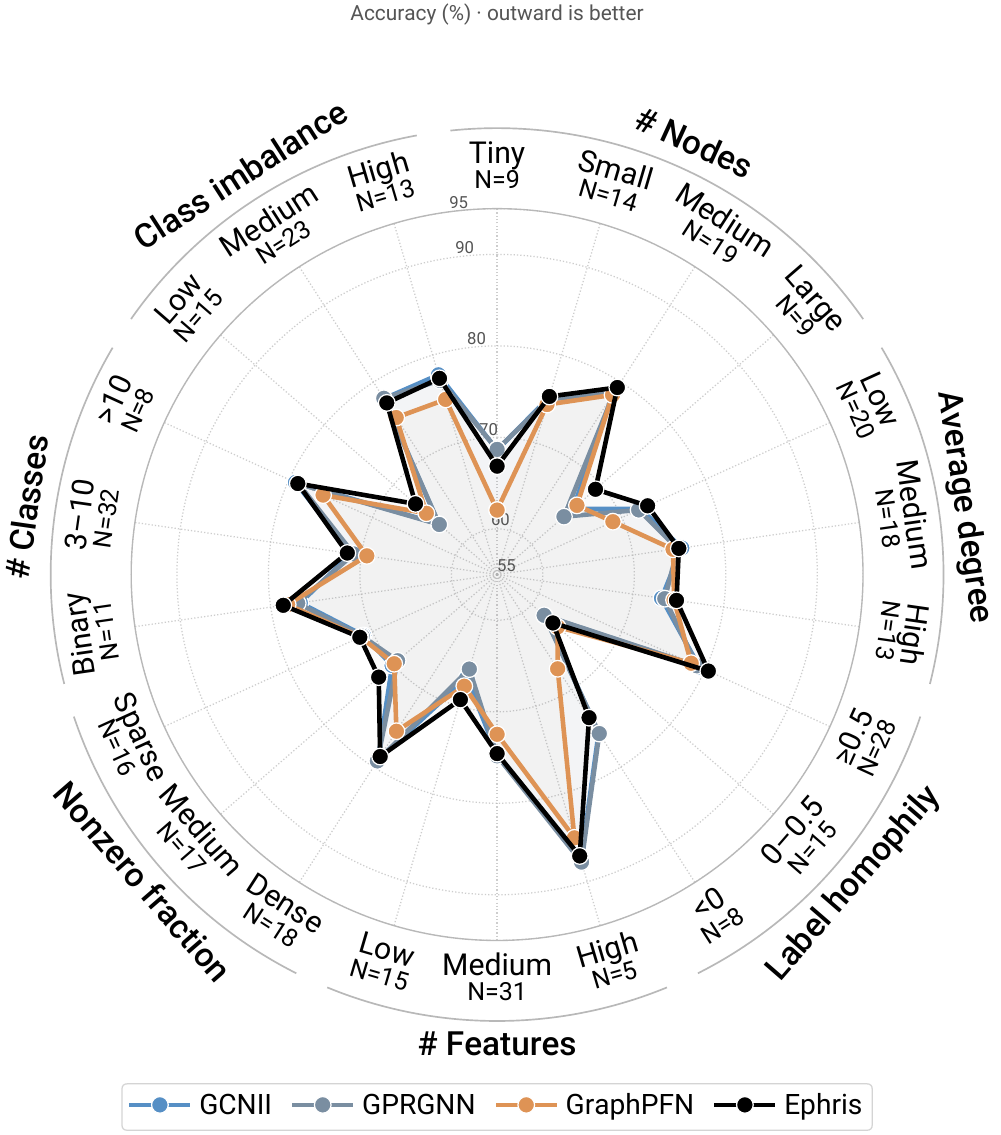}
        {\small (d) Average accuracy}
    \end{minipage}

    \caption{
    \textbf{Performance across dataset subgroups under the low-label regime (10/10/80).}
    Radar plots show subgroup performance measured by
    (a) Elo,
    (b) improvability,
    (c) average rank, and
    (d) average accuracy.
    Each axis corresponds to a dataset subgroup, providing a more detailed view of performance across different dataset characteristics.
    }
    \label{fig:subgroup-radar-low}
\end{figure*}

\mypar{Performance across dataset characteristics.}
\Cref{fig:subgroup-radar-high,fig:subgroup-radar-low} compare performance
across subgroups defined by the dataset characteristics in
\Cref{tab:subgroup-definitions}.

\begin{table}[h]
\centering
\small
\vspace{-4pt}
\caption{\textbf{Dataset subgroup definitions.}}
\vspace{-4pt}
\label{tab:subgroup-definitions}
\setlength{\tabcolsep}{4pt}
\renewcommand{\arraystretch}{1.08}
\begin{tabular}{@{}ll@{}}
\toprule
Characteristic & Subgroups \\
\midrule
Nodes $N$
& Tiny ($\leq2$K), Small ($2$K to $10$K), Medium ($10$K to $100$K), Large ($>100$K) \\
Avg. degree $\bar d$
& Low ($\leq5$), Medium ($5$ to $20$), High ($>20$) \\
Adj. homophily $h_{\mathrm{adj}}$
& Negative ($<0$), Low ($0$ to $0.5$), High ($\geq0.5$) \\
Features $F$
& Low ($\leq500$), Medium ($500$ to $5$K), High ($\geq5$K) \\
Nonzero fraction $r$
& Sparse ($\leq1\%$), Medium ($1\%$ to $50\%$), Dense ($>50\%$) \\
Classes $C$
& Binary ($2$), Medium ($3$ to $10$), Many ($>10$) \\
Class imbalance
& Low ($\leq2$), Medium ($2$ to $10$), High ($>10$) \\
\bottomrule
\end{tabular}
\end{table}

Across these groups, \method{} outperforms GraphPFN in nearly
every subgroup under both label regimes.
This consistent advantage indicates that the improvement of graph ICL
is not tied to a particular graph size, connectivity pattern, homophily level,
or task characteristic.

A different pattern emerges against the strongest tuned GNNs.
In the high-label regime, \method{} outperforms GCNII and GPRGNN across
nearly all subgroups, whereas under low labels the methods become comparable
and tuned GNNs lead in several cases.
One possible explanation is validation-based adaptation: GCNII and GPRGNN
select among 200 configurations for each dataset using validation labels,
whereas \method{} uses a single fixed checkpoint and no validation labels.
This dataset-specific selection may become particularly valuable when
labeled context is low.

Notably, the two clear exceptions even in the high-label regime are
high-dimensional datasets ($F\geq5$K) and tasks with more than ten classes
($C>10$).
Both fall outside the distribution directly observed by \method{} during
pretraining, which contains at most 1,024 features and ten classes; larger
label spaces are instead handled indirectly through ECOC.
GraphPFN shows the same failure pattern in these two subgroups.
This shared degradation suggests a broader limitation of current graph ICL:
generalization remains strongest within the support of the synthetic
pretraining distribution and weakens when feature or label dimensionality
requires substantial extrapolation.

\begin{table*}[t]
\centering
\definecolor{lbGold}{HTML}{D6AC36}
\definecolor{lbSilver}{HTML}{B6BEC8}
\definecolor{lbBronze}{HTML}{BE8457}
\caption{Comparison across 51 datasets under the high-label (50/25/25, left) and low-label (10/10/80, right) train/validation/test regimes. Each side lists all configurations independently in descending Elo order; entries on the same row need not denote the same method. (D)/(T) denote default/HP-tuned configurations. Imp is mean improvability (\%), Acc.\ is mean accuracy (\%), and Time is mean fit-time overhead on a $\log_2$ scale. \textcolor{lbGold}{\ensuremath{\bullet}}\,Gold, \textcolor{lbSilver}{\ensuremath{\bullet}}\,silver, and \textcolor{lbBronze}{\ensuremath{\bullet}}\,bronze mark the top three in each predictive metric separately within each regime. Under high-label, Ephris ranks 1 in Elo, 1 in improvability, 1 in average rank, and 1 in accuracy. Under low-label, Ephris ranks 1 in Elo, 1 in improvability, 1 in average rank, and 1 in accuracy.}
\label{tab:leaderboard-paired}
\vspace{4pt}
\begingroup
\fontsize{7.5}{9}\selectfont
\setlength{\tabcolsep}{2pt}
\renewcommand{\arraystretch}{1.08}
\begin{adjustbox}{width=\linewidth}
\begin{tabular}{@{}lrrrrr@{\hspace{5.8pt}\vrule width 0.4pt\hspace{5.8pt}}lrrrrr@{}}
\toprule
\multicolumn{6}{c}{\textbf{High-label (50/25/25)}} & \multicolumn{6}{c}{\textbf{Low-label (10/10/80)}} \\
\cmidrule(lr){1-6}\cmidrule(lr){7-12}
Method & Elo $\uparrow$ & Imp $\downarrow$ & Rank $\downarrow$ & Acc. $\uparrow$ & Time $\downarrow$ & Method & Elo $\uparrow$ & Imp $\downarrow$ & Rank $\downarrow$ & Acc. $\uparrow$ & Time $\downarrow$ \\
\midrule
\textbf{Ephris} & \textcolor{lbGold}{\ensuremath{\bullet}}\,\textbf{1521} & \textcolor{lbGold}{\ensuremath{\bullet}}\,\textbf{8.29} & \textcolor{lbGold}{\ensuremath{\bullet}}\,\textbf{5.13} & \textcolor{lbGold}{\ensuremath{\bullet}}\,\textbf{81.27} & \textbf{2.07} & \textbf{Ephris} & \textcolor{lbGold}{\ensuremath{\bullet}}\,\textbf{1381} & \textcolor{lbGold}{\ensuremath{\bullet}}\,\textbf{9.06} & \textcolor{lbGold}{\ensuremath{\bullet}}\,\textbf{7.55} & \textcolor{lbGold}{\ensuremath{\bullet}}\,\textbf{74.23} & \textbf{2.36} \\
GCNII (T) & \textcolor{lbSilver}{\ensuremath{\bullet}}\,1388 & 18.81 & \textcolor{lbSilver}{\ensuremath{\bullet}}\,8.57 & 78.02 & 11.60 & GCNII (T) & \textcolor{lbSilver}{\ensuremath{\bullet}}\,1371 & \textcolor{lbSilver}{\ensuremath{\bullet}}\,11.58 & \textcolor{lbSilver}{\ensuremath{\bullet}}\,7.83 & \textcolor{lbSilver}{\ensuremath{\bullet}}\,73.59 & 11.55 \\
GPRGNN (T) & \textcolor{lbBronze}{\ensuremath{\bullet}}\,1369 & 20.04 & \textcolor{lbBronze}{\ensuremath{\bullet}}\,9.18 & 77.65 & 9.40 & GPRGNN (T) & \textcolor{lbBronze}{\ensuremath{\bullet}}\,1356 & \textcolor{lbBronze}{\ensuremath{\bullet}}\,12.54 & \textcolor{lbBronze}{\ensuremath{\bullet}}\,8.28 & \textcolor{lbBronze}{\ensuremath{\bullet}}\,73.42 & 9.05 \\
GraphPFN & 1337 & \textcolor{lbSilver}{\ensuremath{\bullet}}\,16.55 & 10.24 & \textcolor{lbSilver}{\ensuremath{\bullet}}\,79.76 & 6.08 & FAGCN (T) & 1311 & 14.18 & 9.78 & 72.45 & 9.00 \\
FAGCN (T) & 1334 & 21.59 & 10.33 & 76.50 & 9.33 & GCN (T) & 1280 & 14.18 & 10.88 & 73.00 & 9.83 \\
GCN (T) & 1331 & 19.27 & 10.45 & 78.49 & 9.94 & GAT (T) & 1254 & 15.43 & 11.86 & 72.65 & 9.96 \\
SGFormer (T) & 1312 & 18.93 & 11.12 & 78.54 & 9.65 & GraphSAGE (T) & 1246 & 15.15 & 12.18 & 72.91 & 9.90 \\
Polynormer (T) & 1311 & \textcolor{lbBronze}{\ensuremath{\bullet}}\,18.61 & 11.15 & \textcolor{lbBronze}{\ensuremath{\bullet}}\,79.14 & 12.44 & APPNP (T) & 1232 & 17.92 & 12.73 & 71.45 & 8.61 \\
GAT (T) & 1307 & 20.59 & 11.31 & 78.50 & 10.08 & SGFormer (T) & 1225 & 14.58 & 13.00 & 73.10 & 9.62 \\
GraphSAGE (T) & 1300 & 21.23 & 11.57 & 78.08 & 9.91 & GraphPFN & 1221 & 16.05 & 13.18 & 72.30 & 6.07 \\
GATv2 (T) & 1274 & 22.52 & 12.52 & 77.96 & 9.60 & GATv2 (T) & 1210 & 17.49 & 13.64 & 72.02 & 9.47 \\
APPNP (T) & 1215 & 27.69 & 14.85 & 74.87 & 8.75 & Polynormer (T) & 1200 & 16.43 & 14.05 & 72.59 & 12.45 \\
GVT & 1202 & 26.56 & 15.38 & 75.43 & 5.18 & GVT & 1195 & 18.76 & 14.23 & 71.19 & 4.71 \\
LINKX (T) & 1193 & 22.30 & 15.76 & 78.71 & 7.44 & NodeFormer (T) & 1177 & 17.85 & 15.01 & 71.90 & 10.64 \\
NAGphormer (T) & 1188 & 26.48 & 15.96 & 76.60 & 10.85 & NAGphormer (T) & 1153 & 20.39 & 16.03 & 70.41 & 10.54 \\
NodeFormer (T) & 1170 & 26.38 & 16.73 & 75.70 & 10.78 & GPRGNN (D) & 1100 & 23.02 & 18.41 & 69.15 & 1.74 \\
GPRGNN (D) & 1127 & 29.52 & 18.58 & 74.25 & 2.00 & FAGCN (D) & 1086 & 22.29 & 19.01 & 69.48 & 1.28 \\
NAGphormer (D) & 1107 & 29.84 & 19.45 & 74.16 & 2.72 & LINKX (T) & 1076 & 22.98 & 19.49 & 71.09 & 7.78 \\
GraphSAGE (D) & 1078 & 31.09 & 20.73 & 73.95 & 1.76 & NAGphormer (D) & 1056 & 23.07 & 20.36 & 68.92 & 2.54 \\
Node4All & 1059 & 34.17 & 21.55 & 73.47 & 3.56 & GCNII (D) & 1036 & 25.26 & 21.25 & 68.18 & 3.67 \\
FAGCN (D) & 1053 & 31.65 & 21.81 & 73.34 & 1.56 & Node4All & 1034 & 25.72 & 21.36 & 69.29 & 3.49 \\
SGFormer (D) & 1049 & 32.09 & 21.96 & 74.12 & 1.80 & APPNP (D) & 1025 & 25.69 & 21.77 & 68.15 & 2.04 \\
Polynormer (D) & 1004 & 36.55 & 23.84 & 71.21 & 4.02 & SGC (T) & 1025 & 26.70 & 21.77 & 68.51 & 9.89 \\
MLP (T) & 1003 & 37.10 & 23.89 & 72.93 & 7.49 & GraphSAGE (D) & 1018 & 24.73 & 22.04 & 69.15 & 1.80 \\
GCN (D) & 1000 & 36.04 & 24.01 & 70.98 & 1.66 & GCN (D) & 1000 & 27.06 & 22.85 & 67.93 & 1.48 \\
SGC (T) & 999 & 37.11 & 24.06 & 70.98 & 10.13 & SGFormer (D) & 996 & 26.67 & 23.02 & 69.04 & 1.90 \\
G2T-FM & 986 & 39.40 & 24.56 & 72.11 & 4.09 & MLP (T) & 977 & 29.51 & 23.83 & 68.31 & 7.34 \\
NodeFormer (D) & 973 & 35.36 & 25.08 & 73.14 & 4.04 & NodeFormer (D) & 967 & 28.26 & 24.24 & 68.46 & 3.92 \\
GCNII (D) & 973 & 36.43 & 25.11 & 72.00 & 3.84 & G2T-FM & 964 & 33.26 & 24.38 & 65.69 & 5.79 \\
LINKX (D) & 964 & 38.21 & 25.45 & 72.15 & 0.15 & NodePFN & 921 & 32.51 & 26.13 & 67.75 & 11.09 \\
NodePFN & 961 & 37.79 & 25.58 & 72.71 & 9.92 & GATv2 (D) & 916 & 30.61 & 26.32 & 66.27 & 1.83 \\
GATv2 (D) & 958 & 38.20 & 25.69 & 69.96 & 2.00 & GAT (D) & 897 & 31.17 & 27.05 & 66.09 & 1.69 \\
APPNP (D) & 955 & 37.82 & 25.80 & 70.55 & 2.16 & SGC (D) & 896 & 33.42 & 27.10 & 66.23 & 2.56 \\
GAT (D) & 922 & 40.22 & 27.05 & 69.47 & 1.80 & MLP (D) & 889 & 33.60 & 27.34 & 66.49 & 0.75 \\
MLP (D) & 909 & 41.47 & 27.52 & 71.21 & 0.71 & GraphAny & 875 & 35.83 & 27.88 & 65.74 & 0.90 \\
GraphAny & 824 & 46.59 & 30.31 & 67.77 & 2.25 & Polynormer (D) & 867 & 36.30 & 28.16 & 63.05 & 4.13 \\
SGC (D) & 810 & 46.14 & 30.72 & 67.86 & 2.80 & LINKX (D) & 843 & 40.84 & 29.00 & 62.99 & 0.32 \\
\bottomrule
\end{tabular}
\end{adjustbox}
\endgroup
\end{table*}

\begin{figure}[t]
    \centering
    \includegraphics[width=0.96\linewidth]{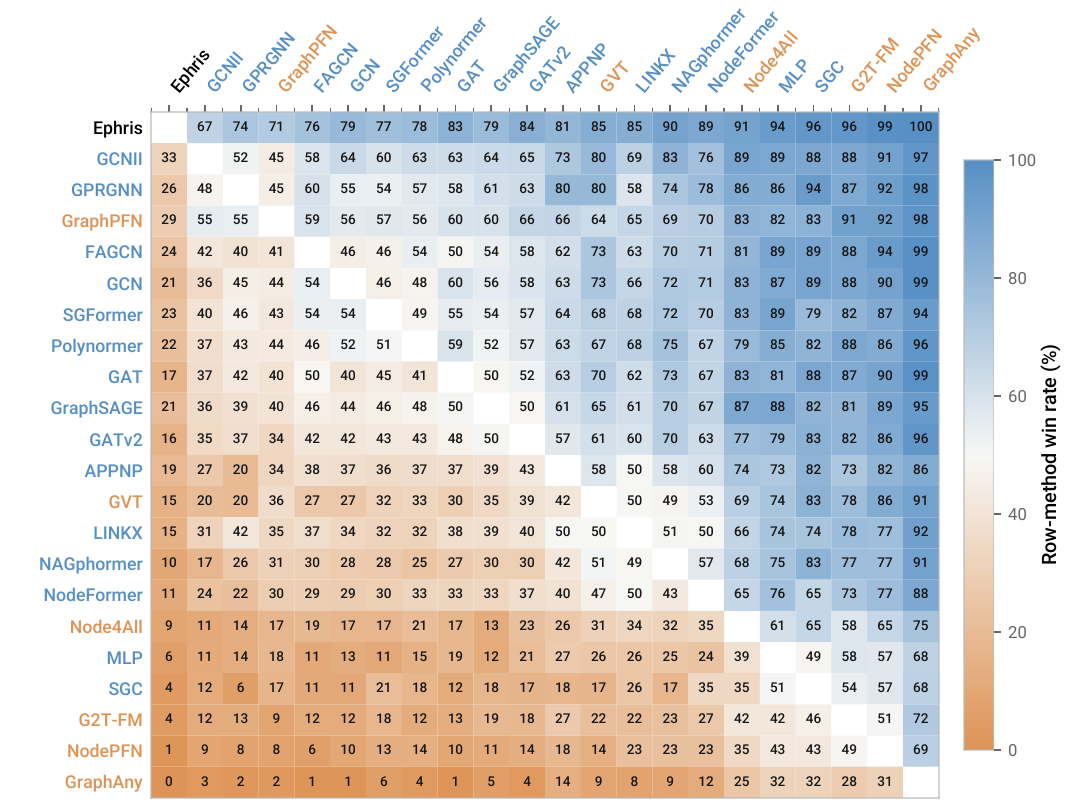}
    {\small (a) High-label regime (50/25/25)}

    \includegraphics[width=0.96\linewidth]{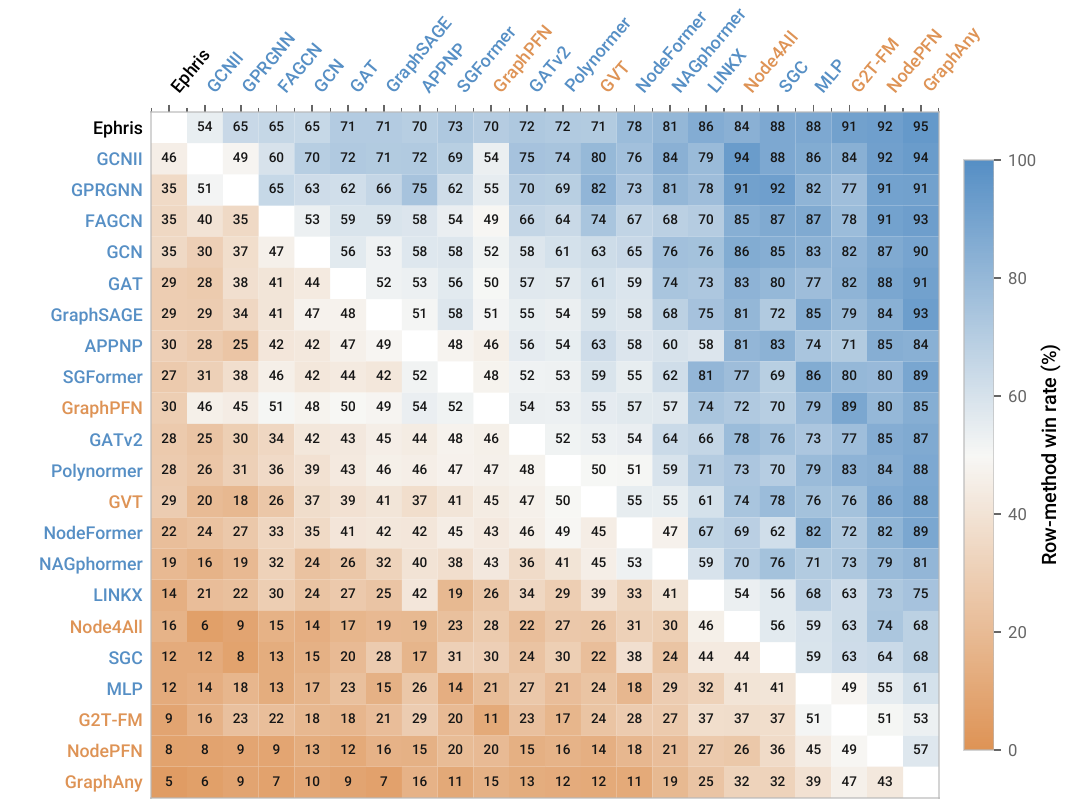}
    {\small (b) Low-label regime (10/10/80)}

    \caption{
    \textbf{Pairwise win-rate matrices.}
    Pairwise comparisons under
    (a) the high-label and
    (b) the low-label regime.
    Each entry reports the fraction of datasets on which the row method achieves higher test accuracy than the column method.
    }
    \label{fig:win-rate-full}
\end{figure}

\end{document}